\documentclass{article}
\usepackage[numbers]{natbib}
\usepackage{vida_arxiv,times}

\usepackage[utf8]{inputenc} 
\usepackage[T1]{fontenc}    
\usepackage{hyperref}       
\usepackage{url}            
\usepackage{booktabs}       
\usepackage{amsfonts}       
\usepackage{nicefrac}       
\usepackage{microtype}      
\usepackage{xcolor}         
\usepackage{placeins}

\usepackage{multirow}
\usepackage{makecell}
\usepackage{graphicx}
\usepackage{subcaption}

\definecolor{mygreen}{RGB}{0,120,60}
\definecolor{myred}{RGB}{180,30,30}

\newcommand{\cmark}{\ding{51}}  
\newcommand{\xmark}{\ding{55}}  
\usepackage{booktabs, multirow, xcolor, amssymb, pifont}

\usepackage{tabularx}
\usepackage{array}
\usepackage{nameref}

\usepackage[ruled,vlined,linesnumbered]{algorithm2e}
\usepackage{enumitem}

\usepackage{etoc}
\usepackage{amsmath}

\usepackage{tcolorbox}
\usepackage{cleveref}

\usepackage{float}

\newtcolorbox{methodbox}[1]{
    colback=gray!8,      
    colframe=black!70,   
    boxrule=0.8pt,       
    arc=3pt,             
    left=6pt,
    right=6pt,
    top=6pt,
    bottom=6pt,
    title=#1,
    fonttitle=\bfseries,
}

\etocdepthtag.toc{main}

\newcolumntype{C}[1]{>{\centering\arraybackslash}p{#1}}
\newcolumntype{Y}{>{\centering\arraybackslash}X}

\title{From Pixels to Policy: A Multi-Agent System for Intervention and Geo-Spatial Decision Support}

\iclrfinalcopy

\author{%
  \begin{tabular}{c@{\hspace{1.2cm}}c}
    \textbf{Hosam Elgendy}
    &
    \textbf{Utkarsh Mall}
    \\[2pt]
    \texttt{hosam.elgendy@mbzuai.ac.ae}
    &
    \texttt{utkarsh.mall@mbzuai.ac.ae}
  \end{tabular}
  \\[10pt]
  Mohamed bin Zayed University of Artificial Intelligence, Abu Dhabi, UAE
}

\begin{document}
\newcommand{\utkarsh}[1]{{\textcolor{red}{UM\@: {#1}}}}
\newcommand{\hosam}[1]{\textcolor{magenta}{HE\@: {#1}}}
\newcommand{\compactpara}[1]
{\noindent \textbf{#1}}

\newcommand{\vida}{\textsc{VIDA-Geo}}
\newcommand{\dashfill}{\leavevmode\leaders\hbox{-}\hfill\kern0pt}
\definecolor{myorange}{RGB}{200, 90, 0}
\definecolor{mydarkblue}{RGB}{0, 60, 125}     
\definecolor{mymagenta}{RGB}{200, 0, 150}     

\newcommand{\orange}[1]{\textcolor{myorange}{#1}}
\newcommand{\darkblue}[1]{\textcolor{mydarkblue}{#1}}
\newcommand{\magenta}[1]{\textcolor{mymagenta}{#1}}

\newenvironment{packed_item}{
\begin{itemize}[leftmargin=1em, itemsep=1pt, topsep=0pt, parsep=0pt, partopsep=0pt]
}{\end{itemize}}

\maketitle

\begin{abstract}
Urban environments are shaped by design choices with long-term implications for health, safety, and quality of life, yet evaluating proposed interventions remains costly, time-consuming, and often impractical.  
Existing geospatial vision methods largely focus on monitoring urban indicators from aerial and street-view imagery, rather than proposing interventions and estimating their effects on such indicators.
Moving beyond recognition, we introduce the problem of discovering interventions that improve target indicators for a given aerial or street-view image.
We argue that a black-box indicator model, combined with a generative editing model, can serve as an \emph{implicit digital twin} for testing intervention hypotheses.
We present \vida~, a multi-agent system that explores this intervention space by coordinating segmentation, diffusion-based inpainting, and indicator scoring models to produce interventions that are both perceptually realistic and aligned with real-world policies.
We evaluate our system on 8 indicators across aerial and street-view imagery, measuring changes in factors such as perceived safety and greenery.
Our approach outperforms existing baselines in many cases, achieving up to $2\times$ higher perceptual quality and policy alignment scores.
Finally, our model provides users with multiple candidate interventions, supporting an expert city-planner-in-the-loop workflow. Code is available at \texttt{https://github.com/HosamGen/VIDA-GEO} 

\end{abstract}

\section{Introduction}
\label{sec:intro}
Visual geospatial data is used in several real-world applications for monitoring and decision-making.
Urban planners make use of satellite images to monitor various indicators such as road safety~\citep{road_1,elallaf2026betarisk} and greenery~\citep{green_1}.
Similarly, climate scientists make use of satellite images to understand pollution~\citep{pollution_1,pollution_3}, perceived temperatures~\citep{perceived_1,perceived_2} and thermal exposure~\citep{thermal_1,thermal_2}.
Beyond satellite images, experts also make use of other visual information, such as street view imagery or surveillance footage, to monitor perceived safety~\citep{safety_1,safety_2}, road-safety~\citep{traffic_1,traffic_2}, or walkability~\citep{walk_2,walk_3}.

Computer vision research on such visual geospatial data has primarily focused on improving recognition, and consequently, we see highly performing models for many of the aforementioned tasks. 
However, monitoring via recognition systems is only half of the story.
The goal of such planners and scientists is not just to monitor indicators but to intervene when necessary and make changes in accordance with policies.
For example, an urban planner might suggest that a particular street is perceived as unsafe due to the presence of scaffolding, and therefore, based on the city's policy, the scaffolding must be replaced with safer scaffolding.
While monitoring is well explored, identifying such recommendations remains underexplored.

\begin{figure}[tb]
  \centering
  \begin{minipage}[c]{0.5\linewidth}
    \centering
    \includegraphics[width=\linewidth]{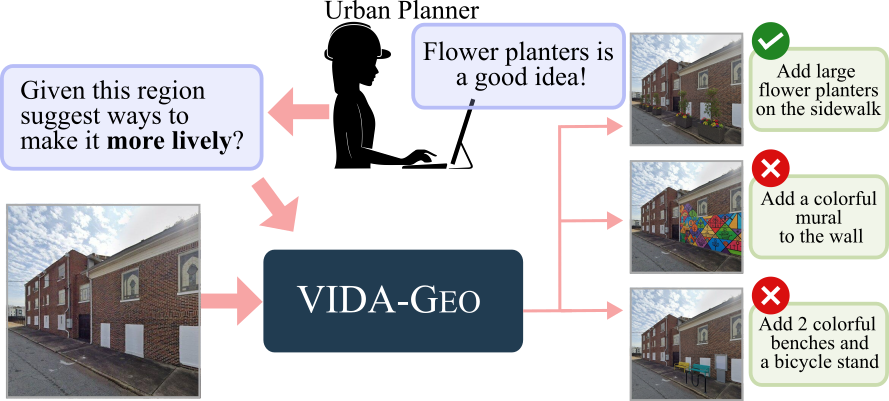}
  \end{minipage}
  \hfill
  \begin{minipage}[c]{0.49\linewidth}
  \small
    \caption{
    Given a geospatial image (street view or aerial), a user can ask \vida~to suggest interventions that improve a target indicator, such as liveliness.
    Our system proposes \textbf{multiple} realistic, policy-grounded interventions to support decision-making.
    The user can then choose the intervention most suitable for their use case.
    }
    \label{fig:teaser}
  \end{minipage}
  \vspace{-20pt}
\end{figure}

Moving beyond visual monitoring, we propose a new framework for \emph{visual intervention for decision support}. 
Given an input geospatial image of a region, such as an aerial or street view image, and a black-box indicator prediction model, such as road safety, our framework outputs realistic interventions to the image that improve the indicator score.
Our framework could aid policymakers in domains such as urban planning or climate science by providing a diverse set of interventions before implementing interventions in the real world (see \cref{fig:teaser}).

The current paradigm for finding interventions in urban planning is to run simulations in digital twins of a city~\citep{twin_1,twin_2}.
However, this has several challenges: 1) they require regular and expensive data collection to maintain the digital twin, 2) the verification of interventions requires expensive and time-consuming simulations, and 3) they could only model specific types of interventions depending on the simulation model.

We instead posit that, rather than creating expensive digital twins, \emph{a combination of an indicator monitoring model (recognition) and an image editing model (generative model) can serve as an implicit digital twin (IDT) for creating such interventions.}
Indicator recognition/monitoring models~\citep{safety_3,indicator_1} have been shown to be highly accurate.
Moreover, as we show in the results (see \cref{fig:counterfactuals}), they are also good at handling counterfactuals, which is an important requirement to be treated as a component of an implicit digital twin.
However, a prediction model alone cannot serve as an IDT, as it cannot generate interventions.
Modern generative models have both realistic generation capabilities and controllable editing~\citep{generation_1,generation_2}, including in domains such as satellite imagery~\citep{cher2025vectorsynth,sastry2024geosynth}.
Therefore, we use generative editing models to generate realistic interventions.

However, one key question remains: how do we propose realistic interventions that can actually be deployed in the real world, rather than simply optimizing the indicator score?
To ensure that interventions are grounded from a policymaker's perspective, we use large language models (LLMs) for intervention proposals.
Specifically, we introduce a multi-agent system with several sub-agents.
We start with a high-level \emph{planner} agent that, given an image, identifies regions of interest (ROIs) that can potentially be intervened upon.
The ROIs, along with the image, are passed to a \emph{policy sub-agent}, which uses real-world knowledge to suggest realistic interventions in the ROIs.
These ROIs are also passed to a \emph{segmentation agent}, which uses several foundation models to find relevant masks for intervention.
The \emph{generation sub-agent} then uses these prompts to generate realistic edits.
The edited images are evaluated both for improvement in the black-box indicator score and for whether they remain grounded in the goals of the \emph{policy agent}.
All trials are kept in memory and passed back to the \emph{planner agent}, which decides the next set of ROIs.
Framing this as a multi-agent system allows interventions to be regionally and policy-grounded while remaining modular and understandable to experts.
We call this a multi-agent system \textbf{\vida} (\textbf{V}isual \textbf{I}ntervention and \textbf{D}ecision-support \textbf{A}gent for \textbf{Geo}spatial data).

To evaluate our method and future works on this problem, we introduce a new benchmark with evaluation metrics for these 8 tasks.
We show that our agentic method can find realistic and grounded interventions by testing it on satellite and street view images across 8 indicator tasks.
Our main contributions are as follows.
\begin{packed_item}
    \item We introduce a new problem setup that moves beyond geospatial monitoring towards intervention for policy-based decision-making.
    \item We provide a new benchmark for evaluating this task, with evaluation metrics that measure the realism of interventions from both image-based and policy-oriented perspectives. 
    \item We introduce a multi-agent system \vida, that proposes, executes, and quantitatively evaluates geospatial interventions. We frame this system as an Implicit Digital Twin and show that it outperforms methods related to this problem setting.
\end{packed_item}

\section{Related Works}
\label{sec:related}

\noindent\textbf{{Urban Indicator Estimation}}
Overhead-images have been used for safety and mobility indicators, including crash-risk signals~\citep{roadsafety,trajectories,elallaf2026betarisk} and pedestrian-network extraction through sidewalks~\citep{walk_mapping,walk_3}.
Segmentation methods~\citep{openearthmap,deepglobe} have also been used to measure vegetation~\citep{treepedia,green_space}. 
Street-view imagery has been widely used to study perceived urban qualities such as safety and liveliness~\citep{placepulse,2024_global_streetscapes,safety_3}.
Recent work also uses it to model human perceptions~\citep{svi_percept,svi_ped}
and for multimodal LLM-based evaluation~\citep{svi_llm}.
While such models are well established, we use them to measure and guide visual interventions that improve an indicator score at a given place.

\noindent\textbf{{Intervention and Decision Making}}
Digital twins combine city data and simulations to support evidence-based
decision making~\citep{twin_2,twin_3}. 
However, such interventions are domain-specific and expensive to simulate.
Prior works use counterfactual models to understand the inputs' effect on the model's prediction~\citep{steex, octet}. Others use diffusion models to generate counterfactual explanations for black-box models~\citep{counterfactual,diff_usion,lance}. 
While these works aim to understand model behavior, our goal is instead to use them as real-world actions.

\noindent\textbf{{Agents for Intervention and Interpretability}} 
Remote-sensing VLMs extend instruction-following models to overhead imagery \citep{geochat,zhan2025skyeyegpt}, with recent work exploring agentic systems as well~\citep{rsagent,shabbir2025thinkgeo, kao2025univearth}. 
Other works show how agents can design experiments to explain model behavior and discover failure modes~\citep{maia,find}. 
Related works use generative models to simulate street-view interventions and changes in perceived urban quality~\citep{ursimulator,zhao2026street}.  
Unlike these works, \vida~formulates urban intervention as a multimodal decision problem. 
Moreover, the proposed actions are optimized not only for target urban indicators but also for physical realism and policy feasibility.


\section{Methodology}
\label{sec:methodology}


\vida\ is a multi-agent system that takes an image and an improvement goal instruction to produce grounded intervention proposals consisting of \emph{a region, an edit, and a textual description}. 
Instead of editing the whole image in one step, it decomposes the problem into specialized agents that reason about where to intervene, what to change, which changes are permissible, and how to generate plausible edits. 
This creates changes that are localized and atomic and therefore much easier to implement in the real world.
The agentic system includes a multi-epoch feedback loop, where summaries of prior epochs inform the subsequent planning and steering away from repeated failure modes.
 
\begin{figure}[t]
    \centering
    \includegraphics[width=\textwidth]{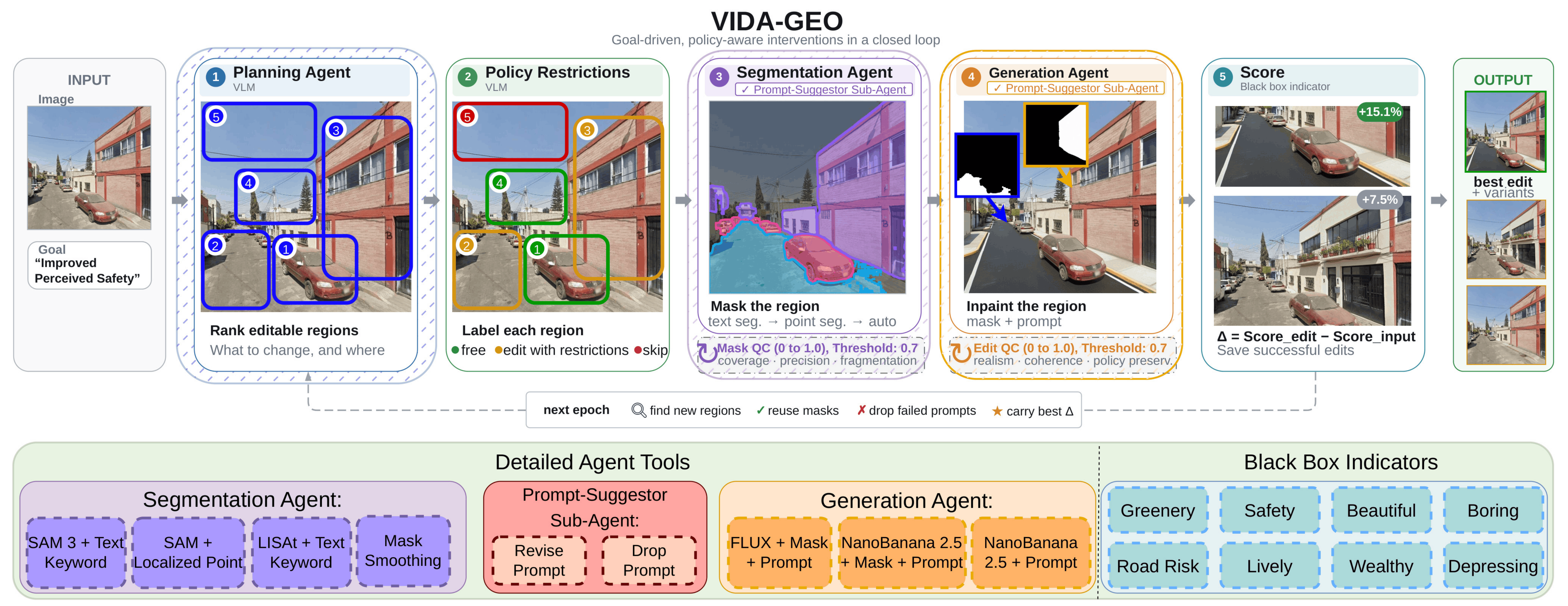}
    \vspace{-6pt}
    \caption{Overview of the \vida~ pipeline.  Given an image and a target urban-improvement instruction, a group of agents identifies ROIs, segments, and edits with task-specific constraints to produce candidate intervention edits. 
    Quality checks assess realism and policy preservation. 
    Candidates are finally evaluated by a black-box indicator scorer.}
    \label{fig:overview}
    \vspace{-10pt}
\end{figure}

The \vida\ pipeline (see \cref{fig:overview}) consists of four specialized agents followed by a scoring and multi-epoch orchestration. 
The \textbf{Planning Agent} decides where and what to change; the \textbf{Policy Agent} determines which changes are permissible and which regions are constrained; the \textbf{Segmentation Agent} identifies which pixels correspond to each region of interest; and the \textbf{Generation Agent} creates the visual interventions through prompted inpainting. 
Candidate interventions are evaluated by the black-box scorer, and the resulting outcomes are used by the \textbf{Orchestration} mechanism to guide later epochs. Each epoch corresponds to one complete planning-to-evaluation cycle. 
We next present these agents.

\subsection{Planning Agent}
Directly editing the entire image is highly unconstrained and may produce unrealistic or infeasible changes. 
Therefore, the \textbf{Planning Agent} first identifies and ranks multiple ROIs in the image as candidate intervention locations. 
For each ROI, the planner produces ranked spatial coordinates, bounding boxes, and semantic descriptions or keywords required for the downstream segmentation task in~\cref{sec:segmentation_agent}. 
In later epochs, the planner additionally receives a summary of previous attempts; we describe this feedback mechanism in~\cref{sec:multi-epoch}.



\subsection{Policy Agent}
\label{sec:policy_agent}
The \textbf{Policy Agent} then determines the physical and semantic constraints governing each ROI. 
These constraints fall into three categories: \textit{free} regions, where any modification is permissible; \textit{constrained} regions, where edits are allowed but must respect specific rules; and \textit{skip} regions, which contain critical infrastructure that must remain unaltered. 
These constraints determine how each ROI may be modified and are also passed to the Generation Agent in~\cref{sec:generation_agent}.
For street-view images, the policy agent also enforces general street-scene plausibility, ensuring that buildings, entrances, and other functional elements remain coherent in any edited output. An example of such policy violations can be seen in~\cref{fig:bad_policy}.

\subsection{Segmentation Agent}
\label{sec:segmentation_agent}
The \textbf{Segmentation Agent} receives each ROI from the planner and segments each region using one of several segmentation tools (\cref{fig:overview} (bottom)).
These include text-referred segmentation driven by the planner's suggested prompt and open-vocabulary models using the planner's keyword.
When both fail, the agent falls back to point-prompted segmentation models from the planner's estimated ROI center. 
Examples of segmentation failures can be seen in~\cref{fig:bad_masks} and the fallback for segmentation is provided in~\cref{sec:fallbacks}.
 
 
\noindent\textbf{{Segmentation QC Sub-Agent.}}
Because different segmentation methods can produce inaccurate, overly broad, or fragmented masks, each candidate mask is evaluated by an LLM-based quality-control (QC) step that compares the original image against the binary mask.
This evaluation considers three criteria.
\textit{Coverage} checks whether the mask spans the correct target area.
\textit{Precision} estimates how much unrelated regions are excluded.
\textit{Fragmentation} measures how spatially connected the mask is, penalizing, e.g., floating islands.

\noindent\textbf{{Feedback loop.}}
Failed masks and their QC descriptions are returned to the Segmentation Agent to adjust the mask selection strategy. 
The Prompt Suggestion sub-agent revises the segmentation keyword or prompt. 
This inner closed-loop mechanism improves poor-quality masks before presenting them to downstream intervention stages.

\subsection{Generation Agent}
\label{sec:generation_agent}
The \textbf{Generation Agent} receives the input image and all accepted masks to produce edited images intended to improve the black-box indicator. 
The images are edited through a mask-and-text-prompted diffusion process, with the \textit{Policy Agent}'s constraint annotations injected into the text prompt. 
These constraints ensure that edits remain consistent with the physical and functional structure of the scene.
When mask-based inpainting is not feasible, the agent falls back on mask-free text-guided generation models (see examples in~\cref{sec:fallbacks}). 

\noindent\textbf{{Generation QC Sub-Agent.}}
Because edits may contain unrealistic, inconsistent, or semantically incorrect modifications, the images undergo a QC step that evaluates three properties.
\textit{Realism} checks whether the edited region can plausibly exist in the real world, flagging, e.g., unnatural textures.
\textit{Coherence} verifies that the edit is contextually consistent with the rest of the image.
Finally, \textit{prompt adherence} confirms that the generation agent followed the suggested editing prompt. The complete prompts are present in~\cref{sec:prompts}.

\noindent\textbf{{Policy compliance check.}}
As previously mentioned in~\cref{sec:policy_agent}, each ROI is associated with policy constraints, verified using the Generation QC to ensure the physical coherence of edits.
For example, 
a residential cluster replacing a retail parking lot is implausible policywise.
Violations are fed back to the generation agent so that it can suggest better-aligned prompts.
In this work, the regulations come from the LLM's knowledge; however, \emph{a future direction would be to improve policy compliance by incorporating city regulations rulebooks}.

\noindent\textbf{{Feedback loop.}}
Failed outputs, including examples shown in~\cref{fig:bad_edits}, and their corresponding QC feedback are returned to the Generation Agent. The agent analyzes whether the edits lacked realism, broke coherence, or deviated from the prompt. The Prompt Suggestion sub-agent then produces a revised editing prompt for the next generation attempt. 


\subsection{Scoring and Multi-Epoch Orchestration}
\label{sec:multi-epoch}
Finally, for each ROI, multiple edited images may pass quality control. 
All accepted candidates are scored using the black-box indicator and compared against the original image. 
Candidates are ranked based on the difference between the two scores (delta score).
This process is repeated over multiple cycles, with the results from each cycle passed back to the planning agent. 
To encourage the agents to identify new regions, produce new masks, and generate alternative edits, all previous candidates and their corresponding delta scores are retained and carried forward to the subsequent cycle.

This accumulated history discourages the system from repeating unsuccessful strategies while encouraging the exploration of diverse editing directions. As a result, this approach yields multiple edited variants from a single input image, each of which adheres to the policy constraints while improving the black-box indicator score.



\section{Results}
\label{sec:results}
In this section, we discuss the implementation details of our agent, present the benchmark and baselines, and show that our multi-agent system provides realistic and grounded interventions. 
The implementation details are presented in the appendix (\cref{sec:setup}).
\subsection{Benchmark}
\label{sec:dataset}
To perform a comprehensive evaluation with different types of visual geospatial modalities, we sample both aerial images and Street View images.
More specifically, we source high-resolution MapBox aerial images and Google Street View images to evaluate interventions.

For intervention and decision-making, we use black-box indicator prediction models. 
For aerial images, we select two models: First is BetaRisk model~\citep{elallaf2026betarisk}, trained to predict fatal-crash risk values. 
The second is OpenEarthMap~\citep{openearthmap} used for greenery prediction. 
For Street View images, we use the models released with Global Streetscapes~\citep{2024_global_streetscapes}, which were pretrained on MIT Place Pulse 2.0~\citep{placepulse}, and predict six standard urban perception attributes: safety, lively, beautiful, wealthy, less boring, and less depressing, on a 0--10 scale which we rescale to 0--100. 

Because the evaluation requires repeated multi-agent and generative model interactions, we prioritize careful assessment over dataset size.
For each of the 8 indicators,
we score randomly sampled images with the corresponding black-box model and filter
out those already scoring above $90$.
From the remaining pool, a VLM selects 100 images per indicator, ensuring diversity in score, geography, and  appearance. See \cref{app:data_statistics} for benchmark statistics.

\subsection{Evaluation Metrics}
\label{sec:evaluation_metrics}
Evaluating interventions is a challenging problem, as \vida\ and baselines can generate a wide range of possible interventions, and it is not directly observable whether these interventions would be effective in the real world.
Therefore, we evaluate such interventions across two complementary axes.
First, the \emph{\textbf{perceptual quality}} of the generated interventions; by determining whether they appear perceptually realistic.
Second, the \emph{\textbf{policy alignment}}; by determining whether the interventions are realistic and actionable from a policy perspective.

\noindent\textbf{{FID-Proxy Through Mahalanobis Distance}} is used to assess if generated edits remain within the distribution of real images~\citep{mahalanobis2018generalized}.
FID requires a set of generated images; however, since we have only one image at a time, we use a proxy (see \cref{sec:fid}).

\noindent\textbf{{LLM-as-a-Judge Evaluation}.}
Since human evaluation at scale is expensive, we primarily rely on LLM judges~\citep{zheng2023judging}, using GPT 5.2 and Qwen 3VL~\citep{singh2025openai,bai2025qwen3} (results in~\cref{sec:llm_judge}). 
Each edited image is scored independently (more information in~\cref{sec:prompts}) on a discrete 1-10 (later scaled to 1-100) scale across three criteria: 
\begin{packed_item}
    \item ~\textbf{Visual quality}: The technical quality of the edit and how natural the image looks. This metric also measures the \emph{perceptual quality}. 
    \item ~\textbf{Realism}: Assess whether the edited world is physically and semantically plausible. 
    Objects should have plausible function, material, scale, geometry, and spatial relationships.   
    \item ~\textbf{Policy preservation}: How well the edit preserves the original scene’s structure, layout, function, and geographic identity, while limiting changes to the intended local intervention.
\end{packed_item}
A composite judge score is computed as the mean of these 3 LLM-as-a-judge scores.
The latter two metrics assess whether the intervention is plausible and \emph{policy-aligned}.

\noindent\textbf{Human Expert Evaluation.}
We additionally evaluate a subset of our data with seven graduate students and five experts recruited through Prolific~\citep{prolific2026}, specializing in urban planning and policy design (see~\cref{sec:evaluators_background}). Evaluators make pairwise comparisons between \vida\ and a selected baseline, providing human validation of perceptual realism and policy compatibility beyond LLM-based evaluation.
We also report high agreement between LLM and expert decisions, suggesting that LLM judges can serve as a scalable and reproducible proxy for human judgments (see~\cref{sec:evaluators_results}).


\subsection{Baselines}
Since we propose a new problem of discovering interventions, there are not many relevant strong baselines. 
The most closely related method is URSimulator~\citep{ursimulator}, which performs perception-guided iterative editing of street-view images with a diffusion model. 
URSimulator tunes text prompts according to predicted human-perception scores but does not adhere to realism and policy. 
We additionally adapt methods that generate counterfactuals for interpretability to compare with \vida. 
DIFFusion~\citep{diff_usion} interprets differences between 2 classes by using a diffusion model to transform an image's class.  
It is a differentiable method that edits via gradient optimization instead of a grounded discrete search like \vida. 
LANCE~\citep{lance} utilizes discrete search to find good counterfactuals for a vision classifier.
We use it to find edits maximizing indicator scores.
We also compare our method against NanoBanana 2.5 in the zero-shot setting (NB2.5 (ZS)), where we directly prompt it to edit an image to improve the black-box score.
For fair comparison, all models get the same number of calls to generative models.

\begin{table*}[t]
    \centering
    \caption{Task performance, perceptual quality, and policy alignment for edited images judged with \textbf{GPT5.2}. 
    \textcolor{mygreen}{Green} and \textcolor{myred}{red} tasks favor higher and lower task scores respectively. 
    Delta Avg. is signed toward the task objective. 
    FID-Proxy and judge-score columns report average \emph{95\% confidence-intervals}. The best judge score for each task and evaluation is boldfaced.}
    \label{tab:main_results_gpt52}
    \small
    \setlength{\tabcolsep}{5pt}
    \resizebox{\textwidth}{!}{%
    \begin{tabular}{llccccccc}
        \toprule
        \textbf{Task} & & &
        & \multicolumn{2}{c}{\textbf{Perceptual Quality}}
        & \multicolumn{2}{c}{\textbf{Policy Alignment}}
        & \\
        \cmidrule(lr){5-6}\cmidrule(lr){7-8}
        & \textbf{Method}
        & \makecell{\textbf{Output}\\\textbf{Avg.} (\%)}
        & \makecell{\textbf{Delta}\\\textbf{Avg.} (\%)}
        & \makecell{\textbf{FID-Proxy}\\($\downarrow$)\\{\scriptsize $\pm$ 1.00}}
        & \makecell{\textbf{Visual}\\\textbf{Quality} (\%)\\{\scriptsize $\pm$ 0.70}}
        & \makecell{\textbf{Realism} (\%)\\{\scriptsize $\pm$ 0.87}}
        & \makecell{\textbf{Policy}\\\textbf{Pres.} (\%)\\{\scriptsize $\pm$ 1.25}}
        & \makecell{\textbf{LLM Judge}\\\textbf{Avg.} (\%)\\{\scriptsize $\pm$ 0.76}} \\
        \midrule

        \multirow{4}{*}{\textcolor{mygreen}{\textbf{Greenery}}}
        & \textbf{\vida} & 35.6 & 6.3 & 44.8 & 72.2 & 67.2 & 64.1 & 67.8 \\
        & LANCE & 25.3 & -4.1 & 57.3 & 37.5 & 28.6 & 16.9 & 27.7 \\
        & DIFFusion & 59.5 & 30.1 & 59.6 & 26.1 & 25.5 & 26.0 & 25.9 \\
        & NB2.5 (ZS) & 43.0 & 13.6 & 44.7 & 71.9 & 66.5 & 66.1 & 68.2 \\

        \midrule

        \multirow{4}{*}{\textcolor{myred}{\textbf{Road Risk}}}
        & \textbf{\vida} & 82.1 & 4.0 & \textbf{47.5} & \textbf{72.8} & \textbf{67.4} & 57.8 & \textbf{66.0} \\
        & LANCE & 68.7 & 17.4 & 58.2 & 38.6 & 31.5 & 19.0 & 29.7 \\
        & DIFFusion & 13.3 & 72.8 & 60.2 & 24.6 & 26.8 & 26.0 & 25.8 \\
        & NB2.5 (ZS) & 81.6 & 4.5 & 49.8 & 68.1 & 61.9 & 57.5 & 62.5 \\

        \specialrule{1.2pt}{0.8ex}{0.8ex}

        \multirow{5}{*}{\textcolor{mygreen}{\textbf{Safety}}}
        & \textbf{\vida} & 20.9 & 3.8 & 56.5 & \textbf{66.5} & \textbf{73.5} & \textbf{70.3} & \textbf{70.1} \\
        & LANCE & 33.8 & 16.6 & 61.0 & 58.5 & 55.8 & 32.8 & 49.0 \\
        & DIFFusion & 50.5 & 33.4 & 70.3 & 28.6 & 35.0 & 32.1 & 31.9 \\
        & URSimulator & 33.0 & 15.9 & 57.4 & 57.9 & 50.8 & 40.0 & 49.6 \\
        & NB2.5 (ZS) & 41.3 & 24.2 & 61.4 & 62.3 & 61.8 & 40.1 & 54.7 \\

        \midrule

        \multirow{5}{*}{\textcolor{mygreen}{\textbf{Lively}}}
        & \textbf{\vida} & 25.1 & 11.4 & 61.5 & \textbf{61.4} & \textbf{63.6} & \textbf{65.6} & \textbf{63.5} \\
        & LANCE & 23.0 & 9.4 & 62.6 & 59.4 & 53.9 & 29.3 & 47.6 \\
        & DIFFusion & 39.5 & 25.8 & 66.2 & 34.9 & 42.2 & 36.5 & 37.8 \\
        & URSimulator & 29.6 & 16.0 & \textbf{58.8} & 52.7 & 44.7 & 35.4 & 44.3 \\
        & NB2.5 (ZS) & 52.0 & 38.3 & 68.2 & 54.2 & 53.0 & 46.8 & 51.3 \\

        \midrule

        \multirow{5}{*}{\textcolor{mygreen}{\textbf{Beautiful}}}
        & \textbf{\vida} & 26.1 & 8.4 & 54.5 & \textbf{65.7} & \textbf{74.7} & \textbf{72.4} & \textbf{70.9} \\
        & LANCE & 42.1 & 24.4 & 58.2 & 56.4 & 52.3 & 29.9 & 46.2 \\
        & DIFFusion & 53.0 & 35.3 & 71.8 & 25.9 & 29.9 & 27.6 & 27.8 \\
        & URSimulator & 32.8 & 15.1 & 54.9 & 56.9 & 52.5 & 41.3 & 50.2 \\
        & NB2.5 (ZS) & 43.7 & 25.9 & 61.9 & 53.8 & 66.3 & 53.9 & 58.0 \\

        \midrule

        \multirow{5}{*}{\textcolor{mygreen}{\textbf{Wealthy}}}
        & \textbf{\vida} & 27.0 & 9.1 & 58.0 & \textbf{65.6} & \textbf{70.8} & \textbf{62.8} & \textbf{66.4} \\
        & LANCE & 35.4 & 17.4 & 61.0 & 58.9 & 55.2 & 31.0 & 48.4 \\
        & DIFFusion & 54.9 & 37.0 & 69.8 & 31.0 & 38.7 & 35.7 & 35.1 \\
        & URSimulator & 35.5 & 17.5 & 57.9 & 56.3 & 47.9 & 35.9 & 46.7 \\
        & NB2.5 (ZS) & 68.8 & 50.8 & 62.5 & 59.8 & 38.4 & 15.8 & 38.0 \\

        \midrule

        \multirow{5}{*}{\textcolor{myred}{\textbf{Boring}}}
        & \textbf{\vida} & 74.6 & 6.0 & 58.2 & \textbf{65.7} & \textbf{72.2} & \textbf{68.1} & \textbf{68.7} \\
        & LANCE & 62.8 & 17.8 & 59.5 & 59.8 & 54.3 & 30.1 & 48.1 \\
        & DIFFusion & 55.2 & 25.4 & 62.7 & 35.1 & 42.2 & 37.4 & 38.2 \\
        & URSimulator & 64.4 & 16.2 & 60.5 & 53.1 & 43.2 & 33.4 & 43.2 \\
        & NB2.5 (ZS) & 61.1 & 19.4 & 74.6 & 42.8 & 35.8 & 31.9 & 36.9 \\

        \midrule

        \multirow{5}{*}{\textcolor{myred}{\textbf{Depressing}}}
        & \textbf{\vida} & 75.5 & 4.6 & 57.6 & \textbf{65.6} & \textbf{73.5} & \textbf{70.2} & \textbf{69.7} \\
        & LANCE & 58.9 & 21.2 & 61.0 & 57.3 & 51.6 & 28.1 & 45.7 \\
        & DIFFusion & 52.3 & 27.8 & 69.8 & 30.2 & 37.7 & 33.2 & 33.7 \\
        & URSimulator & 71.2 & 8.9 & 57.9 & 54.9 & 48.8 & 38.3 & 47.3 \\
        & NB2.5 (ZS) & 57.9 & 22.2 & 63.4 & 60.1 & 61.3 & 46.2 & 55.9 \\

        \bottomrule
    \end{tabular}%
    }
\end{table*}

\subsection{Results and Discussion}
We first test \vida\ quantitatively against the aforementioned baselines.
We observe that it achieves the highest overall performance on both performance axes.
Additionally, we qualitatively show that \vida\ produces multiple plausible interventions that increase the indicator scores and can be used to provide multiple interventions sequentially.

\noindent\textbf{{Can \vida\ produce good interventions?}}
We compare \vida\ against other baselines on all 8 tasks in \cref{tab:main_results_gpt52}. 
\vida\ consistently obtains the highest LLM-judge scores on 7 tasks, often scoring 15pp higher than the other models, indicating that its edits are judged to be more policy-aligned and visually coherent compared to the baselines.

In several tasks, DIFFusion produces the largest changes, such as reducing road risk by 72.8\%, but these edits receive much lower perceptual-quality and policy-alignment scores. 
As for LANCE, it may change the target indicator in the unintended direction and receive lower LLM-judge scores than \vida, suggesting that its interventions are less actionable. The overall results suggest that \vida\ can produce good interventions in terms of perceptual quality and can adhere much better to real-world intervention policies.

URSimulator only applies to street view images, where it tends to be more conservative in making edits; moreover, it shows a noticeable gap in policy alignment compared to \vida.
The zero-shot baseline NB2.5 (ZS) which, unlike \vida, is not explicitly designed to search for interventions that satisfy real-world policy constraints. 
Nevertheless, it achieves a larger task-directed change than \vida. 
For example, \textit{Lively} and \textit{Wealthy} scores are improved by 39.1\% and 50.8\%, respectively, compared to 12.7\% and 10.2\% for \vida. 
\emph{However, these larger changes do not consistently translate into high-quality or realistic interventions.}
The exception is the \textit{Greenery} task, where NB2.5 (ZS) performs comparably to \vida\ in perceptual and policy-alignment quality. 
However, this task is relatively simple, as increasing greenery can often be achieved through simple visual edits such as adding vegetation in empty spaces without requiring complex reasoning.

\begin{figure}[t]
  \centering
  \includegraphics[width=0.9\linewidth]{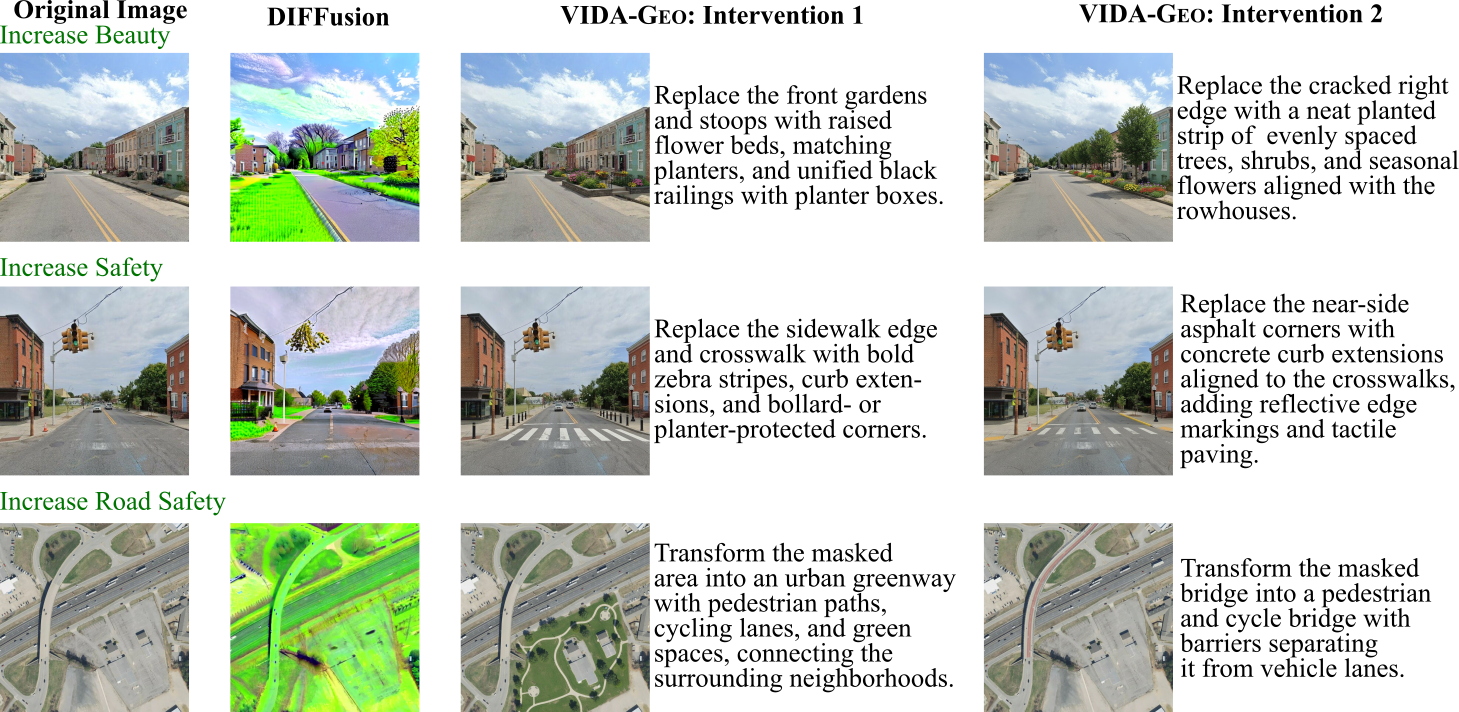}
  \vspace{-6pt}
  \caption{Comparison of intervention outputs from \vida\ vs. DIFFusion for street view (first 2 rows) and aerial inputs (last row). 
  The interventions provided by \vida~are more practical and realizable in the real world, e.g., adding bollards and tactile pavements around the intersection in the second.
  Moreover, \vida\  can provide text instructions along with the images, unlike DIFFusion, which is more useful to a user.}
  \label{fig:qualitative}
  \vspace{-8pt}
\end{figure}

\noindent\textbf{{Do \vida\  interventions make sense to experts?}}
We also asked domain experts to pairwise compare the \vida\ against DIFFusion and LANCE for the same inputs.
The evaluators were asked, "Which image better reflects the target attribute, which appears more realistic, and which edit would be more feasible in urban planning?". 
The results reveal a discrepancy between black-box models and human evaluation. 
Against DIFFusion, experts prefer our agent's outputs 61.3\% of the time on perceptual attribute alignment, compared to only 10\% of the black-box models favoring \vida. 
The same trends held against LANCE, where experts preferred \vida\ 86.7\% of the time on perceptual attributes and 79.3\% of the time on feasibility. 
Overall, \vida\ achieves the strongest overall human-aligned performance (See~\cref{sec:huamn_eval_analysis} for more results). 

While experts disagree with the black-box model, they have a high level of agreement with LLM judges, averaging 88.5\% on the \emph{realism} and 91.6\% on the \emph{policy-preservation} metric (random baselines is 50\%). 
This indicates that LLMs provide evaluations closely aligned with expert judges and can serve as a scalable proxy for more expensive human evaluations.

Finally, \vida's interventions can resemble real-world changes. Such changes increase black-box beauty and safety scores by an average of 24.3\% and 12.46\%, respectively, through interventions such as extended curb geometry, tactile surfaces, and added vegetation. Similar interventions emerge from \vida\ (see~\cref{sec:real_edits}).

\noindent\textbf{{How do interventions look qualitatively?}}
\Cref{fig:qualitative} presents example interventions for 3 tasks.
In most cases, the interventions suggested by DIFFusion are harder to interpret, perceptually unrealistic, and not policy-realizable.
In the second row, it removes the traffic light, making the region less safe. 
In the aerial image, it mainly increases the saturation of the green channel, which is both policy-wise impractical and perceptually unrealistic. 

\vida, on the other hand, provides multiple interventions (along with their text descriptions) that look realistic and are easier to implement. 
For aerial images, it suggests replacing large parking lots with greener parks, replacing a road bridge with a cycling bridge, and making roads safer, especially for pedestrians (refer to~\cref{sec:more_examples} for more examples).

Since aerial images represent a larger physical area, even small interventions, such as replacing the bridge, would be a much bigger undertaking for an urban planner. 
This explains the relatively lower scores for aerial image tasks in \cref{tab:main_results_gpt52} compared to street-view tasks.

\begin{figure}[t]
  \centering
  \begin{subfigure}[c]{0.68\linewidth}
    \centering
    \includegraphics[width=\linewidth]{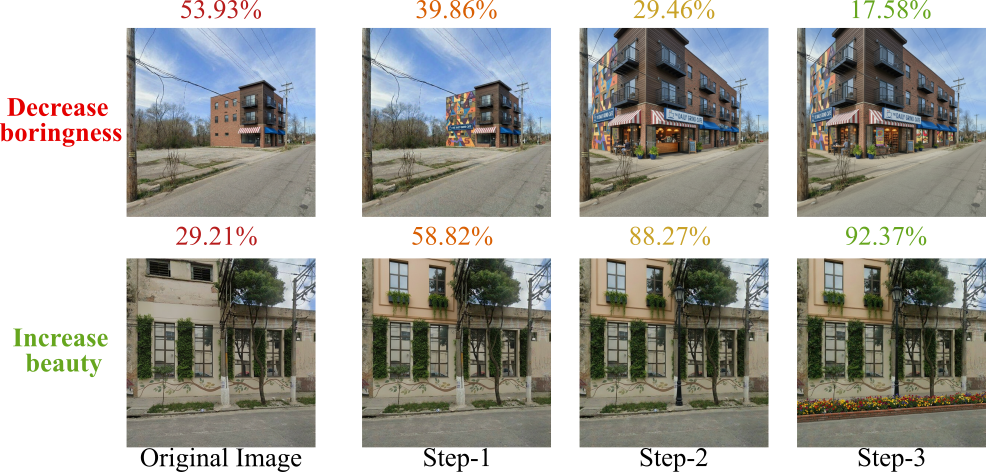}
    \caption{}
    \label{fig:sequential_examples}
  \end{subfigure}
  \hfill
  \begin{subfigure}[c]{0.30\linewidth}
    \centering
    \includegraphics[width=\linewidth]{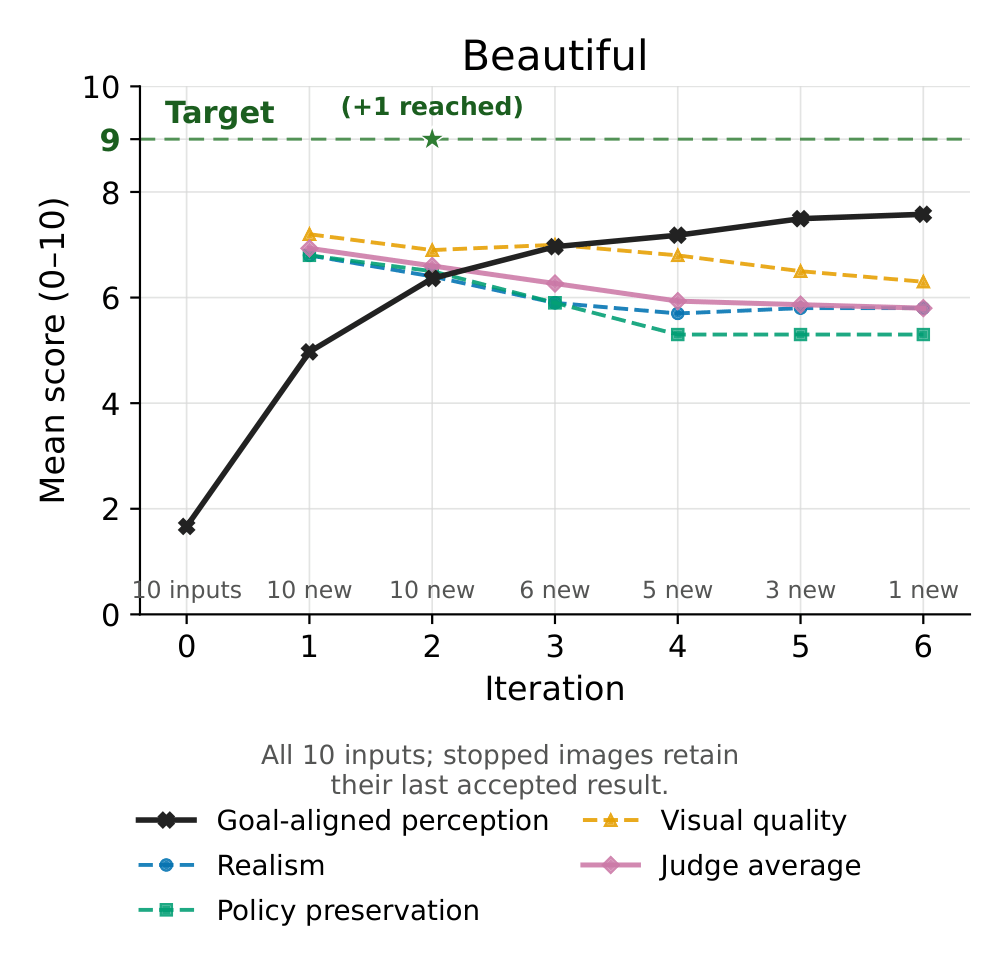}
    \caption{}
    \label{fig:sequential_scores}
  \end{subfigure}

  \vspace{-5pt}
  \caption{Using \vida\ with multiple sequential edits.
  (a) It suggests multiple edits in different regions
  and increases the indicator score. For example, in the
  bottom row, it improves the building facade, then adds a lamppost, followed
  by roadside flowers, increasing the beauty score from 29.2\% to 92.4\%.
  (b)~Quantitative progression over successive iterations.}
  \label{fig:sequential}
  \vspace{-8pt}
\end{figure}

\noindent\textbf{{Can \vida~do sequential edits?}}
In \cref{fig:sequential}, we show the results of applying \vida\ sequentially to outputs from a prior run.
These sequential edits produce interventions that align with policy and increase the indicator score more than a single edit.
For example, in the top row, \vida\ first adds a mural, then expands the buildings, and finally adds texture to the window.
The boring score decreases from 53\% to 17\%, while with single-edit we get the score to 40\% at best.
This suggests that when urban planners seek larger changes, \vida\ can propose sequential edits that progressively improve the target indicator while largely preserving quality and policy alignment. 
As shown in~\cref{fig:sequential_scores}, goal-aligned scores increase with iterations, while LLM-judges metric decrease slightly, suggesting a reasonable trade-off. Additional sequential analysis can be found in \cref{app:sequential}.

\paragraph{Ablations:}
Full ablations of \vida\ can be found in \cref{sec:ablation}.
We show that policy constraints are the largest contributor to policy-aligned outputs. Edit QC and Mask QC also provide substantial gains, and the suggestor offers smaller but consistent improvements.







\section{Conclusion, Limitations, and Future Work}
\label{sec:conclusion}
\paragraph{Conclusion:} We presented a new problem setup moving beyond monitoring towards testing interventions and decision-making.
Testing and evaluating successful intervention is a challenge; therefore, we proposed an evaluation benchmark of 8 different tasks across aerial and street view images, consisting of 800 images.
We also proposed new metrics to evaluate the quality of such interventions, both perceptually and policywise.
To tackle the problem, we introduced \vida~ a multi-agentic method that suggests good grounded interventions.
We believe such intervention can be used as decision support to real-life urban planners and policymakers when deciding to change a physical space in the real world.

\noindent\textbf{{Limitations and Future Work:}}
Since our work introduces the new problem of visual intervention for geospatial support, we present the first method as a solution.
We believe there is substantial room for improvement.
For example, in the current system, edits generated by FLUX, especially for aerial images, frequently fail quality checks, leading to over-reliance on NanoBanana.
Additionally, replacing closed-source models as agents with open-source counterparts is possible but degrades quality significantly.
Furthermore, the current method relies on multiple mask/edit quality checks to ensure good outputs, which could be reduced with better segmentation and editing tools.
Finally, this work does not directly encode full zoning law, engineering standards, or construction budgets. 
However, it is intended as an early-stage decision support for urban planners and policymakers, rather than an implementation system. 
The generated edits and their indicator scores provide visual proxies that can help experts explore and compare candidate interventions, and we hope that future work will include such constraints to further improve the intervention framework. 
\FloatBarrier

\bibliographystyle{iclr2027_conference}
\bibliography{main}
\newpage
\appendix

\etocdepthtag.toc{appendix}

\etocsettagdepth{main}{none}
\etocsettagdepth{appendix}{section}

\etocsettocstyle
  {\section*{Appendix Contents}}
  {}

\tableofcontents

    
    
    

\newpage
\section{\vida\ Algorithm and Fallbacks}
\label{sec:fallbacks}
\SetKwFunction{Planner}{Planner}
\SetKwFunction{PolicyAgent}{PolicyAgent}
\SetKwFunction{SegAgent}{SegmentationAgent}
\SetKwFunction{GenAgent}{GenerationAgent}
\SetKwFunction{UpdateHistory}{UpdateHistory}

\SetKwProg{Fn}{Function}{:}{}

\begin{algorithm}[h]
\caption{\vida\ Agent Procedures}
\label{alg:vida_agents}

\Fn{\Planner{$I,G,\mathcal{H}$}}{
    $\mathcal{R} \leftarrow \mathrm{RankROIs}(I,G,\mathcal{H})$
    \tcp*{identify and rank candidate ROIs}

    \Return{$\mathcal{R}$}\;
}

\BlankLine

\Fn{\PolicyAgent{$I,r,G$}}{
    $C_r \leftarrow \mathrm{AssignConstraint}(I,r,G)$
    \tcp*{free, constrained, or skip}

    \Return{$C_r$}\;
}

\BlankLine

\Fn{\SegAgent{$I,r$}}{

    $M \leftarrow \mathrm{Segment}(I,r)$
    \tcp*{select SAM, SAM3, or LISA}

    $(q_m,f_m) \leftarrow \mathrm{MaskQC}(I,M,r)$\;

    \If{$f_m$ indicates fragmentation}{
        $M \leftarrow \mathrm{SmoothMask}(M)$
        \tcp*{smooth fragmented regions}

        $(q_m,f_m) \leftarrow \mathrm{MaskQC}(I,M,r)$\;
    }

    \While{$q_m = \mathrm{fail}$ and retries remain}{

        $P_s \leftarrow \mathrm{PromptSuggestor}(r,f_m)$
        \tcp*{revise segmentation prompt}

        $M \leftarrow \mathrm{Segment}(I,r,P_s)$
        \tcp*{revise prompt and/or switch segmenter}

        $(q_m,f_m) \leftarrow \mathrm{MaskQC}(I,M,r)$\;
    }

    \Return{$M,q_m$}\;
}

\BlankLine

\Fn{\GenAgent{$I,M,r,G,C_r$}}{

    $P_g \leftarrow \mathrm{PromptSuggestor}(r,G,C_r)$
    \tcp*{propose editing prompt}

    \Repeat{$q_g = \mathrm{pass}$ or retries exhausted}{

        $I' \leftarrow \mathrm{Generate}(I,M,P_g)$
        \tcp*{select available generation tool}

        $(q_g,f_g) \leftarrow
        \mathrm{GenerationQC}(I,I',M,P_g,C_r)$\;

        \If{$q_g = \mathrm{fail}$}{
            $P_g \leftarrow \mathrm{PromptSuggestor}(P_g,f_g)$
            \tcp*{revise editing prompt}
        }
    }

    \Return{$I',q_g$}\;
}

\end{algorithm}

\begin{algorithm}[h]
\caption{\vida\ Multi-Agent Geo-Editing Pipeline}
\label{alg:vida}

\KwIn{Image $I$, goal $G$, scorer $S$}
\KwOut{Accepted interventions $\mathcal{O}$}

$\mathcal{H} \leftarrow \emptyset$,
$\mathcal{O} \leftarrow \emptyset$\;

$s_0 \leftarrow S(I)$\;

\For{$e = 1,\dots,E$}{

    $\mathcal{R} \leftarrow
    \Planner(I,G,\mathcal{H})$\;

    \ForEach{$r \in \mathcal{R}$}{

        $C_r \leftarrow
        \PolicyAgent(I,r,G)$\;

        \If{$C_r = \mathrm{skip}$}{
            \textbf{continue}\;
        }

        $(M,q_m) \leftarrow
        \SegAgent(I,r)$\;

        \If{$q_m = \mathrm{fail}$}{
            \textbf{continue}\;
        }

        $(I',q_g) \leftarrow
        \GenAgent(I,M,r,G,C_r)$\;

        \If{$q_g = \mathrm{pass}$}{

            $\Delta \leftarrow S(I') - s_0$\;

            $\mathcal{O}
            \leftarrow
            \mathcal{O}
            \cup
            \{(r,M,I',\Delta)\}$\;
        }
    }

    $\mathcal{H}
    \leftarrow
    \UpdateHistory(\mathcal{H},\mathcal{O})$
    \tcp*{inform the next epoch}
}

\Return{$\mathcal{O}$}\;

\end{algorithm}

\begin{figure}[h!]
    \centering

    \begin{subfigure}[t]{0.31\textwidth}
        \centering
        \includegraphics[
            width=\linewidth,
            height=0.22\textheight
        ]{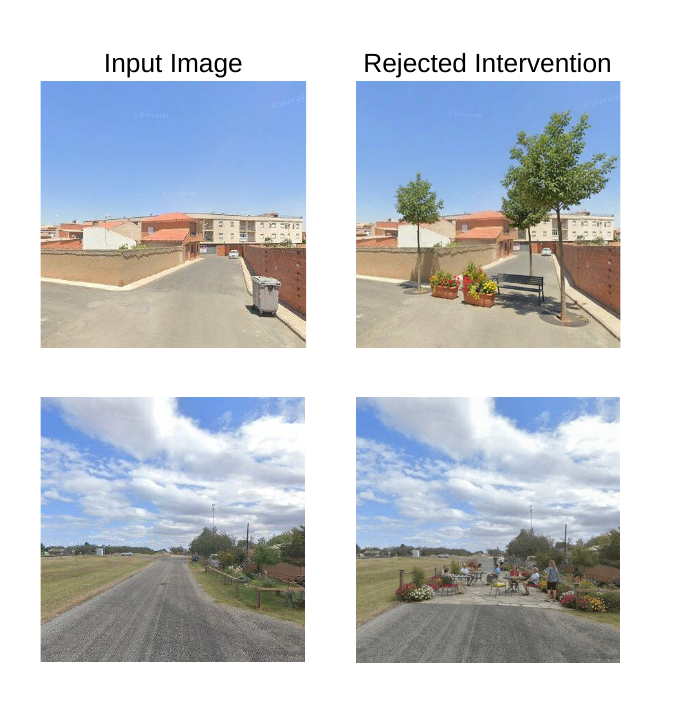}
        \caption{Policy violations.}
        \label{fig:bad_policy}
    \end{subfigure}
    \hspace{0.02\textwidth}
    \begin{subfigure}[t]{0.31\textwidth}
        \centering
        \includegraphics[
            width=\linewidth,
            height=0.22\textheight
        ]{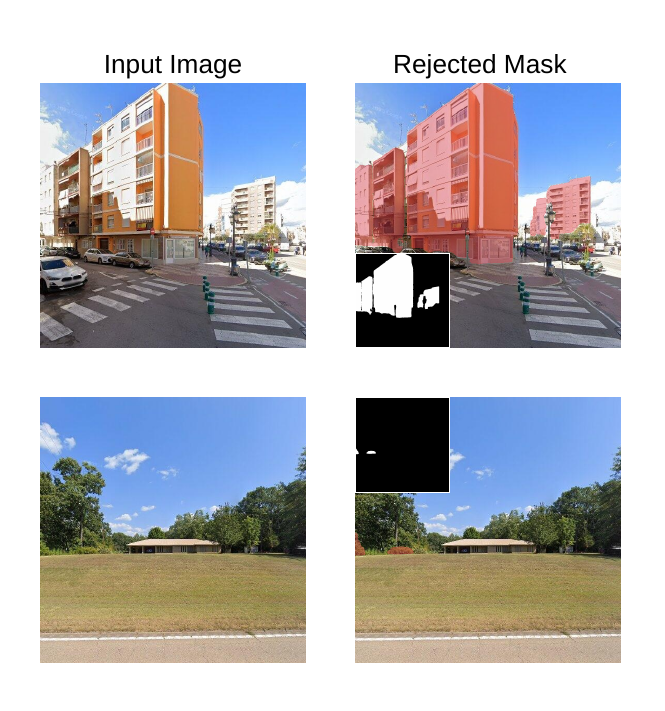}
        \caption{Bad masks.}
        \label{fig:bad_masks}
    \end{subfigure}
    \hspace{0.02\textwidth}
    \begin{subfigure}[t]{0.31\textwidth}
        \centering
        \includegraphics[
            width=\linewidth,
            height=0.22\textheight
        ]{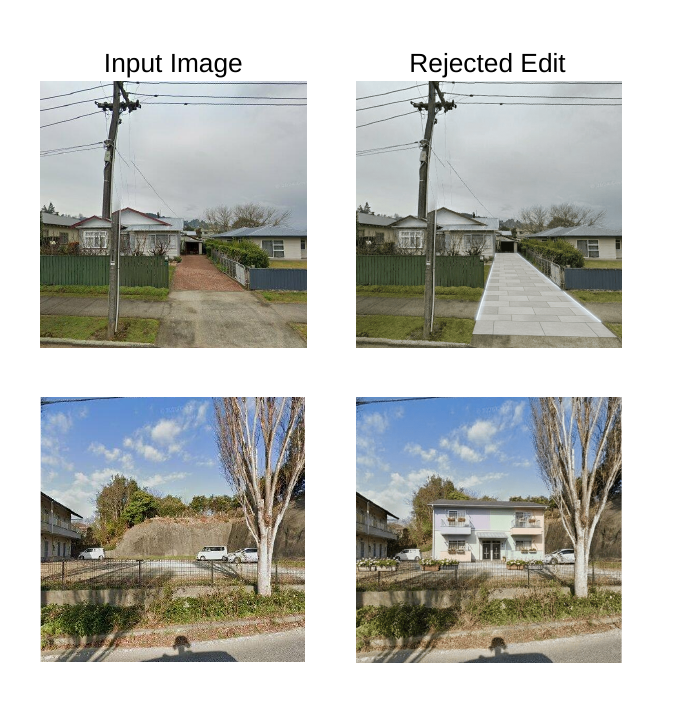}
        \caption{Bad edits.}
        \label{fig:bad_edits}
    \end{subfigure}

    \caption{Examples of failure cases. (a) shows policy violations that our system steers away from. (b) shows segmentation masks rejected by the QC sub-agent, with the top image segmenting "medium rise building" and the bottom "grass shrubs". Whereas subfigure (c) illustrates rejected edits, the top edit includes a low-res and unrealistic edit, and the bottom edit inpainting the wrong region. Fallbacks and fixes are explained in~\cref{sec:fallbacks}.}
    \label{fig:failure_examples}
\end{figure}

The goal of \vida\ is to avoid full-image-level interventions that could lead to artifacts and policy violations shown in~\cref{fig:bad_policy}. The algorithm shown in~\cref{alg:vida} summarizes the overall flow of the \vida\ pipeline that operates on the region-level; including the interaction between planning, policy assignment, segmentation, generation, quality control, and multi-epoch feedback.  

\begin{figure*}[h]
    \centering

    \begin{subfigure}{\textwidth}
        \centering
        \includegraphics[width=0.80\textwidth]{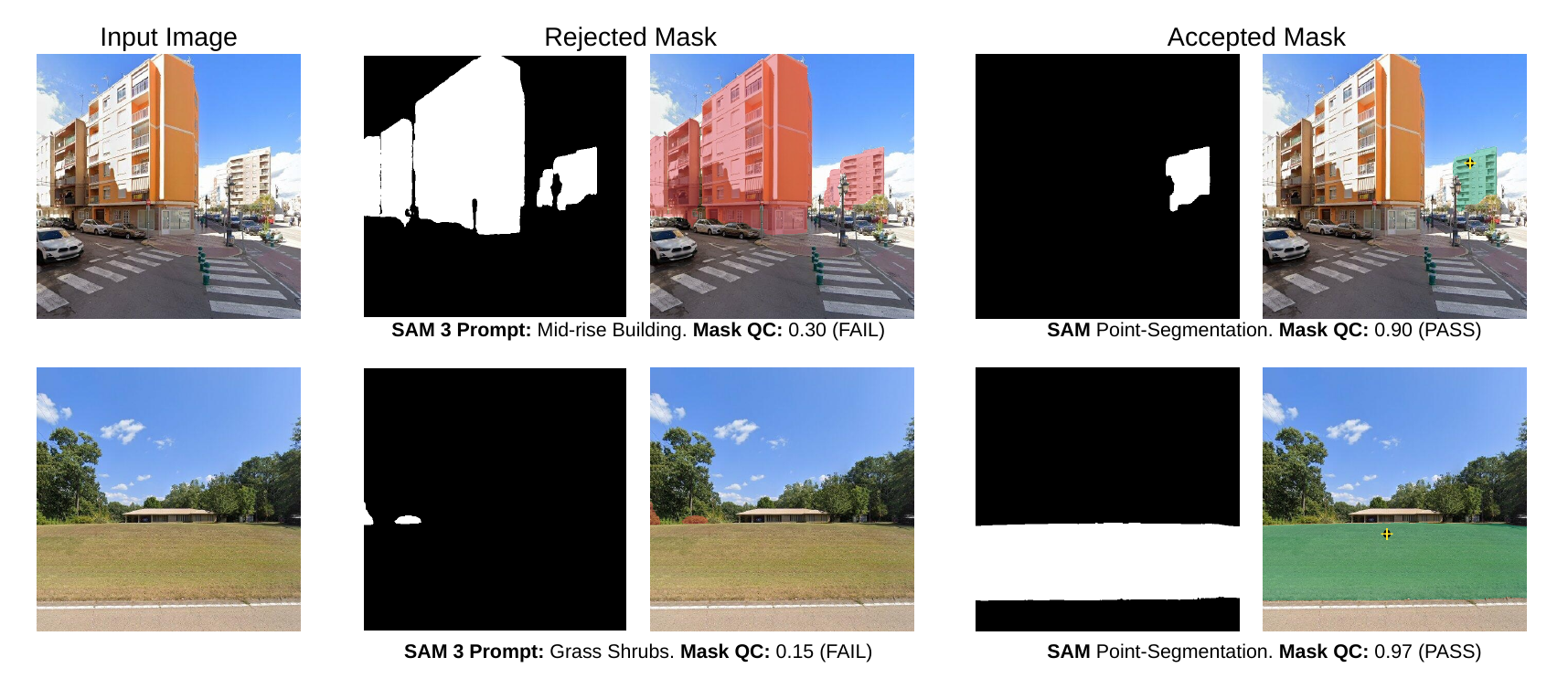}
        \caption{Workaround example for a rejected mask after QC. Another tool was used to generate the successful mask.}
        \label{fig:mask_fallback}
    \end{subfigure}

    \vspace{6pt}

    \begin{subfigure}{\textwidth}
        \centering
        \includegraphics[width=0.80\textwidth]{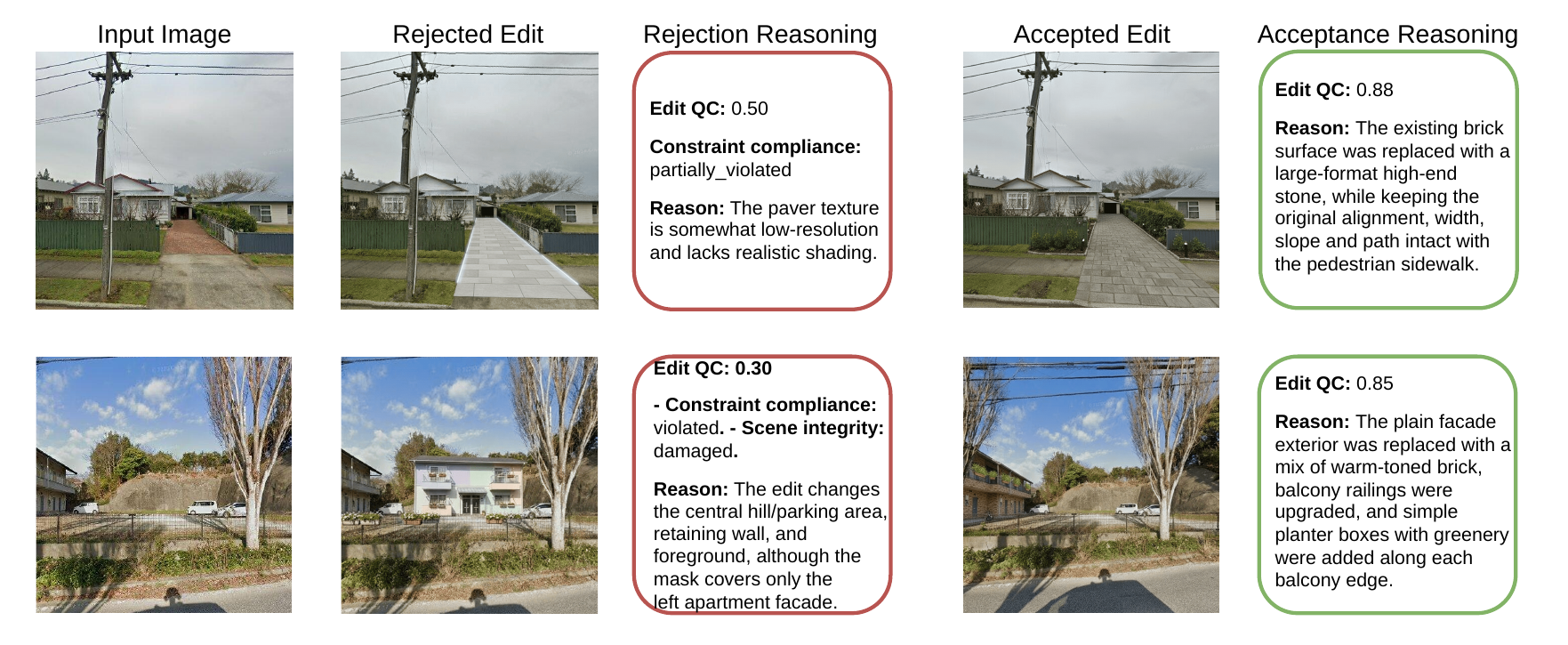}
        \caption{Workaround example for a rejected edit after QC. Prompts were revised to generate better, and more realistic interventions.}
        \label{fig:edit_fallback}
    \end{subfigure}

    \caption{
        Qualitative examples from the Segmentation and Generation Agents outputs, indicating originally failed outputs, that were later corrected to pass the QC.
    }
    \label{fig:fallbacks}
\end{figure*}

In the segmentation stage, failure cases can be addressed by revising the segmentation prompt, switching to an alternative segmentation tool, or applying mask smoothing when fragmentation is detected. In~\cref{fig:bad_masks}, the intended region of interest is the building on the right, but the SAM3 prompt "mid-rise building" would segment every building in the image. Therefore, using the location point from the Planner Agent with SAM could select the correct building in the image as shown in~\cref{fig:mask_fallback}. 

Similarly, generation failures such as those shown in~\cref{fig:bad_edits} can trigger a revised editing prompt before the intervention is regenerated. These fallback mechanisms allow the system to adapt its strategy without restarting the full pipeline, while preserving the same ROI and intervention objective. In the first example, the pavement originally looked unrealistic and was corrected in the next editing stage. In the bottom example, the first edit was made in the wrong region of the image, outside the masked area. It was rejected and corrected in the second iteration as illustrated in~\cref{fig:edit_fallback}.

\section{Details on FID-Proxy}
\label{sec:fid}
A reference Gaussian $\mathcal{N}(\boldsymbol{\mu}, \boldsymbol{\Sigma})$ is estimated from a held-out set of real satellite/street-view images using 2048-dimensional feature vectors extracted from a pretrained InceptionV3 network, with the final classifier replaced by an identity layer. 
For each generated image $x$, the domain typicality score is:

\begin{equation}
    d(x) = \sqrt{\left(f(x) - \boldsymbol{\mu}\right)^{\top}
                  \boldsymbol{\Sigma}^{-1}
                  \left(f(x) - \boldsymbol{\mu}\right)}
    \label{eq:mahalanobis}
\end{equation}

\noindent where $f(x) \in \mathbb{R}^{2048}$ is the embedded feature vector of image $x$, and $\boldsymbol{\Sigma}^{-1}$ is computed via Cholesky decomposition with a small regularization term $\epsilon \mathbf{I}$ for numerical stability.  Unlike set-level FID, this formulation yields a per-image score, enabling image-level comparison across individual edits and models. 
A lower value $d(x)$ indicates that the edited image is more typical of the real aerial or street-view domain.

\section{Human Evaluation Analysis}
\label{sec:huamn_eval_analysis}

\subsection{Evaluators Background}
\label{sec:evaluators_background}
This section dives deeper into the analysis of the human evaluation results. The results are broken down by question type and metric.
Seven graduate student evaluators participated in the study, each reviewing 30 image pairs comparing outputs from \vida\ and DIFFusion or LANCE separately, for a total of 210 pairwise trials per comparison. Similarly, five users with expertise in Urban Planning and Policy Design from Prolific participated in the study. The left–right placement of the two methods was randomized. 

Note that the graduate students were from the university department and were not members of the paper’s author team.  During recruitment, they were informed about the evaluation and provided verbal consent for their responses to be used in the study. Participation was voluntary, and no names, contact details, or other personal information were collected.

\subsection{Human Evaluation Results}
\label{sec:evaluators_results}
Table~\ref{tab:student-eval} summarizes the graduate-student evaluation of \vida\ against DIFFusion and LANCE across perceptual attribute alignment, realism, and urban-planning feasibility. The table additionally provides attribute-level results for the perceptual evaluation.

\noindent\textbf{Expert Evaluation.}
We further evaluate \vida\ against DIFFusion and LANCE using domain experts. Table~\ref{tab:expert-eval} summarizes their preferences across the same three evaluation questions, with a detailed breakdown of the perceptual attributes.

\noindent\textbf{Agreement with LLM Judges.}
We further examine whether human decisions align with the LLM judges' evaluations. Specifically, we use the LLM judges' scores for \vida\ and each baseline to make pairwise decisions analogous to those made by the human evaluators. Table~\ref{tab:llm-human-agreement} reports agreement with both student and expert majority decisions across DIFFusion and LANCE comparisons. Agreement on realism and urban-planning feasibility is consistently above the 50\% random baseline, showing that LLM judges can serve as a scalable proxy for more expensive human evaluation.

\begin{table*}[t]
\centering
\caption{Graduate Student preference for \vida\ over DIFFusion and LANCE.
Bold rows report the overall result for each evaluation question.
For Q1, we additionally report preference by targeted perceptual
attribute. Each entry gives the number and percentage of judgments
favoring \vida.}
\label{tab:student-eval}
\small
\setlength{\tabcolsep}{7pt}
\renewcommand{\arraystretch}{1.12}

\begin{tabular}{lcc}
\toprule
\textbf{Evaluation}
& \textbf{\vida\ vs. DIFFusion}
& \textbf{\vida\ vs. LANCE} \\
\midrule

\textbf{Q1 -- Metric Preference}
& \textbf{123/210 (58.6\%)}
& \textbf{189/210 (90.0\%)} \\

\quad $\hookrightarrow$ Lively
& 28/35 (80.0\%) & 34/35 (97.1\%) \\
\quad $\hookrightarrow$ Beautiful
& 25/35 (71.4\%) & 34/35 (97.1\%) \\
\quad $\hookrightarrow$ Less Boring
& 25/35 (71.4\%) & 29/35 (82.9\%) \\
\quad $\hookrightarrow$ Less Depressing
& 18/35 (51.4\%) & 32/35 (91.4\%) \\
\quad $\hookrightarrow$ Safe
& 17/35 (48.6\%) & 34/35 (97.1\%) \\
\quad $\hookrightarrow$ Wealthy
& 10/35 (28.6\%) & 26/35 (74.3\%) \\

\addlinespace[5pt]

\textbf{Q2 -- Realism}
& \textbf{201/210 (95.7\%)}
& \textbf{160/210 (76.2\%)} \\

\addlinespace[5pt]

\textbf{Q3 -- Urban Planning}
& \textbf{183/210 (87.1\%)}
& \textbf{161/210 (76.7\%)} \\

\bottomrule
\end{tabular}
\end{table*}

\begin{table*}[t]
\centering
\caption{Expert preference for \vida\ over DIFFusion and LANCE.
Bold rows report the overall result for each evaluation question.
For Q1, we additionally report preference by targeted perceptual
attribute. Each entry gives the number and percentage of judgments
favoring \vida.}
\label{tab:expert-eval}
\small
\setlength{\tabcolsep}{7pt}
\renewcommand{\arraystretch}{1.12}

\begin{tabular}{lcc}
\toprule
\textbf{Evaluation}
& \textbf{\vida\ vs. DIFFusion}
& \textbf{\vida\ vs. LANCE} \\
\midrule

\textbf{Q1 -- Metric Preference}
& \textbf{92/150 (61.3\%)}
& \textbf{130/150 (86.7\%)} \\

\quad $\hookrightarrow$ Lively
& 21/25 (84.0\%) & 22/25 (88.0\%) \\
\quad $\hookrightarrow$ Beautiful
& 18/25 (72.0\%) & 24/25 (96.0\%) \\
\quad $\hookrightarrow$ Less Boring
& 15/25 (60.0\%) & 19/25 (76.0\%) \\
\quad $\hookrightarrow$ Less Depressing
& 10/25 (40.0\%) & 23/25 (92.0\%) \\
\quad $\hookrightarrow$ Safe
& 15/25 (60.0\%) & 24/25 (96.0\%) \\
\quad $\hookrightarrow$ Wealthy
& 13/25 (52.0\%) & 18/25 (72.0\%) \\

\addlinespace[5pt]

\textbf{Q2 -- Realism}
& \textbf{142/150 (94.7\%)}
& \textbf{100/150 (66.7\%)} \\

\addlinespace[5pt]

\textbf{Q3 -- Urban Planning}
& \textbf{137/150 (91.3\%)}
& \textbf{119/150 (79.3\%)} \\

\bottomrule
\end{tabular}
\end{table*}


\begin{table}[t]
\centering
\caption{Agreement between LLM judges and human majority preferences.
For each image pair and evaluation question, agreement indicates
whether the LLM judge selected the same preferred output as the
corresponding human majority. Human ties are excluded.}
\label{tab:llm-human-agreement}
\small
\setlength{\tabcolsep}{5pt}
\renewcommand{\arraystretch}{1.12}

\begin{tabular}{llccc}
\toprule
Judge & Human Group
& \makecell{Q1\\Metric}
& \makecell{Q2\\Realism}
& \makecell{Q3\\Planning} \\
\midrule

\multicolumn{5}{l}{\textit{\vida\ vs. DIFFusion}} \\[2pt]

\multirow{2}{*}{Qwen3-VL}
& Students & 46.4\% & 93.1\% & 89.3\% \\
& Experts  & 50.0\% & 96.6\% & 92.9\% \\

\addlinespace[2pt]

\multirow{2}{*}{GPT-5.2}
& Students & 48.3\% & 92.9\% & 96.6\% \\
& Experts  & 51.7\% & 96.4\% & 100.0\% \\

\midrule

\multicolumn{5}{l}{\textit{\vida\ vs. LANCE}} \\[2pt]

\multirow{2}{*}{Qwen3-VL}
& Students & 58.6\% & 77.3\% & 80.8\% \\
& Experts  & 65.5\% & 81.8\% & 80.8\% \\

\addlinespace[2pt]

\multirow{2}{*}{GPT-5.2}
& Students & 55.2\% & 75.0\% & 85.2\% \\
& Experts  & 82.8\% & 79.2\% & 92.6\% \\

\bottomrule
\end{tabular}
\end{table}

\section{Sequential Editing Artifacts}
\label{app:sequential}
\begin{figure}[t]
    \centering
    \includegraphics[width=\linewidth]{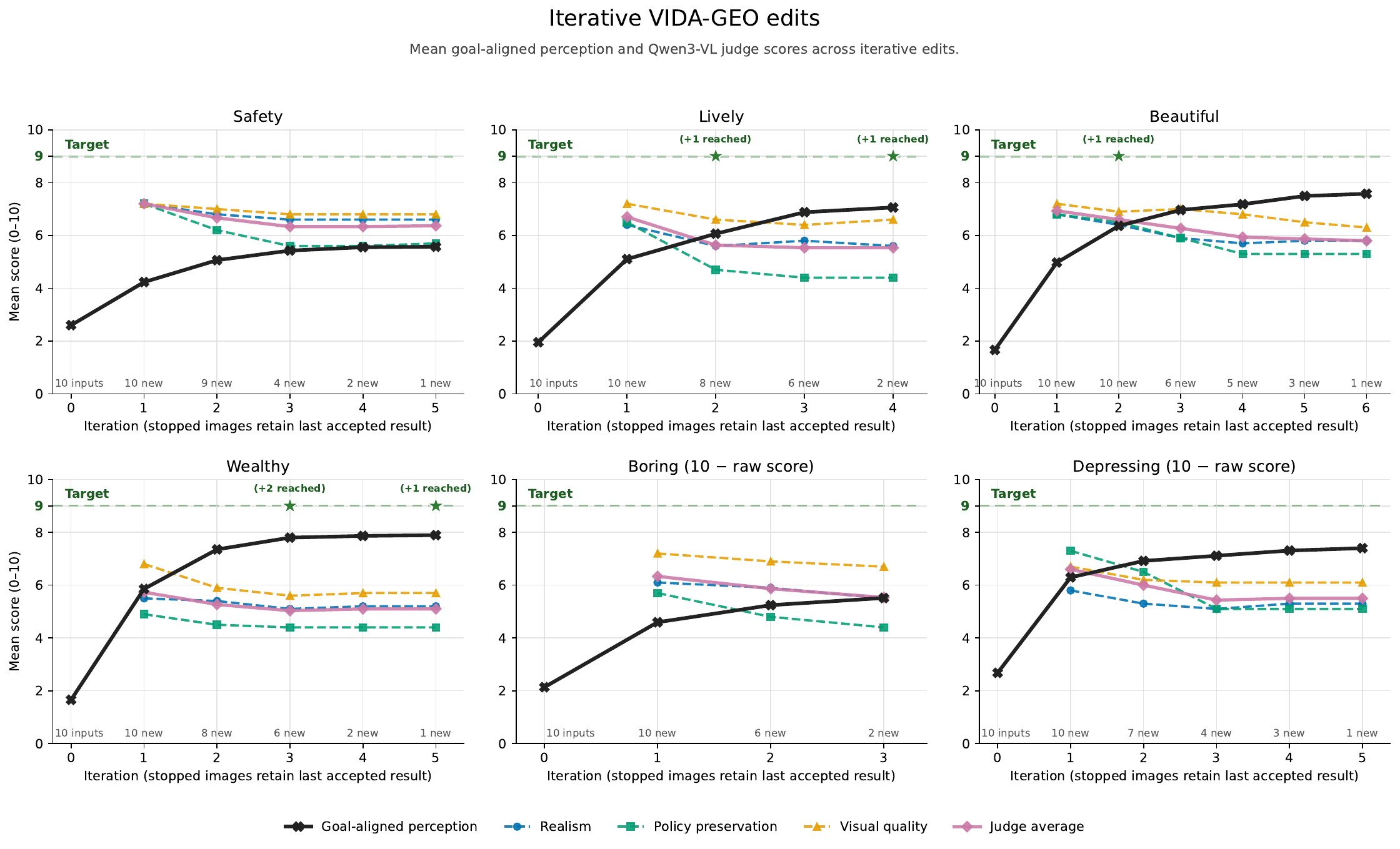}
    \caption{Sequential editing behavior of \vida\ across six Street View perception tasks. 
    Each panel reports the mean goal-aligned perception score together with Qwen3-VL realism, 
    policy-preservation, visual-quality, and average LLM-Judge scores over sequential edits of 
    10 inputs. For \textit{Boring} and \textit{Depressing}, the raw perception score is inverted 
    ($10-\text{raw score}$) so that higher values consistently indicate progress toward the target. 
    Editing stops for an input when no acceptable candidate passes QC or when the target is reached; 
    stopped inputs retain their last accepted result in subsequent cohort averages.}
    \label{fig:iterative_scored}
\end{figure}

One concern with sequential editing is the accumulation of artifacts from one iteration to the next, as some acceptable edits may already receive lower LLM-Judge scores after the first iteration and continue to degrade with additional edits. To further analyze this behavior, we iteratively edit 10 inputs per Street View metric and stop when no acceptable edit passes QC or if the edited image reaches the desired goal for the metric (9.0 for increasing metrics and 1.0 for decreasing metrics). We then evaluate the outputs using the LLM-Judge metrics discussed in \cref{sec:evaluation_metrics}, and the results can be seen in \cref{fig:iterative_scored}. This showcases the trade-off between continual editing and the LLM-as-a-Judge scoring. Across tasks, the first iteration produces the largest improvement in the target indicator and generally the highest average LLM-Judge scores, but the following iterations indicate a slight drop in perceptual quality and policy alignment scores. Despite this decline, the average LLM-Judge scores at the final iteration remain above 5.0, suggesting that the resulting edits remain reasonably realistic and policy-aligned.

\section{Using Generative Models + Black Box Indicator Predictions as IDTs}
\label{sec:world_model}

\begin{figure}[h]
  \centering
  \includegraphics[width=0.75\linewidth]{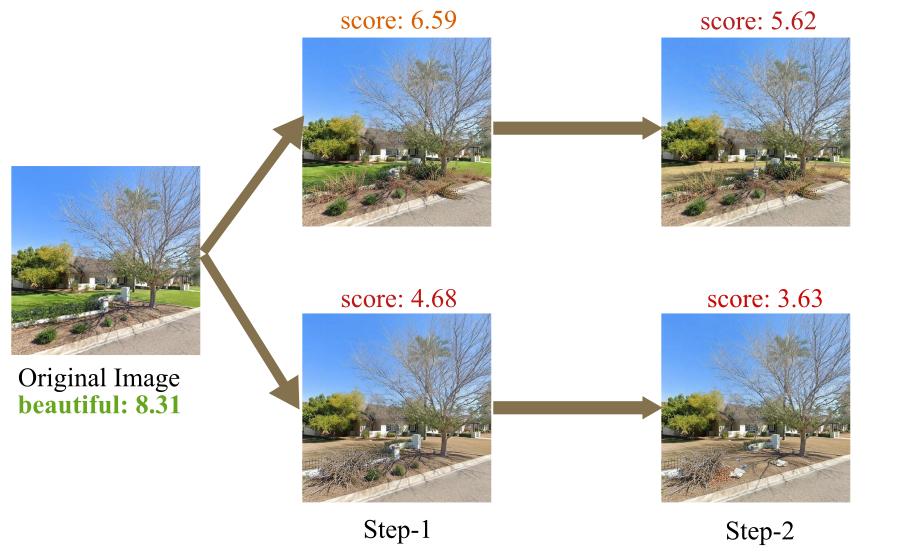}
  \vspace{-6pt}
  \caption{Counterfactual scene edits and corresponding predictor outputs. \vida\ generates alternative versions of the same street-view image through two distinct editing directions. As vegetation is removed and the environment becomes less maintained, the black-box "beauty" predictor assigns progressively lower scores. The consistent and interpretable score changes suggest that the predictor captures meaningful scene attributes under counterfactual interventions, supporting its use as a component of a world model.}
  \label{fig:counterfactuals}
  \vspace{-8pt}
\end{figure}

In \cref{sec:intro}, we mention that a model can be considered an \textit{Implicit Digital Twin} if it has the ability to not only make correct predictions on the in-distribution test data similar to the training data but also on counterfactuals that may be out-of-distribution. 
Here we present brief qualitative evidence that the black-box models we have chosen perform fairly well on counterfactual edits.
\Cref{fig:counterfactuals}, shows changes to the score from these black box model when certain changes are made. 
While exact score is hard to interpret, the relative increase or decrease is logical.

\section{Implementation and Experimental Setup Details}
\label{sec:setup}
The pipeline integrates four models, FLUX-Fill~\cite{flux}, LISAt~\cite{lisat}, SAM~\cite{kirillov2023segment}, and SAM3~\cite{carion2025sam3segment}, which serve as local REST microservices. 
Region planning and all quality-control agents use \textbf{GPT-5.1}~\cite{singh2025openai}. 
Semantically constrained regions rely on \textbf{Gemini 2.5 Flash}~\cite{comanici2025gemini}, also referred to as NanoBanana 2.5 (NB2.5), which edits via scene-level instruction rather than masked pixel replacement. 
All the cloud calls to GPT-5.1 and NB2.5 carry no local GPU requirements. 
The thresholds used in the quality check steps using the LLM judge were {0.7}; the number of trials to mask and edit each region is limited to 3 trials per epoch. We set the steps to 30 and guidance to 20 for FLUX.

All experiments ran on a NVIDIA RTX 5000 Ada GPU with 32GB of memory. The simplest setup for the Agent is to use SAM/SAM3 and the black-box indicators which can occupy up to 10GB of memory. Experiments using the Street-view images, for 3 epochs, and NanoBanana as the generation model, take 6 minutes and 12 seconds on average. Experiments using satellite images, for 3 epochs using NanoBanana, take 13 minutes and 42 seconds on average.

\section{Prompts Used Within \vida}
\label{sec:prompts}
This section reports the prompts and request formats for \vida\ , that affect
region selection, policy constraints, image generation, and evaluation. The complete
domain-specific prompts are provided in the first link of the supplementary repository.

\begin{methodbox}{Text-Segmentation Request Format}
LISAt and SAM3 do not receive a separate natural-language system prompt.
Instead, they receive the planner-generated \texttt{seg\_keyword}: a concrete
one- or two-word noun describing the target region, such as
\texttt{road}, \texttt{parking lot}, \texttt{street lamp}, or
\texttt{tree canopy}.

The SAM fallback receives only point coordinates and foreground/background
labels.
\end{methodbox}

\begin{methodbox}{Policy-Agent Decision Format}
For each proposed region, the Policy Agent receives its identifier, label,
description, edit type, and target metric. It assigns one of four edit classes:

\begin{itemize}
    \item \texttt{free}: the proposed edit is physically plausible and may be
    applied directly;
    \item \texttt{constrained}: the region may be edited, but its physical or
    functional role must be preserved;
    \item \texttt{skip}: the proposed intervention is physically implausible or
    incompatible with the scene.
\end{itemize}

For constrained regions, the agent must provide an explicit
\texttt{strategy\_prompt\_hint} describing what must remain intact. Every
proposed region must appear in the structured output.
\end{methodbox}

\begin{methodbox}{NanoBanana Edit Request Format}
\textbf{Masked editing.} The image editor receives the original image and a
binary mask, where white pixels denote the only region that may be changed.

\medskip
\noindent
\textit{Task:} \texttt{\{suggested\_edit\_prompt\}}

\begin{itemize}
    \item Modify only the white mask region.
    \item Preserve all pixels outside the white region.
\end{itemize}

\textbf{Maskless fallback.} If masked editing fails, the editor receives the
original image and the same edit instruction, together with the following
constraints:

\begin{itemize}
    \item Keep all unrelated scene content unchanged.
    \item Blend the modification naturally with the existing scene.
    \item Do not add text, labels, watermarks, or borders.
\end{itemize}
\end{methodbox}

For LLM-as-a-Judge evaluation, the LLM-Judge receives the original image first and the
edited image second. All three criteria are scored jointly in one request.

\begin{methodbox}{LLM-as-a-Judge Prompt}
You are a strict and consistent judge of an image edit. Be critical and
discerning.

You receive two images: first the original/reference image and then the
edited/output image. Judge only the edited image, using the original image to
identify what changed.

Score each of the following criteria on an integer scale from 1 to 10. Use the
full range, reserve 9--10 for nearly flawless results, and assign low scores
when substantial failures are present.

\begin{enumerate}
    \item \textbf{realism}: Assess whether the edited world is physically, semantically, and contextually plausible in the specific scene. Objects should have plausible function, material, scale, geometry, attachment, and spatial relationships. Roads, sidewalks, doors, windows, wires, roofs, trees, parking areas, and buildings should fit the surrounding structure and land use. Penalize edits that insert context-breaking or excessive features, such as objects pasted into an implausible setting, repeated objects that would not actually exist there, or changes that conflict with the overall building, street, or neighborhood context. 
    \item \textbf{policy\_preservation}: Assess whether
    the edited image preserves the original scene’s core structure and function. Compare against the reference image. Penalize removal or severe distortion of roads, buildings, sidewalks, vehicles, street layout, vegetation layout, utility poles/wires, doors/windows, building envelopes, or essential infrastructure. Reward edits that keep the original scene recognizable and operational while changing only the intended local appearance. 
    \item \textbf{visual\_quality}: Assess whether the edited image looks like a natural, authentic image in the same capture modality as the original. For street-view images, it should look like a real GSV/photo capture, not a cartoon, CGI render, painting, collage, or oddly formatted/generated image. Penalize synthetic style, over-smoothed AI texture, strange color processing, low sharpness, seams, warping, ghosting, garbled textures, melted objects, obvious compositing artifacts, or any visual presentation that makes the image itself feel unnatural. This criterion evaluates image naturalness and artifact quality, not whether the edited world is a good intervention. 
    \end{enumerate}

Judge each criterion holistically. Return only a JSON object containing exactly
the following integer fields and no additional text:

\begin{center}
\texttt{\{"realism": $s_r$, "policy\_preservation": $s_p$,
"visual\_quality": $s_v$\}}
\end{center}

where $s_r$, $s_p$, and $s_v$ are integers from 1 to 10.
\end{methodbox}

The reported LLM Judge Average is the arithmetic mean of the realism,
policy-preservation, and visual-quality scores.

\section{Handling of Failure Cases}
\label{sec:fail_cases}
Inputs are not excluded when \vida\ produces no accepted edit. This occurs when the original image already has a relatively optimal indicator score, or when no candidate edit satisfies the acceptance criteria; therefore, \vida\ returns the original input image and flags it as “No Successful Edit.” The input still remains part of the benchmark rather than being removed from the evaluation. A representation of the failure rate using the Street View images is as follows in \cref{tab:edit_failures}. This showcases an error rate of 1.7\% for \vida's pipeline.

\begin{table}[ht]
    \centering
    \caption{Edit success and failure counts by metric. Each metric contains 100 evaluated inputs.}
    \label{tab:edit_failures}
    \small
    \begin{tabular}{lccc}
        \toprule
        \textbf{Metric}
        & \textbf{Successful}
        & \makecell{\textbf{No Improving}\\\textbf{Edit}}
        & \makecell{\textbf{Failure}\\\textbf{per 100}} \\
        \midrule
        Safety     & 96  & 4 & 4\% \\
        Lively     & 100 & 0 & 0\% \\
        Beautiful  & 98  & 2 & 2\% \\
        Wealthy    & 99  & 1 & 1\% \\
        Boring     & 99  & 1 & 1\% \\
        Depressing & 98  & 2 & 2\% \\
        \midrule
        \textbf{Total} & \textbf{590} & \textbf{10} & \textbf{1.7\%} \\
        \bottomrule
    \end{tabular}
\end{table}

\section{Analysis of the Best Output for Each Model}
\label{app:best_gpt}
In order to make a fairer comparison between the outputs of the models, a smaller sample of outputs is selected to represent the best single output image for each model for each task. The results are shown in \cref{tab:best_gpt52}, which summarizes the perceptual quality and policy-alignment results for every "Best" output for each input image. It is clear that \vida\ still outperforms LANCE and DIFFusion even when the best selected image is isolated and separately evaluated.


\begin{table*}[h]
    \centering
    \caption{Human-aligned quality, intervention strength, and distributional realism proxy for the \textit{\textbf{Best}} edited satellite and Street View images across urban intervention tasks, evaluated by \textbf{GPT-5.2}. Tasks shown in \textcolor{mygreen}{green} correspond to indicators where higher scores are desirable, while tasks shown in \textcolor{myred}{red} correspond to indicators where lower scores are desirable. Bold judge scores indicate that the best-performing method is statistically separated from all alternatives by non-overlapping 95\% confidence intervals.}
    \label{tab:best_gpt52}
    \small
    \setlength{\tabcolsep}{6pt}
    \resizebox{\textwidth}{!}{%
    \begin{tabular}{llccccccc}
        \toprule
        \textbf{Task} & & &
        & \multicolumn{2}{c}{\textbf{Perceptual Quality}}
        & \multicolumn{2}{c}{\textbf{Policy Alignment}}
        & \\
        \cmidrule(lr){5-6}\cmidrule(lr){7-8}
        & \textbf{Method}
        & \makecell{\textbf{Output}\\\textbf{Edit Avg.} (\%)}
        & \makecell{\textbf{Delta}\\\textbf{Avg.} (\%)}
        & \makecell{\textbf{FID-Proxy}\\($\downarrow$)\\{\scriptsize $\pm$ 1.97}}
        & \makecell{\textbf{Visual}\\\textbf{Quality} (\%)\\{\scriptsize $\pm$ 2.48}}
        & \makecell{\textbf{Realism} (\%)\\{\scriptsize $\pm$ 2.73}}
        & \makecell{\textbf{Policy}\\\textbf{Pres.} (\%)\\{\scriptsize $\pm$ 3.45}}
        & \makecell{\textbf{LLM Judge}\\\textbf{Avg.} (\%)\\{\scriptsize $\pm$ 2.42}} \\
        \midrule

        \multirow{4}{*}{\textcolor{mygreen}{\textbf{Greenery}}}
        & \textbf{\vida}  & 44.6 & 15.3 & 45.6 & 68.3 & 59.2 & 46.1 & 57.9 \\
        & LANCE           & 36.5 & 7.1 & 57.5 & 37.3 & 27.4 & 14.6 & 26.4 \\
        & DIFFusion       & 77.8 & 48.5 & 58.3 & 21.8 & 22.1 & 27.2 & 23.7 \\
        & NB2.5 (ZS)      & 56.3 & 26.9 & 47.7 & 65.0 & 57.3 & 51.0 & 57.8 \\

        \midrule

        \multirow{4}{*}{\textcolor{myred}{\textbf{Road Risk}}}
        & \textbf{\vida}  & 70.6 & 15.5 & 49.6 & 69.7 & 61.5 & 45.8 & 59.0 \\
        & LANCE           & 38.8 & 47.3 & 61.9 & 30.0 & 27.5 & 20.4 & 26.0 \\
        & DIFFusion       & 0.0 & 86.0 & 63.2 & 19.0 & 17.2 & 13.0 & 16.4 \\
        & NB2.5 (ZS)      & 63.7 & 22.4 & 50.7 & 65.5 & 59.1 & 51.3 & 58.6 \\

        \specialrule{1.2pt}{0.8ex}{0.8ex}

        \multirow{5}{*}{\textcolor{mygreen}{\textbf{Safety}}}
        & \textbf{\vida}  & 35.1 & 17.9 & 57.2 & \textbf{66.0} & \textbf{70.5} & \textbf{66.6} & \textbf{67.7} \\
        & LANCE           & 56.1 & 38.9 & 62.0 & 53.6 & 49.8 & 26.2 & 43.2 \\
        & DIFFusion       & 78.1 & 60.9 & 67.9 & 27.7 & 35.8 & 31.7 & 31.7 \\
        & URSimulator     & 50.2 & 33.0 & 57.4 & 57.6 & 49.7 & 40.8 & 49.4 \\
        & NB2.5 (ZS)      & 68.9 & 51.8 & 62.3 & 55.8 & 54.0 & 30.2 & 46.7 \\

        \midrule

        \multirow{5}{*}{\textcolor{mygreen}{\textbf{Lively}}}
        & \textbf{\vida}  & 49.2 & 35.6 & 62.6 & 58.6 & \textbf{56.7} & \textbf{52.6} & \textbf{55.9} \\
        & LANCE           & 40.2 & 26.6 & 63.8 & 56.8 & 47.8 & 23.6 & 42.7 \\
        & DIFFusion       & 66.0 & 52.4 & 65.0 & 32.5 & 39.1 & 32.3 & 34.6 \\
        & URSimulator     & 49.8 & 36.2 & 59.1 & 50.5 & 40.1 & 32.2 & 40.9 \\
        & NB2.5 (ZS)      & 79.4 & 65.7 & 67.9 & 49.6 & 48.1 & 36.5 & 44.7 \\

        \midrule

        \multirow{5}{*}{\textcolor{mygreen}{\textbf{Beautiful}}}
        & \textbf{\vida}  & 45.4 & 27.7 & 55.7 & \textbf{63.9} & 68.6 & \textbf{63.1} & \textbf{65.2} \\
        & LANCE           & 63.7 & 45.9 & 58.8 & 50.3 & 45.7 & 25.3 & 40.4 \\
        & DIFFusion       & 81.4 & 63.7 & 67.4 & 25.2 & 30.6 & 29.2 & 28.3 \\
        & URSimulator     & 48.1 & 30.3 & 55.2 & 55.5 & 51.5 & 40.8 & 49.3 \\
        & NB2.5 (ZS)      & 70.9 & 53.2 & 62.7 & 51.4 & 63.7 & 48.7 & 54.6 \\

        \midrule

        \multirow{5}{*}{\textcolor{mygreen}{\textbf{Wealthy}}}
        & \textbf{\vida}  & 48.0 & 30.1 & 59.0 & \textbf{64.8} & \textbf{65.5} & \textbf{52.5} & \textbf{60.9} \\
        & LANCE           & 54.7 & 36.8 & 61.5 & 55.6 & 50.0 & 28.0 & 44.5 \\
        & DIFFusion       & 77.9 & 60.0 & 71.4 & 24.7 & 32.5 & 29.0 & 28.7 \\
        & URSimulator     & 54.3 & 36.4 & 58.9 & 55.8 & 45.0 & 34.3 & 45.0 \\
        & NB2.5 (ZS)      & 87.2 & 69.2 & 63.6 & 60.7 & 34.3 & 13.0 & 36.0 \\

        \midrule

        \multirow{5}{*}{\textcolor{myred}{\textbf{Boring}}}
        & \textbf{\vida}  & 57.3 & 23.3 & 58.9 & \textbf{65.6} & \textbf{70.6} & \textbf{64.3} & \textbf{66.8} \\
        & LANCE           & 42.5 & 38.1 & 59.2 & 52.3 & 49.4 & 27.8 & 43.2 \\
        & DIFFusion       & 27.7 & 52.8 & 61.7 & 28.1 & 36.0 & 31.0 & 31.7 \\
        & URSimulator     & 44.7 & 35.9 & 60.9 & 51.6 & 41.2 & 32.2 & 41.7 \\
        & NB2.5 (ZS)      & 31.4 & 49.1 & 73.5 & 40.1 & 28.7 & 28.6 & 32.5 \\

        \midrule

        \multirow{5}{*}{\textcolor{myred}{\textbf{Depressing}}}
        & \textbf{\vida}  & 58.9 & 21.2 & 59.2 & \textbf{64.1} & \textbf{69.7} & \textbf{64.4} & \textbf{66.0} \\
        & LANCE           & 36.2 & 43.9 & 61.8 & 49.4 & 44.2 & 23.9 & 39.2 \\
        & DIFFusion       & 27.9 & 52.2 & 73.0 & 25.6 & 32.4 & 28.3 & 28.8 \\
        & URSimulator     & 54.6 & 25.5 & 59.6 & 53.9 & 46.5 & 37.1 & 45.8 \\
        & NB2.5 (ZS)      & 32.8 & 47.3 & 65.4 & 52.8 & 53.6 & 36.0 & 47.5 \\

        \bottomrule
    \end{tabular}%
    }
\end{table*}


\begin{table*}[t]
    \centering
    \caption{Perceptual quality and policy alignment for edited satellite and Street View images. \textcolor{mygreen}{Green} tasks correspond to indicators where higher scores are desirable, while \textcolor{myred}{red} tasks correspond to indicators where lower scores are desirable. Visual Quality, Realism, and Policy Preservation are evaluated using \textbf{Qwen3-VL 32B}, and LLM Judge Avg. is their mean. All columns report average 95\% confidence-interval half-widths. Bold values indicate the best-performing method for each task and metric.}
    \label{tab:main_results_qwen_compact}
    \small
    \setlength{\tabcolsep}{6pt}
    \resizebox{\textwidth}{!}{%
    \begin{tabular}{llcccc}
        \toprule
        \textbf{Task}
        & \textbf{Method}
        & \makecell{\textbf{Visual}\\\textbf{Quality} (\%)\\{\scriptsize $\pm$ 0.67}}
        & \makecell{\textbf{Realism} (\%)\\{\scriptsize $\pm$ 0.86}}
        & \makecell{\textbf{Policy}\\\textbf{Pres.} (\%)\\{\scriptsize $\pm$ 1.15}}
        & \makecell{\textbf{LLM Judge}\\\textbf{Avg.} (\%)\\{\scriptsize $\pm$ 0.76}} \\
        \midrule

        \multirow{4}{*}{\textcolor{mygreen}{\textbf{Greenery}}}
        & \textbf{\vida} & \textbf{80.0} & 68.7 & 69.1 & 72.6 \\
        & LANCE             & 41.5 & 37.3 & 23.8 & 34.2 \\
        & DIFFusion         & 28.1 & 25.2 & 30.0 & 27.7 \\
        & NB2.5 (ZS)        & 78.1 & \textbf{70.2} & \textbf{77.7} & \textbf{75.3} \\

        \midrule
        \multirow{4}{*}{\textcolor{myred}{\textbf{Road Risk}}}
        & \textbf{\vida} & \textbf{80.5} & \textbf{71.5} & 65.6 & \textbf{72.6} \\
        & LANCE             & 45.1 & 43.4 & 26.9 & 38.5 \\
        & DIFFusion         & 25.3 & 23.3 & 29.1 & 25.9 \\
        & NB2.5 (ZS)        & 69.0 & 62.2 & \textbf{66.6} & 65.9 \\

        \specialrule{1.2pt}{0.8ex}{0.8ex}

        \multirow{5}{*}{\textcolor{mygreen}{\textbf{Safety}}}
        & \textbf{\vida} & \textbf{78.7} & \textbf{77.3} & \textbf{75.8} & \textbf{77.3} \\
        & LANCE             & 62.5 & 60.8 & 40.5 & 54.6 \\
        & DIFFusion         & 30.8 & 33.8 & 33.5 & 32.7 \\
        & URSimulator       & 60.7 & 49.7 & 39.3 & 49.9 \\
        & NB2.5 (ZS)        & 65.6 & 65.3 & 45.8 & 58.9 \\

        \midrule
        \multirow{5}{*}{\textcolor{mygreen}{\textbf{Lively}}}
        & \textbf{\vida} & \textbf{71.8} & \textbf{66.7} & \textbf{65.8} & \textbf{68.1} \\
        & LANCE             & 65.3 & 62.5 & 39.4 & 55.7 \\
        & DIFFusion         & 38.0 & 40.5 & 37.8 & 38.8 \\
        & URSimulator       & 54.7 & 43.8 & 34.9 & 44.4 \\
        & NB2.5 (ZS)        & 49.4 & 50.9 & 41.5 & 47.2 \\

        \midrule
        \multirow{5}{*}{\textcolor{mygreen}{\textbf{Beautiful}}}
        & \textbf{\vida} & \textbf{76.4} & \textbf{74.8} & \textbf{74.9} & \textbf{75.3} \\
        & LANCE             & 58.0 & 55.6 & 35.8 & 49.8 \\
        & DIFFusion         & 27.5 & 28.8 & 29.1 & 28.5 \\
        & URSimulator       & 62.1 & 53.2 & 38.9 & 51.4 \\
        & NB2.5 (ZS)        & 50.3 & 61.0 & 56.0 & 55.8 \\

        \midrule
        \multirow{5}{*}{\textcolor{mygreen}{\textbf{Wealthy}}}
        & \textbf{\vida} & \textbf{74.9} & \textbf{70.8} & \textbf{66.1} & \textbf{70.6} \\
        & LANCE             & 62.9 & 60.5 & 39.4 & 54.3 \\
        & DIFFusion         & 32.7 & 36.0 & 36.0 & 34.9 \\
        & URSimulator       & 56.7 & 45.3 & 35.2 & 45.7 \\
        & NB2.5 (ZS)        & 53.7 & 34.4 & 18.3 & 35.4 \\

        \midrule
        \multirow{5}{*}{\textcolor{myred}{\textbf{Boring}}}
        & \textbf{\vida} & \textbf{75.8} & \textbf{71.8} & \textbf{70.3} & \textbf{72.6} \\
        & LANCE             & 63.4 & 59.6 & 38.0 & 53.7 \\
        & DIFFusion         & 38.0 & 41.0 & 37.0 & 38.6 \\
        & URSimulator       & 54.3 & 42.7 & 32.7 & 43.2 \\
        & NB2.5 (ZS)        & 36.2 & 32.3 & 29.1 & 32.5 \\

        \midrule
        \multirow{5}{*}{\textcolor{myred}{\textbf{Depressing}}}
        & \textbf{\vida} & \textbf{75.4} & \textbf{72.4} & \textbf{74.1} & \textbf{73.9} \\
        & LANCE             & 61.2 & 57.2 & 35.7 & 51.4 \\
        & DIFFusion         & 32.3 & 36.2 & 35.1 & 34.5 \\
        & URSimulator       & 58.0 & 47.7 & 36.8 & 47.5 \\
        & NB2.5 (ZS)        & 55.4 & 56.7 & 43.8 & 52.0 \\

        \bottomrule
    \end{tabular}%
    }
\end{table*}

%

\begin{table*}[h]
    \centering
    \caption{Human-aligned quality and distributional realism proxy for the \textit{\textbf{Best}} edited satellite and Street View images across urban intervention tasks, evaluated by \textbf{Qwen3-VL 32B}. Tasks shown in \textcolor{mygreen}{green} correspond to indicators where higher scores are desirable, while tasks shown in \textcolor{myred}{red} correspond to indicators where lower scores are desirable. FID-Proxy and judge-score columns report average 95\% confidence-interval half-widths. Bold judge scores indicate that the best-performing method is statistically separated from all alternatives by non-overlapping 95\% confidence intervals.}
    \label{tab:best_qwen3vl}
    \small
    \setlength{\tabcolsep}{6pt}

    \resizebox{\textwidth}{!}{%
    \begin{tabular}{llccccc}
        \toprule

        \textbf{Task}
        & \textbf{Method}
        & \multicolumn{2}{c}{\textbf{Perceptual Quality}}
        & \multicolumn{2}{c}{\textbf{Policy Alignment}}
        & \textbf{LLM Judge} \\

        \cmidrule(lr){3-4}
        \cmidrule(lr){5-6}

        &
        & \makecell{\textbf{FID-Proxy}\\$\downarrow$\\{\scriptsize $\pm$ 1.97}}
        & \makecell{\textbf{Visual}\\\textbf{Quality} (\%)\\{\scriptsize $\pm$ 2.77}}
        & \makecell{\textbf{Realism} (\%)\\{\scriptsize $\pm$ 3.01}}
        & \makecell{\textbf{Policy}\\\textbf{Pres.} (\%)\\{\scriptsize $\pm$ 3.19}}
        & \makecell{\textbf{Avg.} (\%)\\{\scriptsize $\pm$ 2.50}} \\

        \midrule

        \multirow{4}{*}{\textcolor{mygreen}{\textbf{Greenery}}}
        & \textbf{\vida} & 45.6 & \textbf{75.6} & 59.0 & 51.5 & 62.0 \\
        & LANCE & 57.5 & 40.3 & 36.4 & 22.4 & 33.0 \\
        & DIFFusion & 58.3 & 22.3 & 22.3 & 30.7 & 25.1 \\
        & NB2.5 (ZS) & 47.7 & 68.0 & 58.4 & \textbf{70.3} & 65.6 \\

        \midrule

        \multirow{4}{*}{\textcolor{myred}{\textbf{Road Risk}}}
        & \textbf{\vida} & 49.6 & \textbf{77.7} & 66.1 & 54.2 & 66.0 \\
        & LANCE & 61.9 & 38.0 & 40.5 & 27.4 & 35.3 \\
        & DIFFusion & 63.2 & 19.2 & 12.3 & 15.9 & 15.8 \\
        & NB2.5 (ZS) & 50.7 & 65.7 & 60.8 & 61.8 & 62.8 \\

        \specialrule{1.2pt}{0.8ex}{0.8ex}

        \multirow{5}{*}{\textcolor{mygreen}{\textbf{Safety}}}
        & \textbf{\vida} & 57.2 & \textbf{76.3} & \textbf{73.0} & \textbf{72.1} & \textbf{73.8} \\
        & LANCE & 62.0 & 57.2 & 52.9 & 33.4 & 47.8 \\
        & DIFFusion & 67.9 & 30.6 & 33.5 & 30.9 & 31.7 \\
        & URSimulator & 57.4 & 58.5 & 46.2 & 38.6 & 47.8 \\
        & NB2.5 (ZS) & 62.3 & 56.5 & 58.8 & 35.8 & 50.4 \\

        \midrule

        \multirow{5}{*}{\textcolor{mygreen}{\textbf{Lively}}}
        & \textbf{\vida} & 62.6 & \textbf{68.7} & \textbf{63.3} & \textbf{52.4} & \textbf{61.5} \\
        & LANCE & 63.8 & 62.7 & 56.9 & 34.7 & 51.4 \\
        & DIFFusion & 65.0 & 36.0 & 37.3 & 32.4 & 35.2 \\
        & URSimulator & 59.1 & 51.4 & 39.4 & 31.1 & 40.6 \\
        & NB2.5 (ZS) & 67.9 & 43.5 & 45.3 & 32.1 & 40.3 \\

        \midrule

        \multirow{5}{*}{\textcolor{mygreen}{\textbf{Beautiful}}}
        & \textbf{\vida} & 55.7 & \textbf{72.1} & \textbf{66.8} & \textbf{64.7} & \textbf{67.9} \\
        & LANCE & 58.8 & 52.9 & 51.1 & 32.2 & 45.4 \\
        & DIFFusion & 67.4 & 27.5 & 29.3 & 29.8 & 28.9 \\
        & URSimulator & 55.2 & 60.9 & 51.1 & 36.3 & 49.4 \\
        & NB2.5 (ZS) & 62.7 & 45.7 & 56.6 & 49.4 & 50.6 \\

        \midrule

        \multirow{5}{*}{\textcolor{mygreen}{\textbf{Wealthy}}}
        & \textbf{\vida} & 59.0 & \textbf{70.9} & \textbf{63.4} & \textbf{54.9} & \textbf{63.1} \\
        & LANCE & 61.5 & 57.3 & 54.3 & 37.4 & 49.7 \\
        & DIFFusion & 71.4 & 26.7 & 30.0 & 29.5 & 28.7 \\
        & URSimulator & 58.9 & 55.4 & 43.6 & 33.3 & 44.1 \\
        & NB2.5 (ZS) & 63.6 & 53.1 & 31.0 & 15.2 & 33.1 \\

        \midrule

        \multirow{5}{*}{\textcolor{myred}{\textbf{Boring}}}
        & \textbf{\vida} & 58.9 & \textbf{74.1} & \textbf{69.6} & \textbf{66.0} & \textbf{69.9} \\
        & LANCE & 59.2 & 55.4 & 55.0 & 36.3 & 48.9 \\
        & DIFFusion & 61.7 & 31.2 & 34.4 & 30.9 & 32.2 \\
        & URSimulator & 60.9 & 53.7 & 41.3 & 31.8 & 42.3 \\
        & NB2.5 (ZS) & 73.5 & 32.5 & 26.1 & 25.7 & 28.1 \\

        \midrule

        \multirow{5}{*}{\textcolor{myred}{\textbf{Depressing}}}
        & \textbf{\vida} & 59.2 & \textbf{71.2} & \textbf{65.6} & \textbf{66.8} & \textbf{67.9} \\
        & LANCE & 61.8 & 53.4 & 52.5 & 31.9 & 45.9 \\
        & DIFFusion & 73.0 & 28.2 & 32.3 & 31.5 & 30.7 \\
        & URSimulator & 59.6 & 55.8 & 43.9 & 34.8 & 44.8 \\
        & NB2.5 (ZS) & 65.4 & 47.2 & 50.2 & 35.0 & 44.1 \\

        \bottomrule
    \end{tabular}%
    }
\end{table*}

\section{Analysis of Different LLM Judge}
\label{sec:llm_judge}

\Cref{tab:main_results_qwen_compact} illustrates the results of choosing a different LLM model, Qwen3 VL (qwen3-vl-32b-instruct), showcasing the consistent results for \vida\ across the perceptual quality and policy alignment evaluation metrics. Similar to \cref{app:best_gpt}, the results for every "Best" output for each input image using the Qwen3 VL judge are summarized in \cref{tab:best_qwen3vl}.

\section{Benchmark Data Statistics}
\label{app:data_statistics}
In this section, we provide additional statistics about the size and diversity of the dataset introduced in \cref{sec:dataset}. The selection process is an automated procedure that excludes cases where meaningful intervention would be infeasible, such as satellite images of an extremely dense urban area or street view images of surrounding cars on a road.
The geographic distribution of the expanded benchmark is: North America (43\%), Europe (18\%), Asia (17\%), South America (14\%), Oceania (8\%). The score distributions for the selected images are summarized in \cref{tab:data_scores} (in \%).

\begin{table}[H]
    \centering
    \caption{Summary statistics of indicator scores.}
    \label{tab:data_scores}
    \small
    \renewcommand{\arraystretch}{0.95}
    \begin{tabular}{lcc}
        \toprule
        \textbf{Indicator} & \textbf{Median Score} & \textbf{Score Range} \\
        \midrule
        Safety      & 13.0 & 3.0--49.0 \\
        Lively      & 11.0 & 2.0--43.0 \\
        Beautiful   & 15.0 & 2.0--47.0 \\
        Wealthy     & 14.0 & 2.3--48.0 \\
        Boring      & 83.0 & 54.0--98.0 \\
        Depressing  & 83.0 & 55.0--98.0 \\
        Greenery    & 29.0 & 5.0--64.0 \\
        Road Risk   & 90.0 & 51.3--99.7 \\
        \bottomrule
    \end{tabular}
\end{table}

For the satellite-image indicators, we measured the average semantic composition of the images: Urban: 57.6\%, Vegetated: 41.1\%, Road: 14.6\%, Building: 18.9\%. Across all Street View images, the average visual composition is: sky (39.9\%), buildings (17.4\%), vegetation (15.5\%), road (14.1\%), open ground (4.5\%), and sidewalk or path (3.3\%). These statistics provide a quantitative characterization of visual diversity in the benchmark.

\section{More Qualitative Examples}
\label{sec:more_examples}

\Cref{tab:qualitative_comparison} shows a few more comparisons of \vida\ vs. DIFFusion, specifically on aerial image indicators. 
While DIFFusion just increases the green saturation of the image for both increasing road safety and greenery, \vida\ provides more realistic and policy-wise practical changes.

\Cref{tab:qual_examples} shows an example of a single forward execution of a segmentation agent and generative agent.
For the first row in greenery, the segmentation agent suggests a reason to improve.
Then the planner suggests a prompt to edit said region with more trees.
Finally, the generator agent calls FLUX model and fills the empty region with mixed species of trees.
The black-box score increased by 7\% with this regional intervention.

\definecolor{mygreen}{RGB}{0,120,60}
\definecolor{myred}{RGB}{180,30,30}

\newcommand{\tableimgwidth}{0.130\textwidth}

\newcommand{\mintask}[1]{%
    {\scriptsize\textcolor{myred}{\textbf{#1}}}%
}

\newcommand{\maxtask}[1]{%
    {\scriptsize\textcolor{mygreen}{\textbf{#1}}}%
}

\newcommand{\inputcell}[1]{%
    \begin{tabular}[c]{@{}c@{}}
        \includegraphics[width=\tableimgwidth]{#1}
    \end{tabular}%
}

\newcommand{\outputcell}[4]{%
    \begin{tabular}[c]{@{}c@{}}
        \includegraphics[width=\tableimgwidth]{#1}\\[-4pt]
        {\scriptsize #2, #3, #4}
    \end{tabular}%
}

\newcommand{\nacell}{%
    \parbox[c][\tableimgwidth][c]{\tableimgwidth}{%
        \centering
        {\scriptsize\textit{N/A}}
    }%
}

\newcommand{\samplerule}{%
    \specialrule{0.10pt}{0.2pt}{0.2pt}
}


\clearpage


\begin{center}

\captionof{table}{
Qualitative comparison across selected intervention examples.
Scores beneath each generated image are reported in the order
LLM-Judge, FID-proxy, and goal gain.
\textcolor{myred}{Red} task labels denote minimization objectives and
\textcolor{mygreen}{green} task labels denote maximization objectives.
}
\label{tab:qualitative_comparison}

\vspace{3pt}

\setlength{\tabcolsep}{1pt}
\renewcommand{\arraystretch}{0.92}

\begin{tabular*}{\textwidth}{
@{\extracolsep{\fill}}
c c c c c c c
@{}
}

\toprule

{\scriptsize\textbf{Task}}
&
{\scriptsize\textbf{Input}}
&
{\scriptsize\textbf{\vida}}
&
{\scriptsize\textbf{DIFFusion}}
&
{\scriptsize\textbf{LANCE}}
&
{\scriptsize\textbf{NB2.5}}
&
{\scriptsize\textbf{URSimulator}}
\\[-2pt]

&
&
\multicolumn{5}{c}{
    {\tiny\textit{
        LLM-Judge $\uparrow$, FID $\downarrow$, Gain $\uparrow$
    }}
}
\\[-2pt]

\midrule


\maxtask{Safety}
&
\inputcell{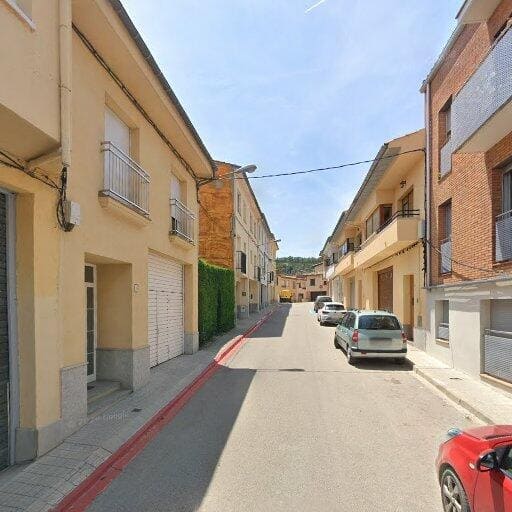}
&
\outputcell
    {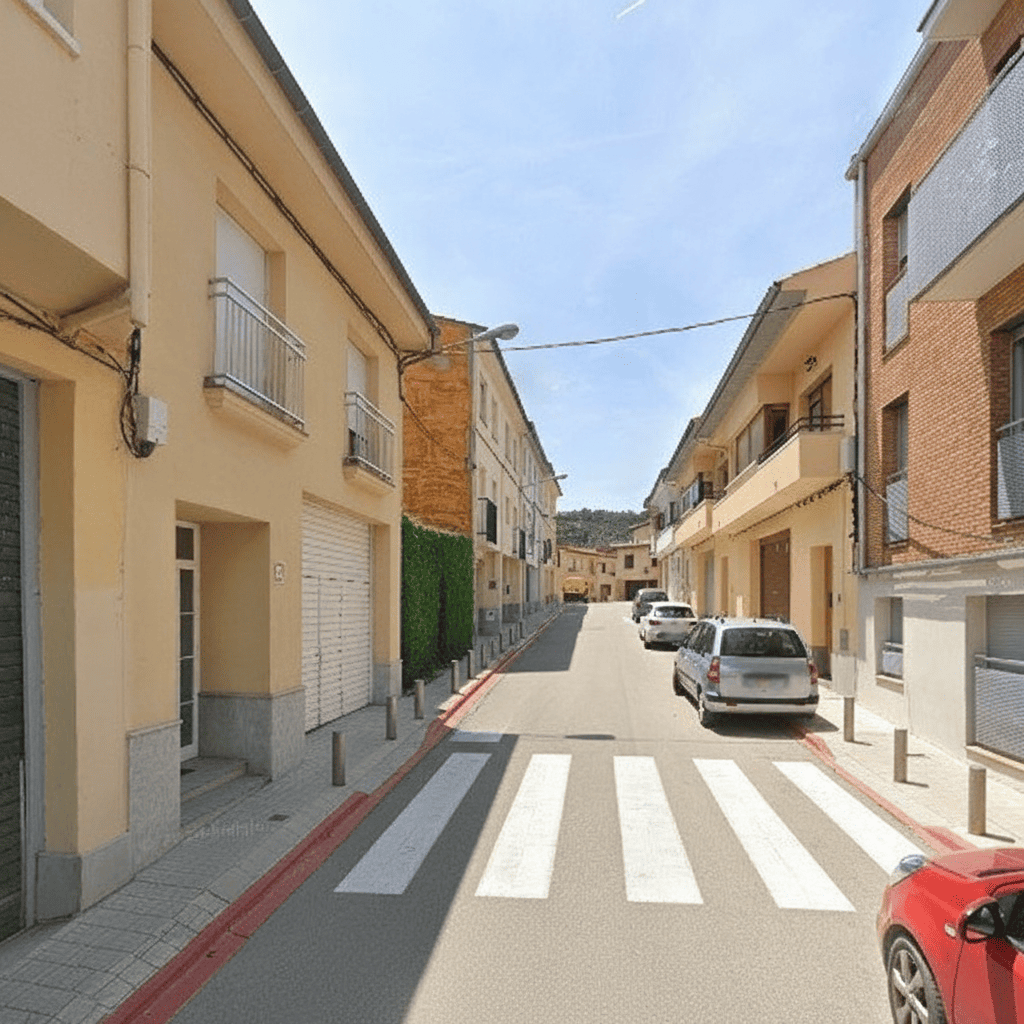}
    {86.7}{52.5}{11.0}
&
\outputcell
    {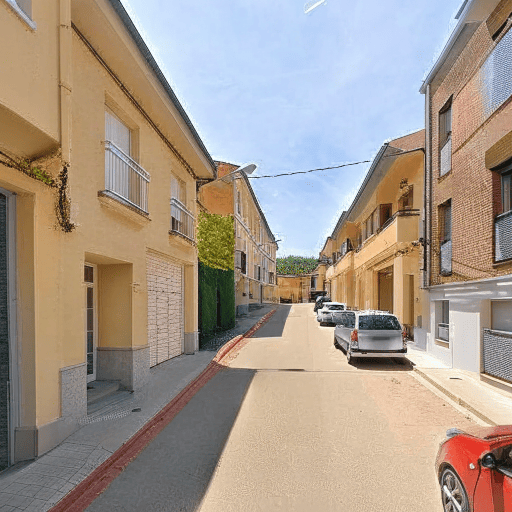}
    {86.7}{48.0}{33.8}
&
\outputcell
    {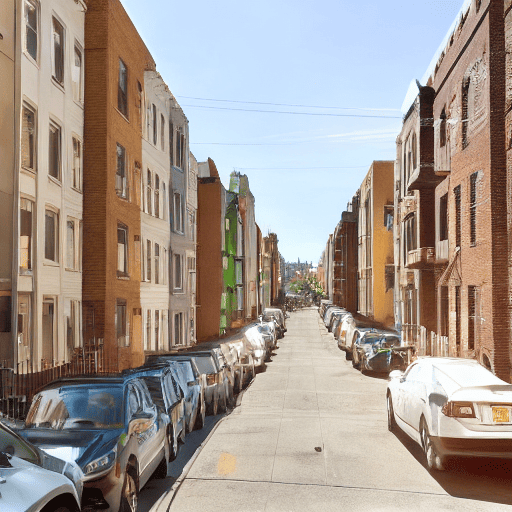}
    {26.7}{58.1}{45.3}
&
\outputcell
    {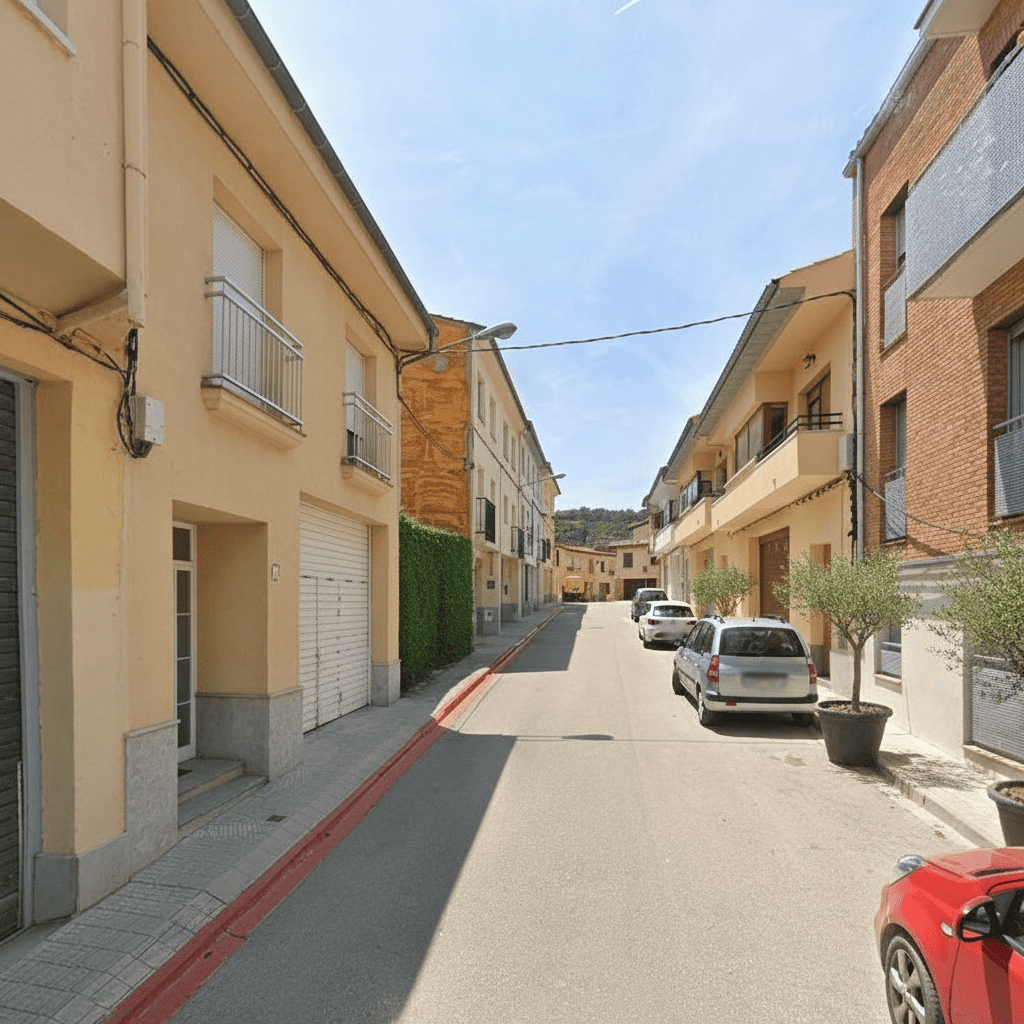}
    {83.3}{50.8}{17.2}
&
\outputcell
    {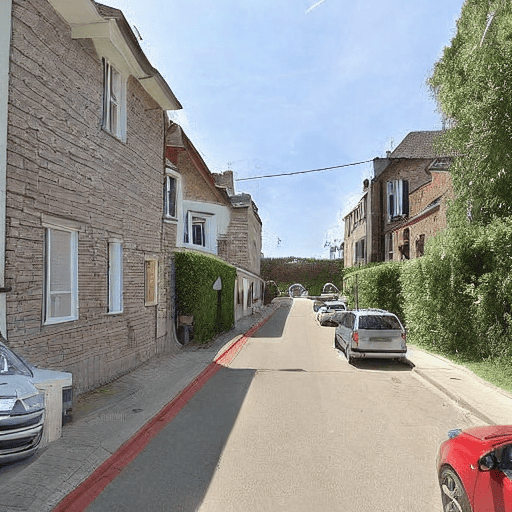}
    {60.0}{53.0}{23.9}
\\
\samplerule


\maxtask{Safety}
&
\inputcell{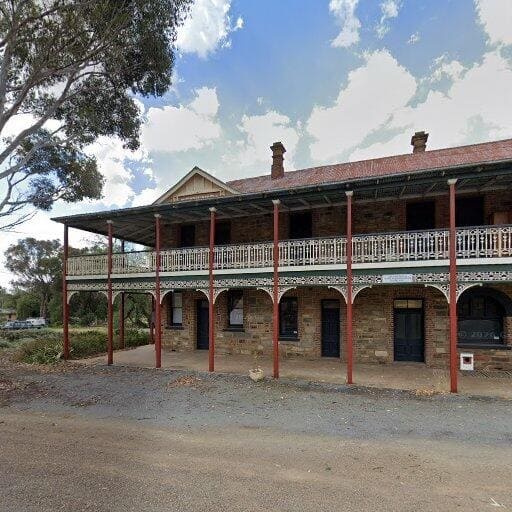}
&
\outputcell
    {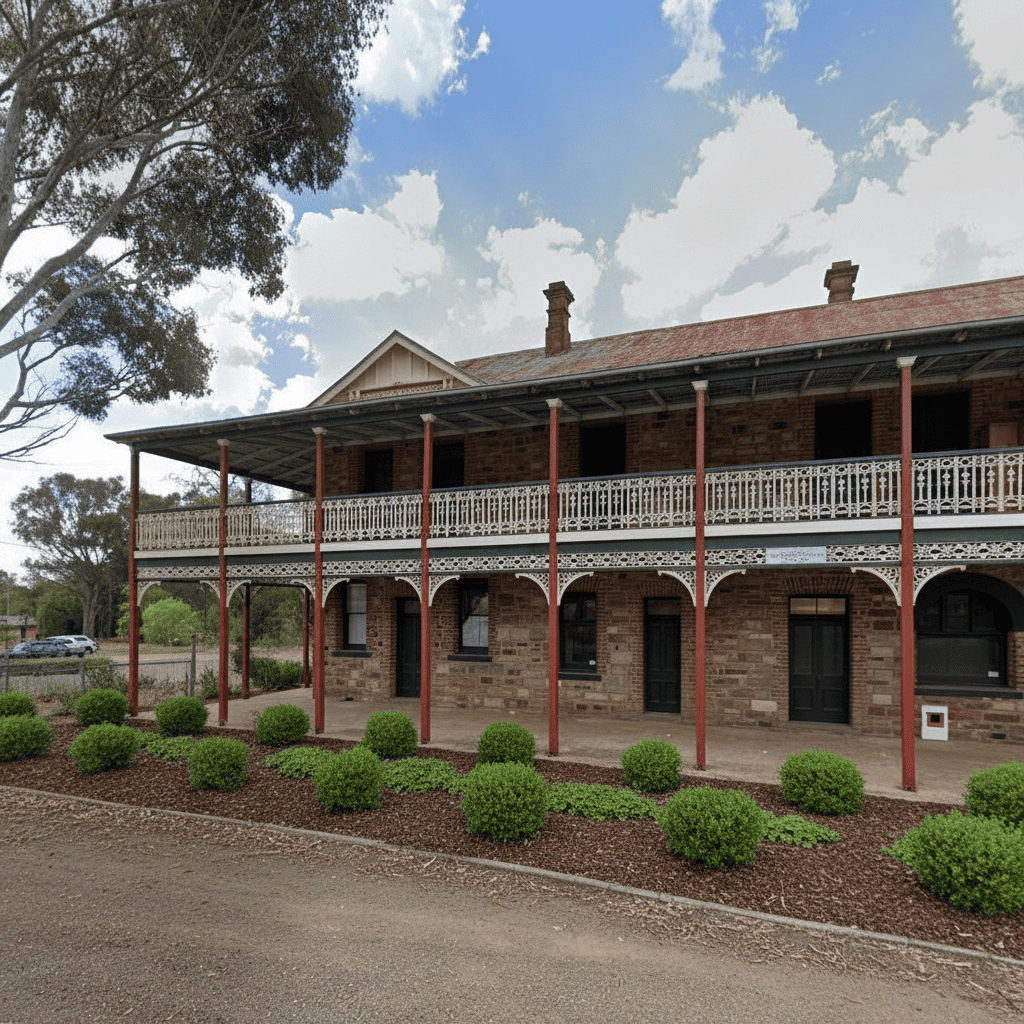}
    {83.3}{57.1}{31.6}
&
\outputcell
    {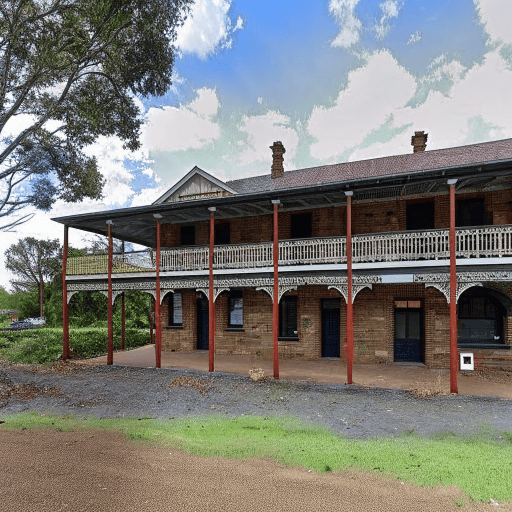}
    {73.3}{62.8}{12.9}
&
\outputcell
    {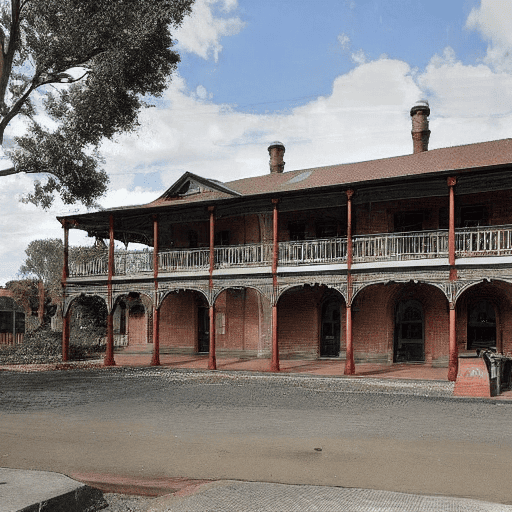}
    {73.3}{69.1}{15.1}
&
\outputcell
    {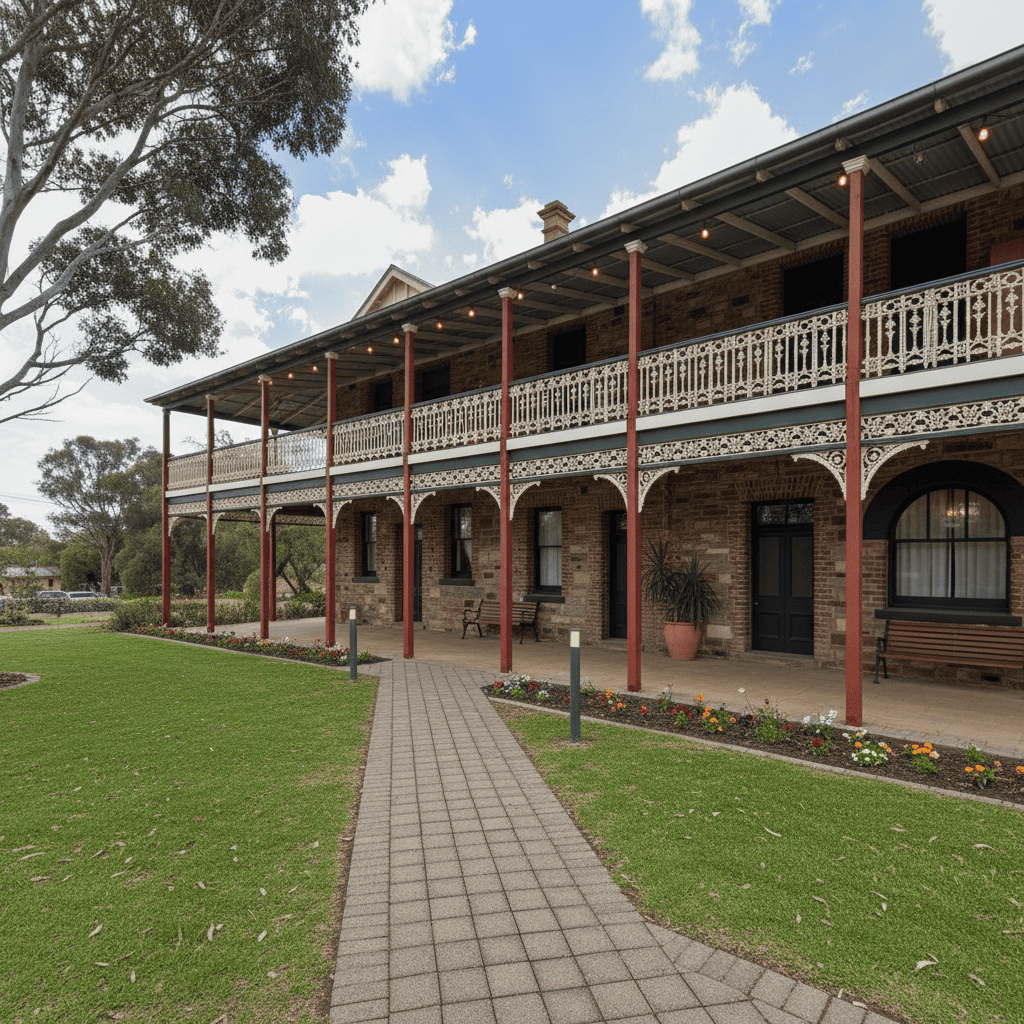}
    {70.0}{73.0}{12.2}
&
\outputcell
    {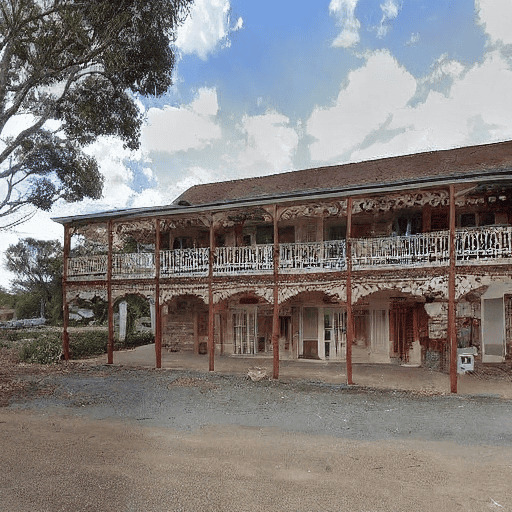}
    {63.3}{59.6}{25.3}
\\
\samplerule


\maxtask{Lively}
&
\inputcell{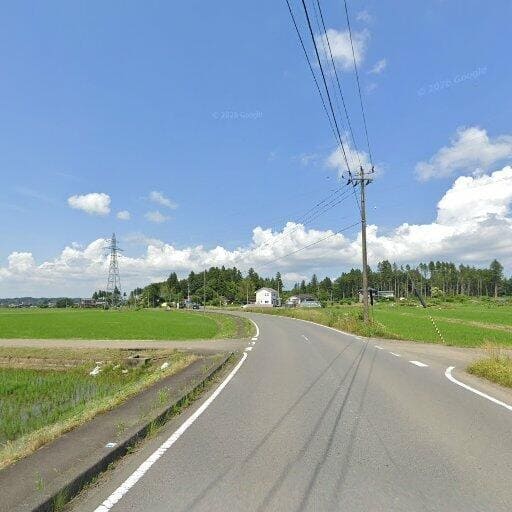}
&
\outputcell
    {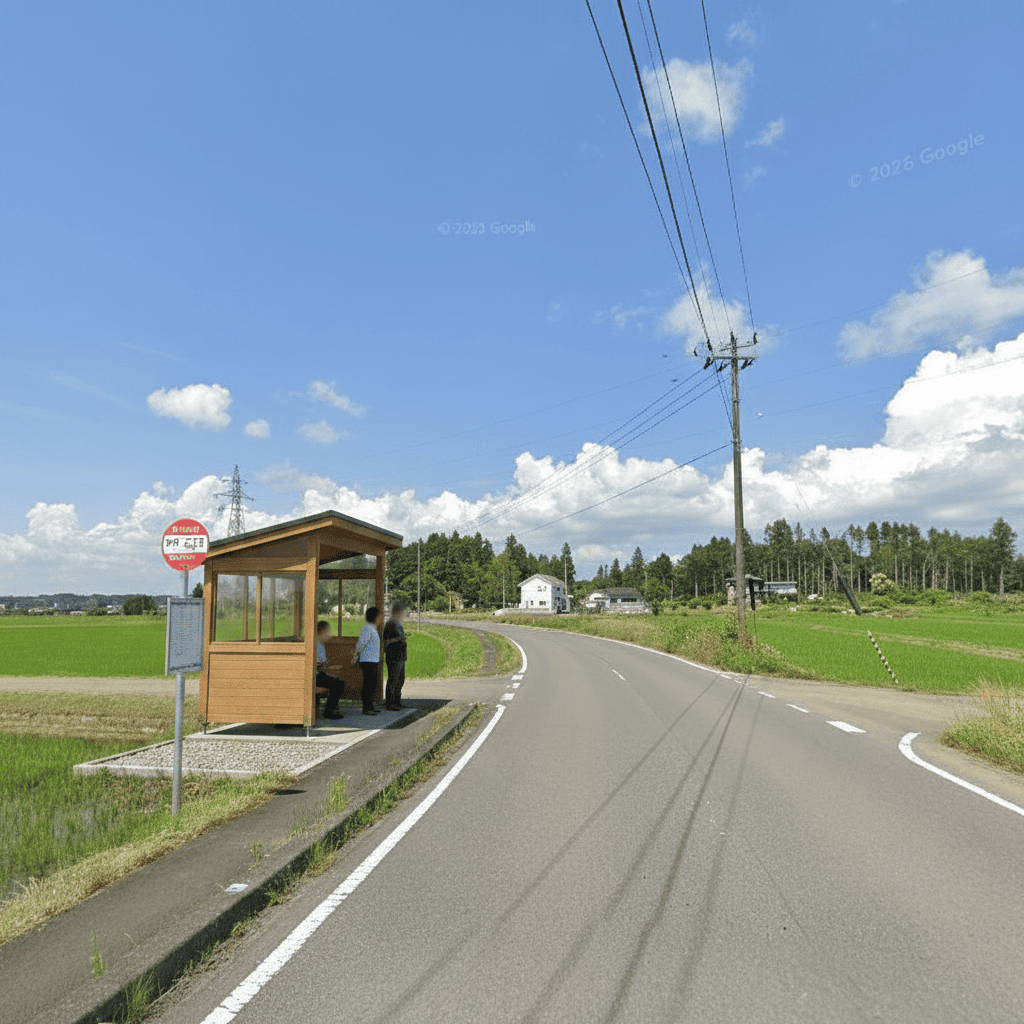}
    {76.7}{50.7}{37.3}
&
\outputcell
    {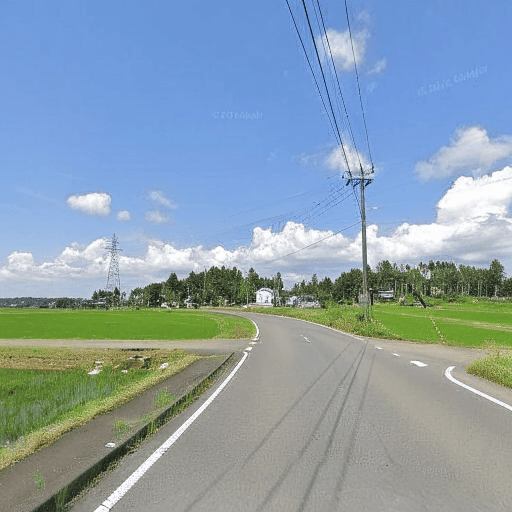}
    {90.0}{59.0}{4.7}
&
\outputcell
    {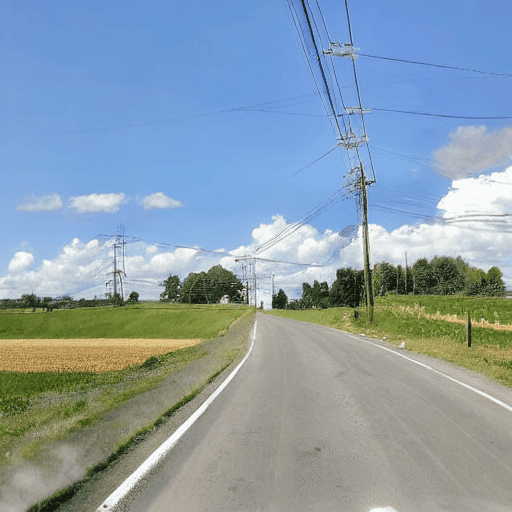}
    {60.0}{60.3}{3.0}
&
\outputcell
    {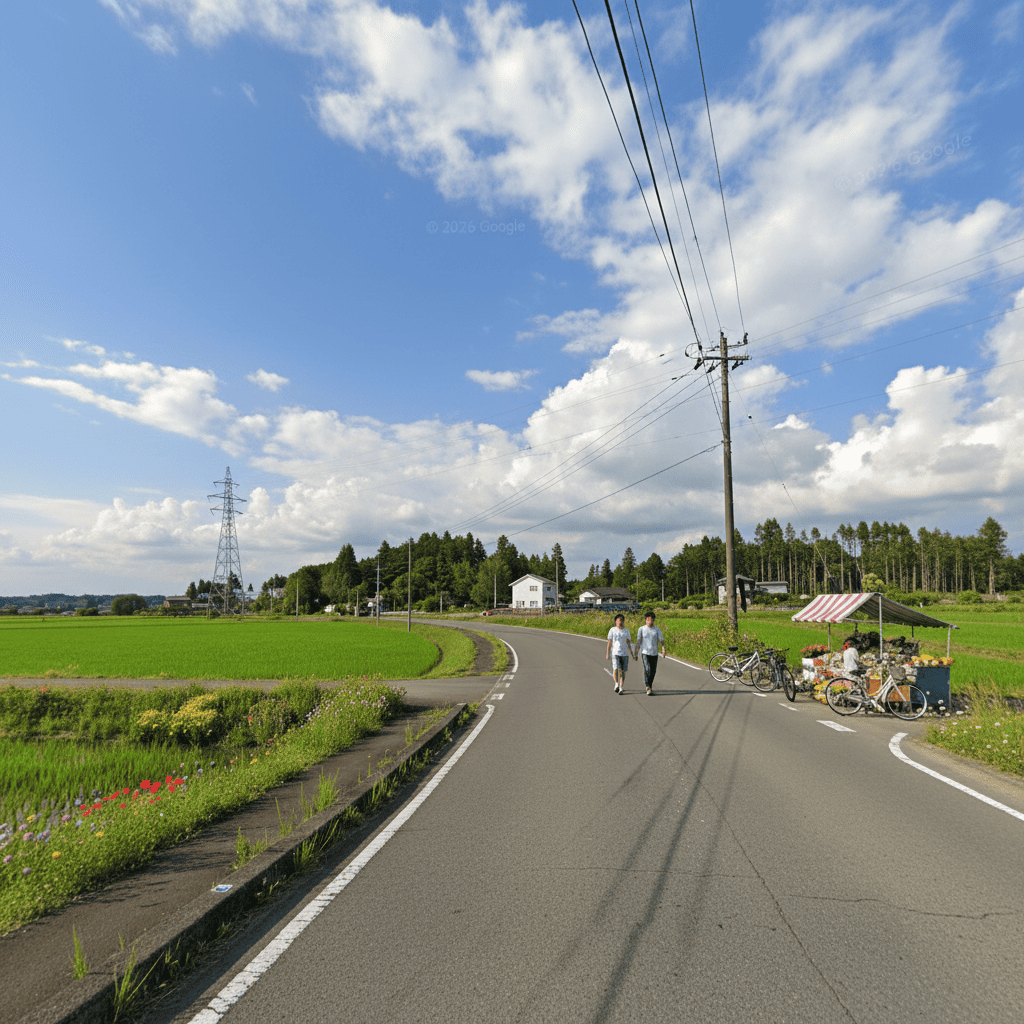}
    {66.7}{62.9}{15.5}
&
\outputcell
    {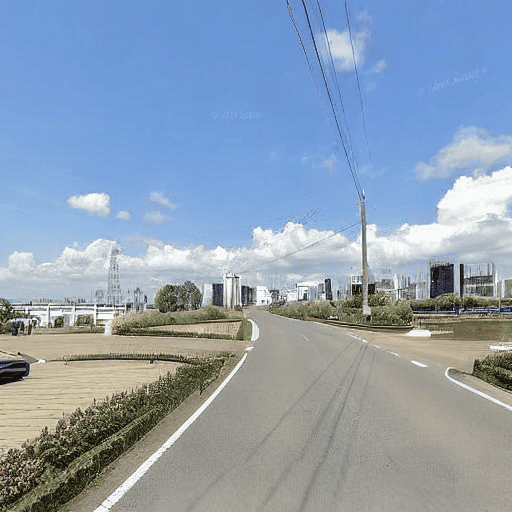}
    {43.3}{56.2}{21.2}
\\
\samplerule


\maxtask{Lively}
&
\inputcell{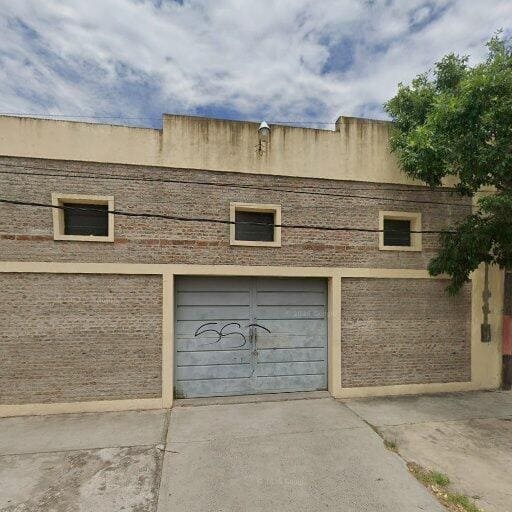}
&
\outputcell
    {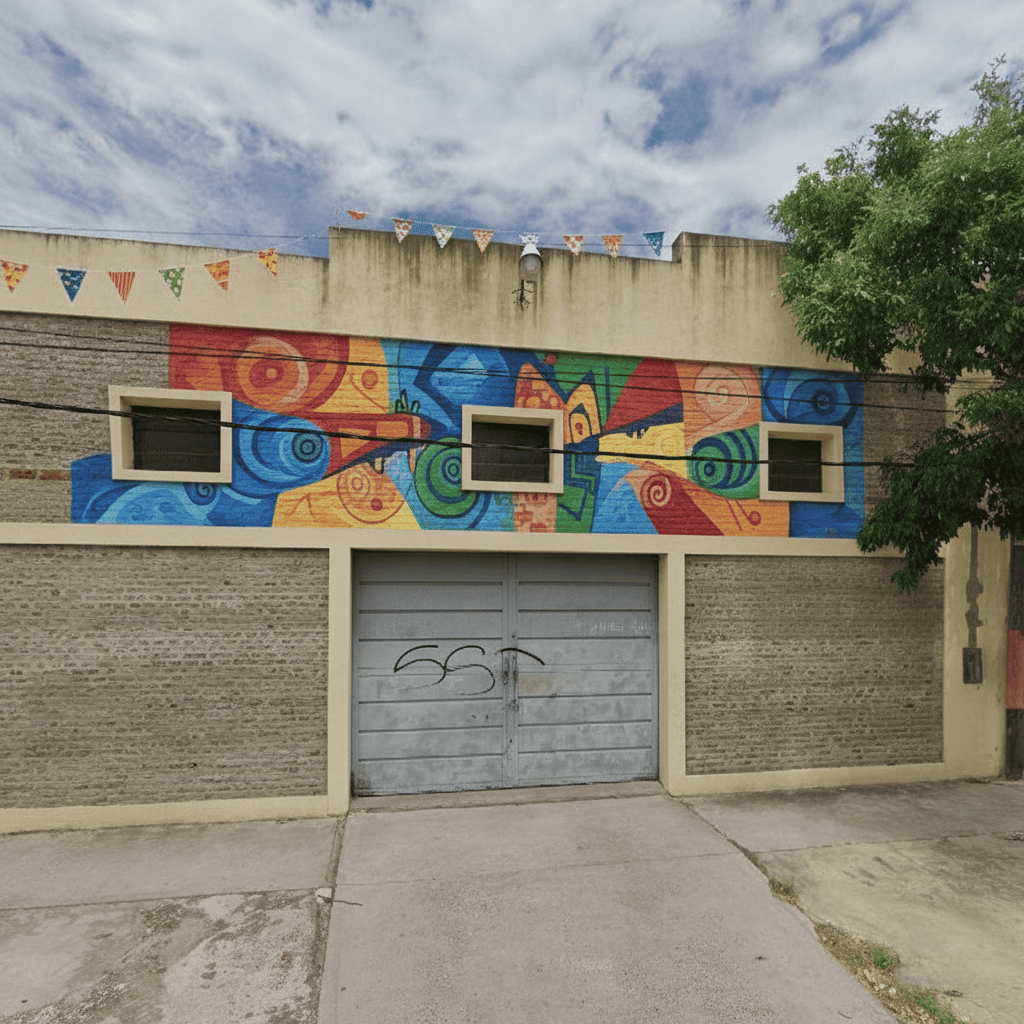}
    {80.0}{81.5}{13.1}
&
\outputcell
    {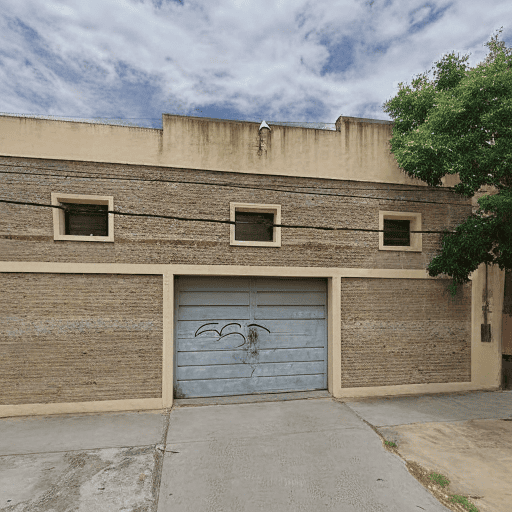}
    {93.3}{66.8}{1.5}
&
\outputcell
    {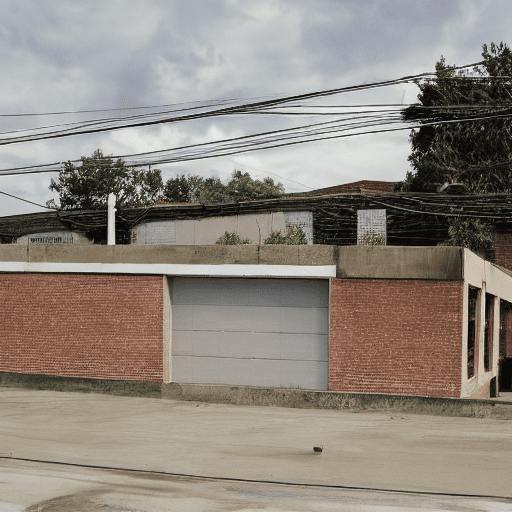}
    {50.0}{50.2}{8.8}
&
\outputcell
    {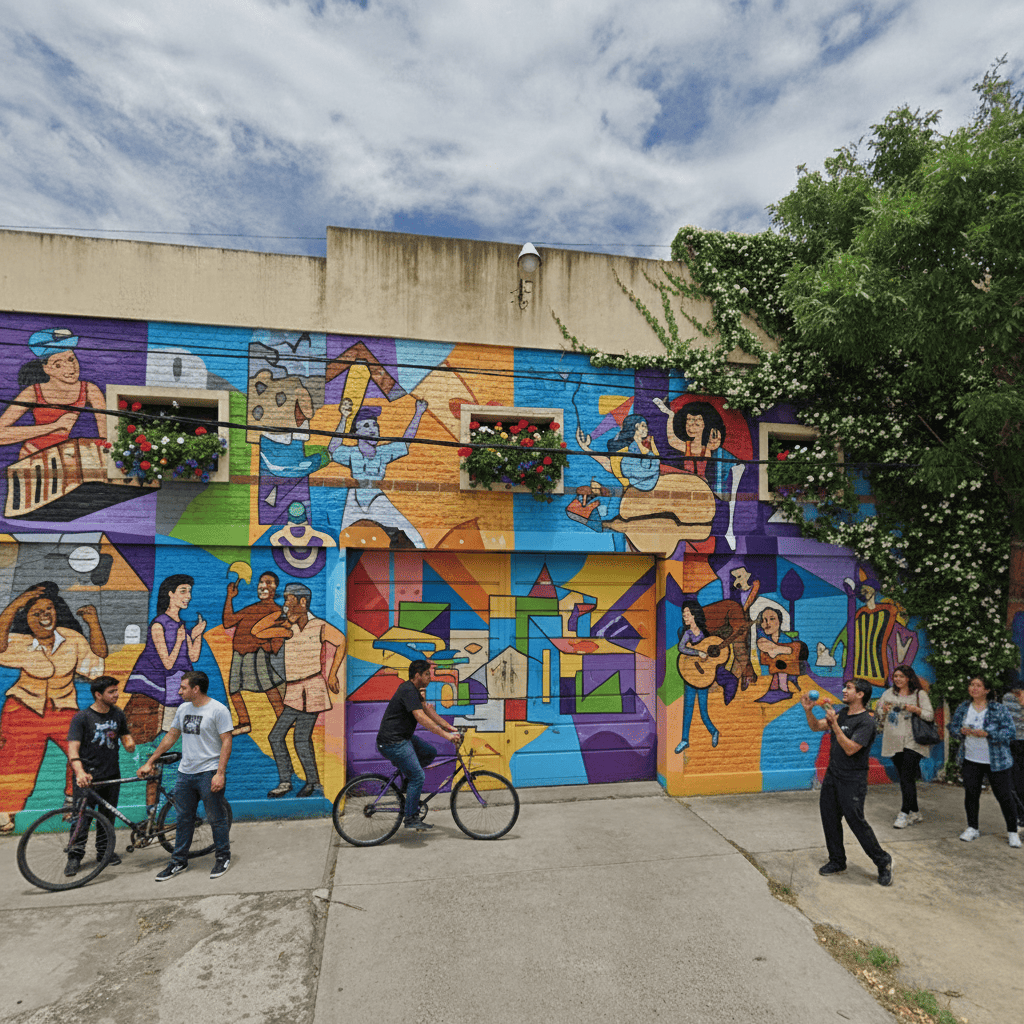}
    {53.3}{87.6}{68.8}
&
\outputcell
    {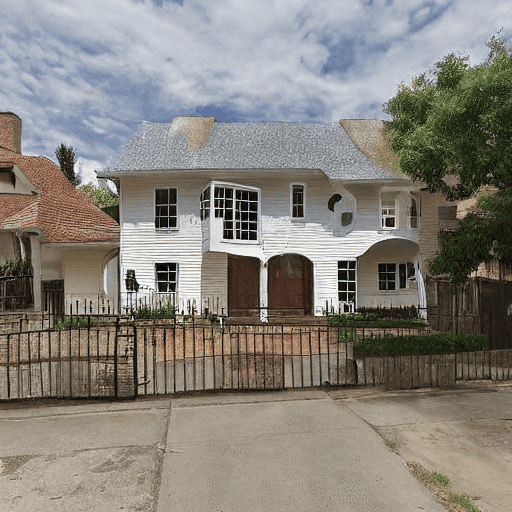}
    {40.0}{59.2}{62.4}
\\
\samplerule


\maxtask{Lively}
&
\inputcell{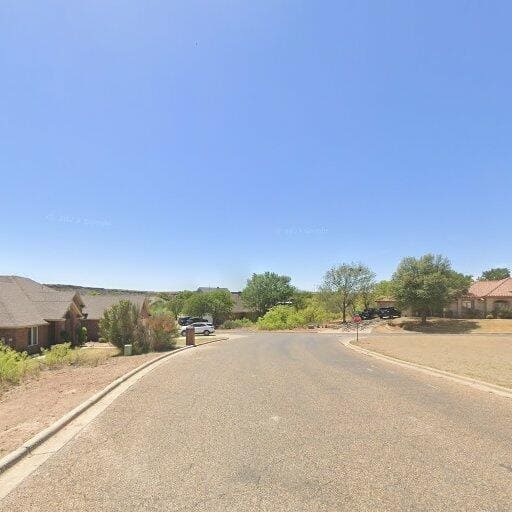}
&
\outputcell
    {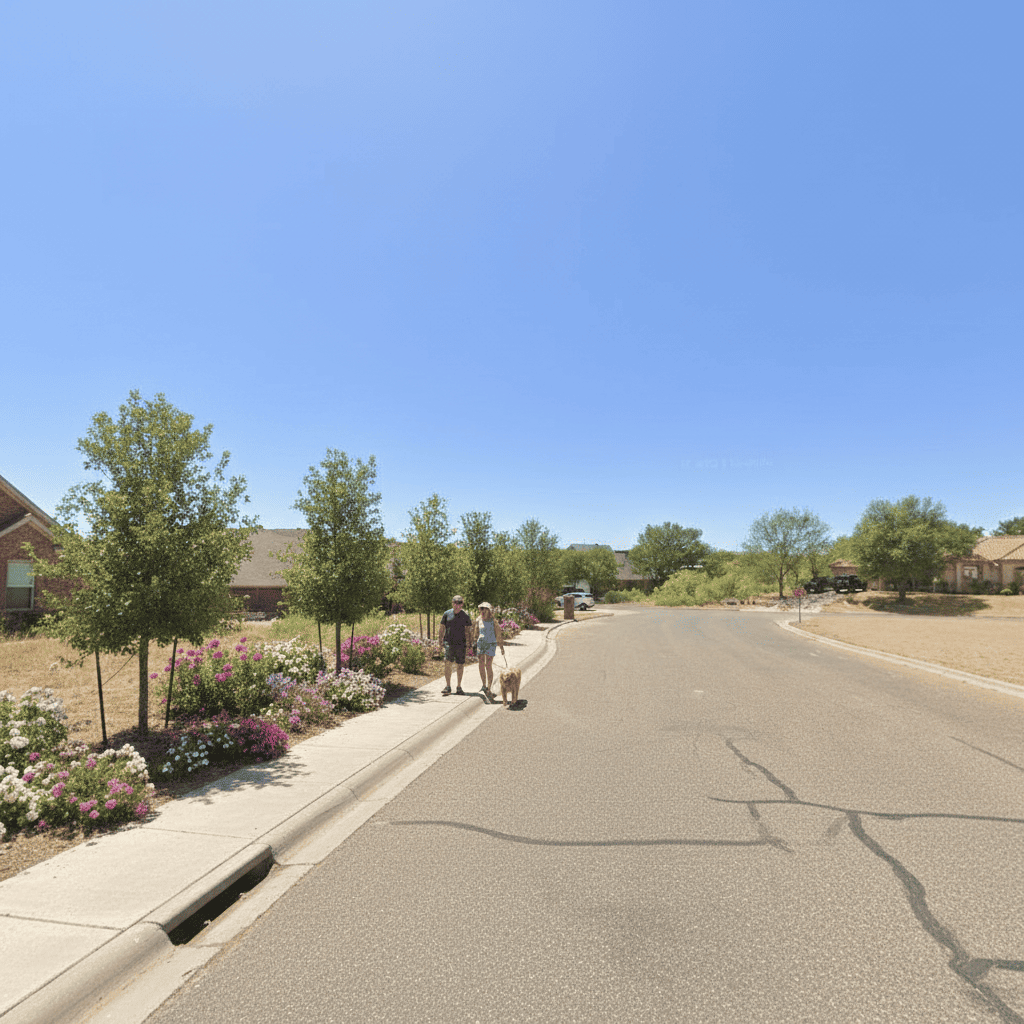}
    {73.3}{73.9}{47.5}
&
\outputcell
    {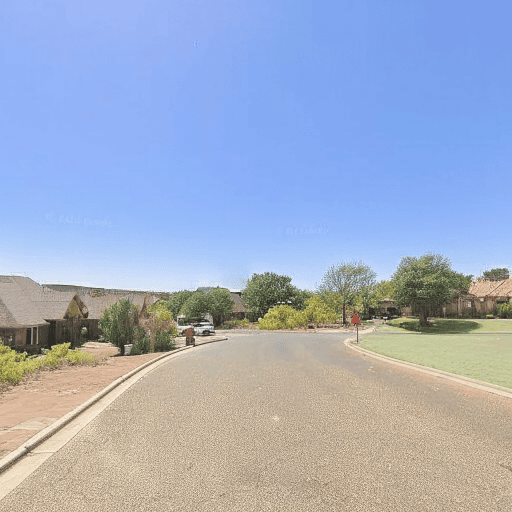}
    {86.7}{69.6}{14.3}
&
\outputcell
    {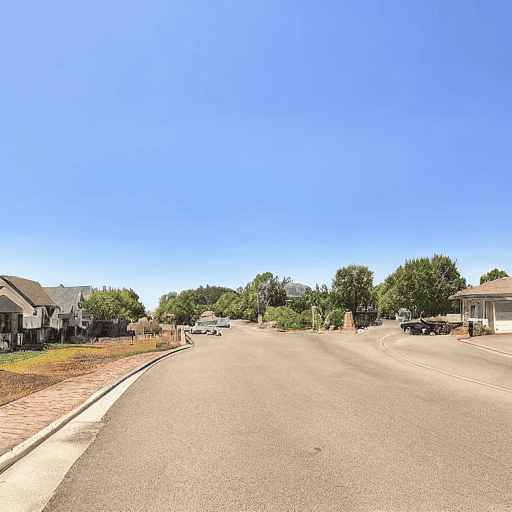}
    {50.0}{74.1}{30.1}
&
\outputcell
    {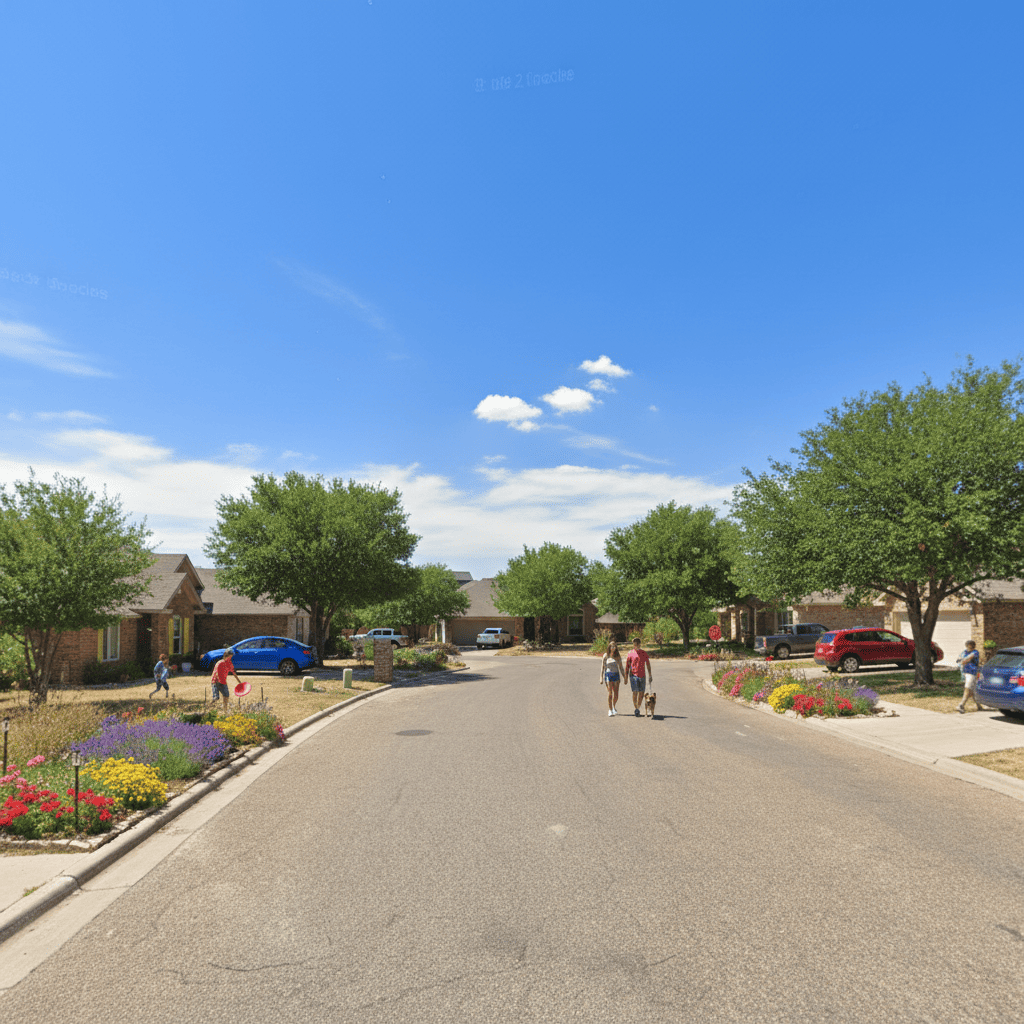}
    {63.3}{72.8}{66.3}
&
\outputcell
    {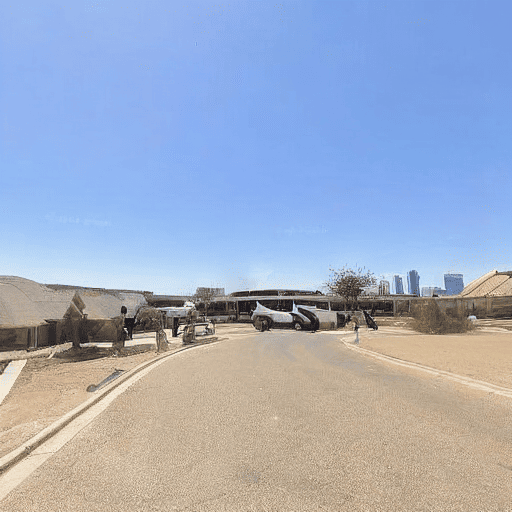}
    {50.0}{62.7}{11.8}
\\
\samplerule


\maxtask{Wealthy}
&
\inputcell{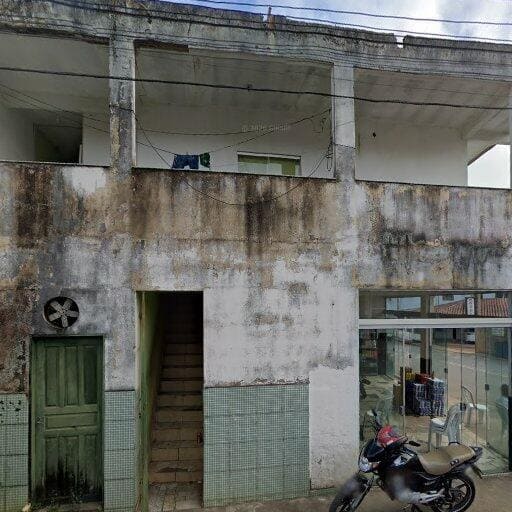}
&
\outputcell
    {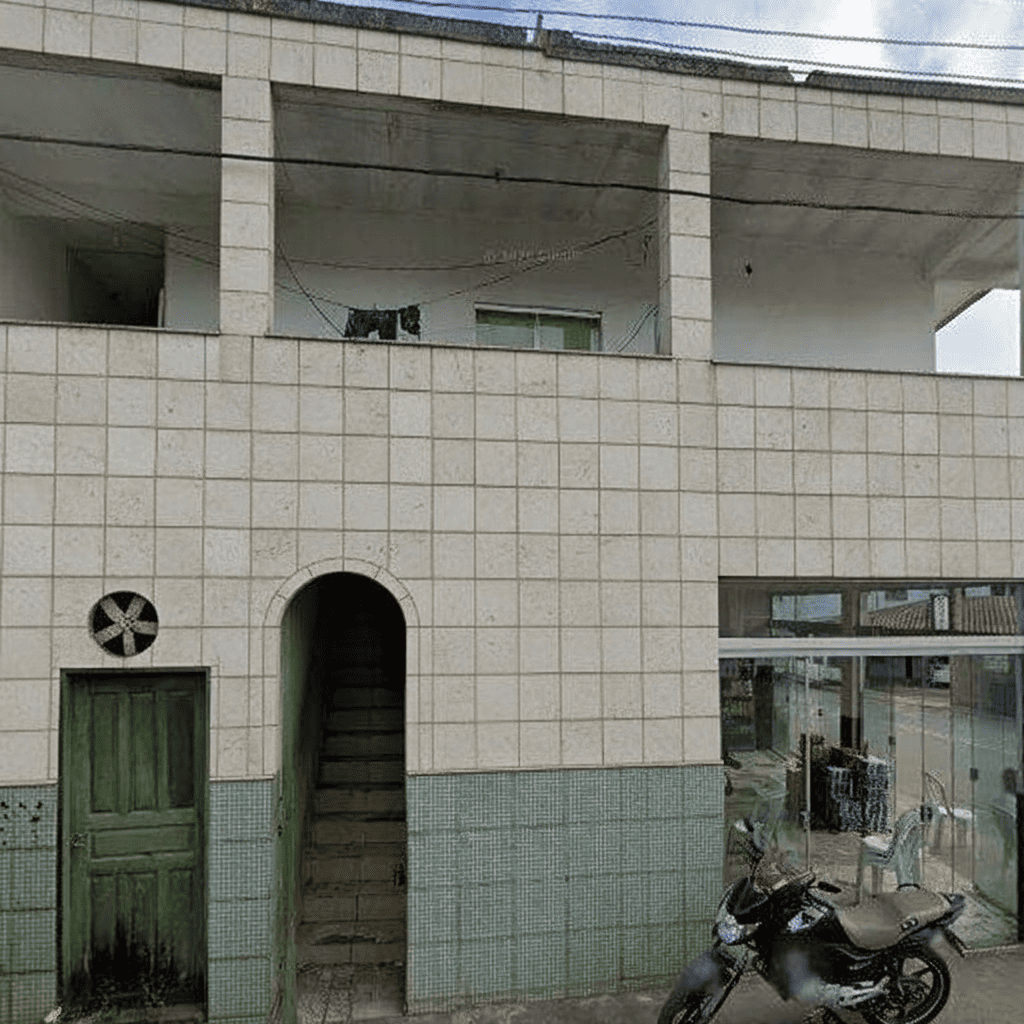}
    {73.3}{69.2}{45.3}
&
\outputcell
    {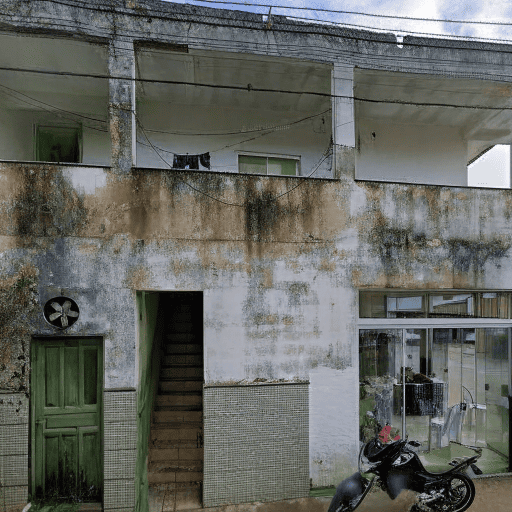}
    {90.0}{69.1}{7.5}
&
\outputcell
    {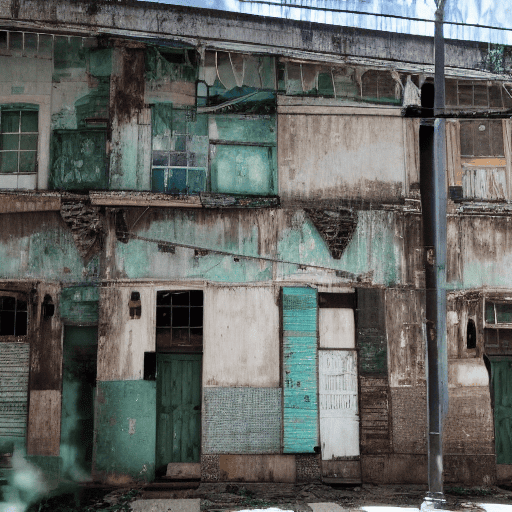}
    {43.3}{65.7}{40.9}
&
\outputcell
    {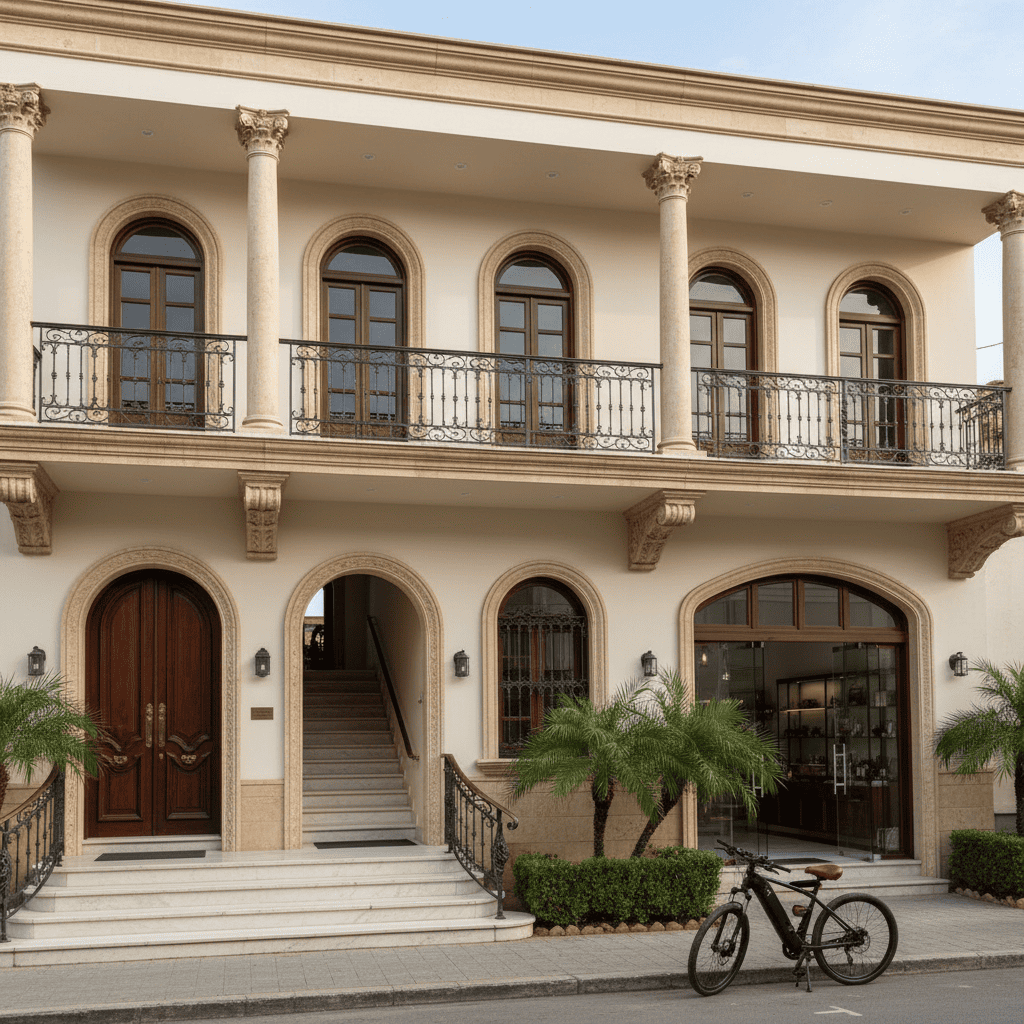}
    {50.0}{77.3}{44.2}
&
\outputcell
    {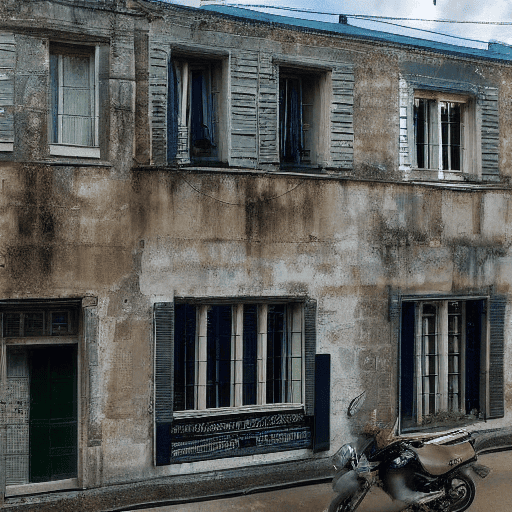}
    {50.0}{82.6}{71.9}
\\

\bottomrule

\end{tabular*}

\end{center}


\newpage


\begin{center}

{\small\textbf{Table \thetable\ continued}}

\vspace{4pt}

\setlength{\tabcolsep}{1pt}
\renewcommand{\arraystretch}{0.92}

\begin{tabular*}{\textwidth}{
@{\extracolsep{\fill}}
c c c c c c c
@{}
}

\toprule

{\scriptsize\textbf{Task}}
&
{\scriptsize\textbf{Input}}
&
{\scriptsize\textbf{\vida}}
&
{\scriptsize\textbf{DIFFusion}}
&
{\scriptsize\textbf{LANCE}}
&
{\scriptsize\textbf{NB2.5}}
&
{\scriptsize\textbf{URSimulator}}
\\[-2pt]

&
&
\multicolumn{5}{c}{
    {\tiny\textit{
        LLM-Judge $\uparrow$, FID $\downarrow$, Gain $\uparrow$
    }}
}
\\[-2pt]

\midrule


\mintask{Depressing}
&
\inputcell{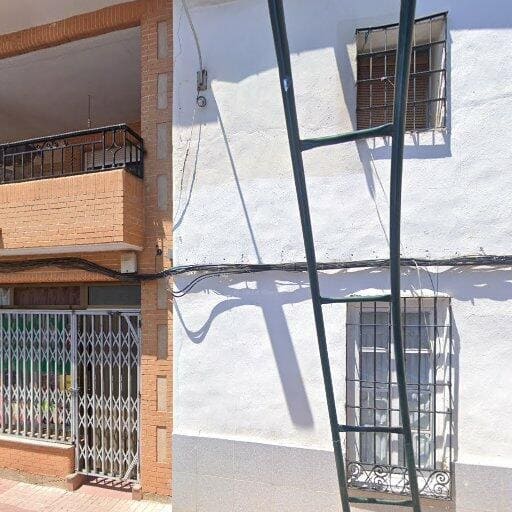}
&
\outputcell
    {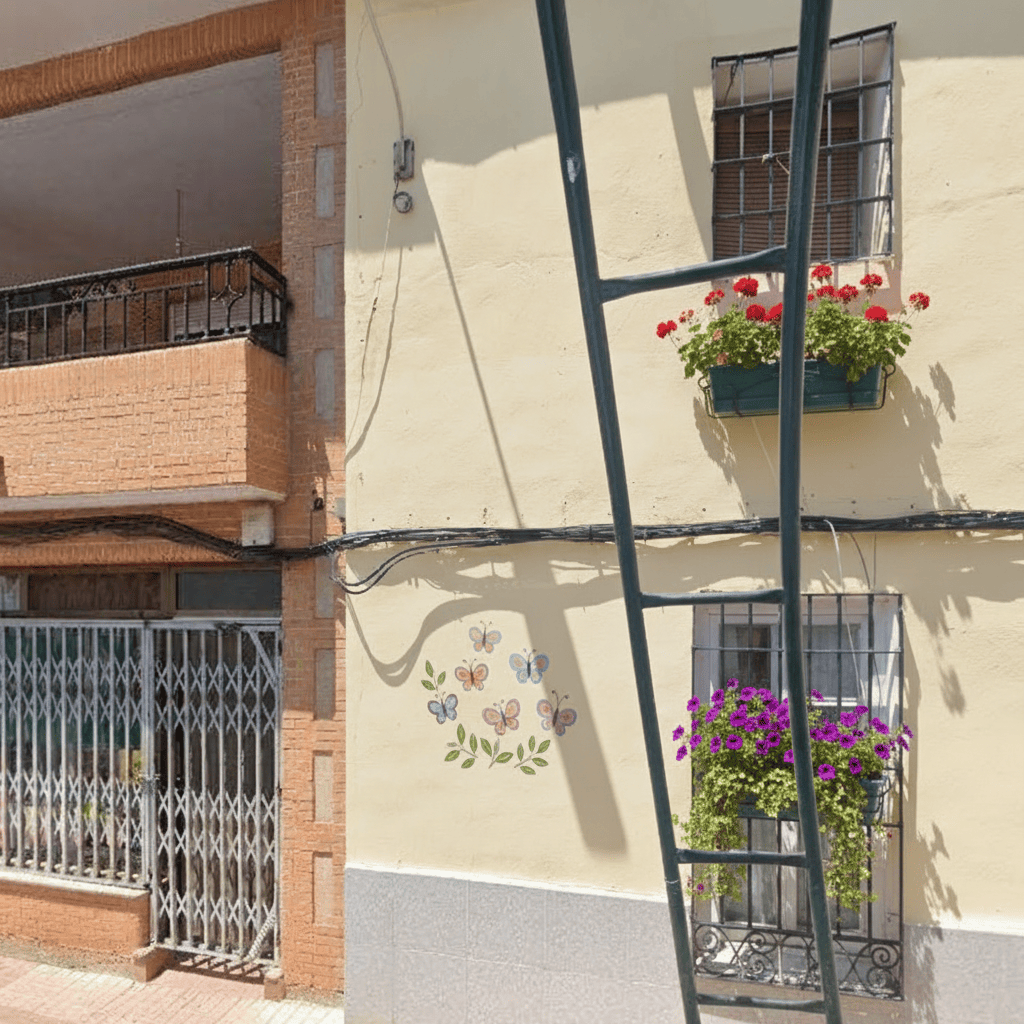}
    {76.7}{67.7}{60.9}
&
\outputcell
    {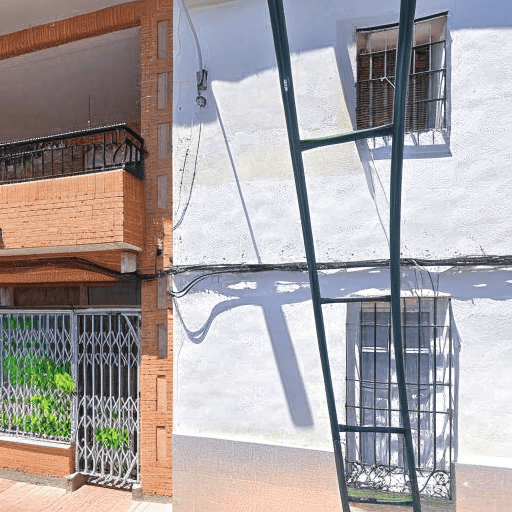}
    {83.3}{69.5}{35.9}
&
\outputcell
    {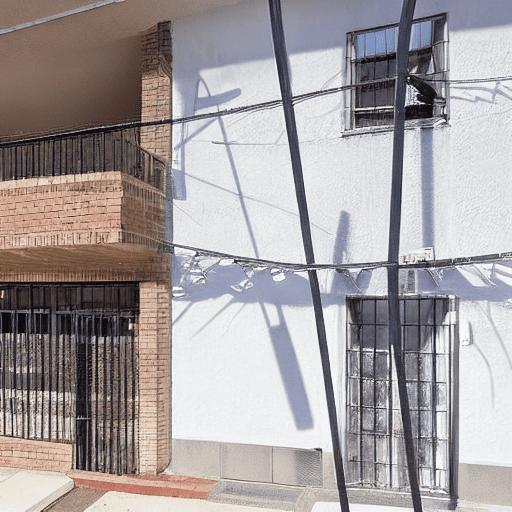}
    {63.3}{65.0}{31.3}
&
\outputcell
    {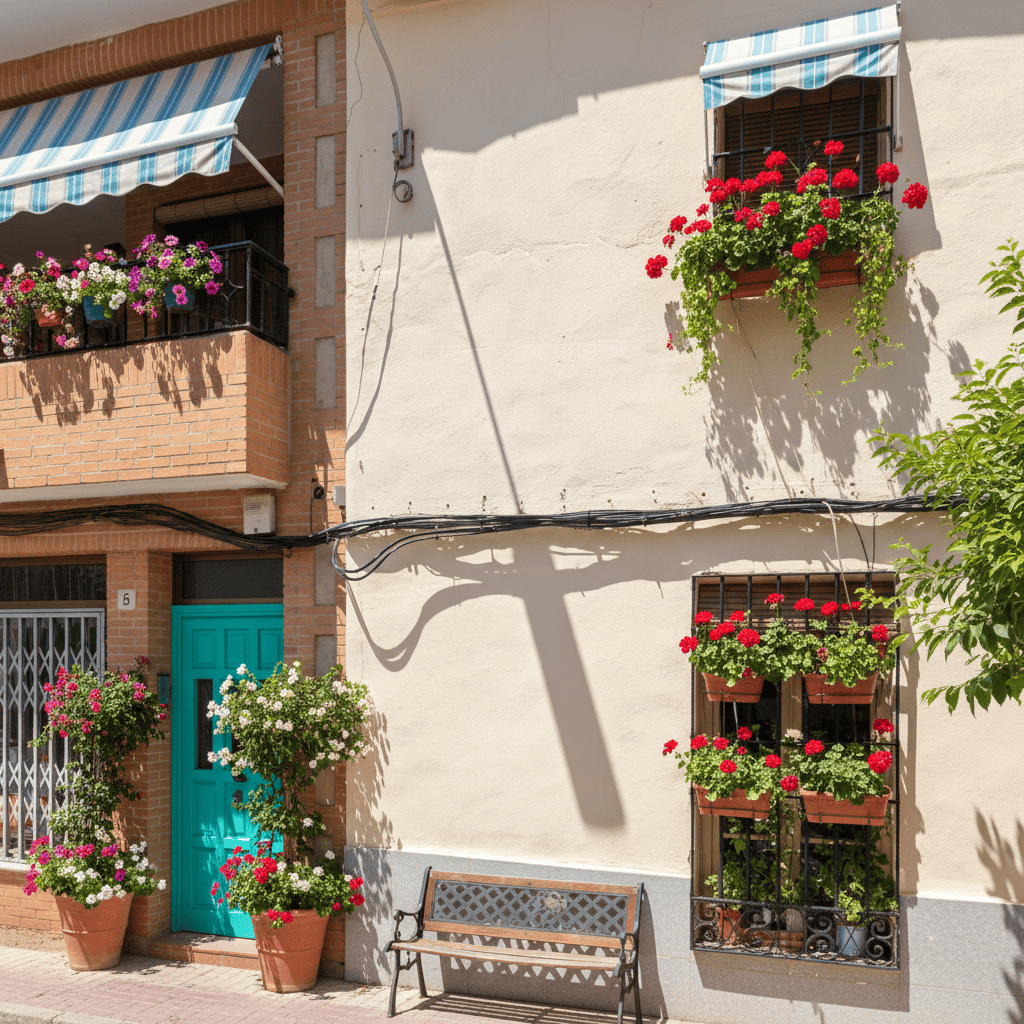}
    {70.0}{87.0}{48.9}
&
\outputcell
    {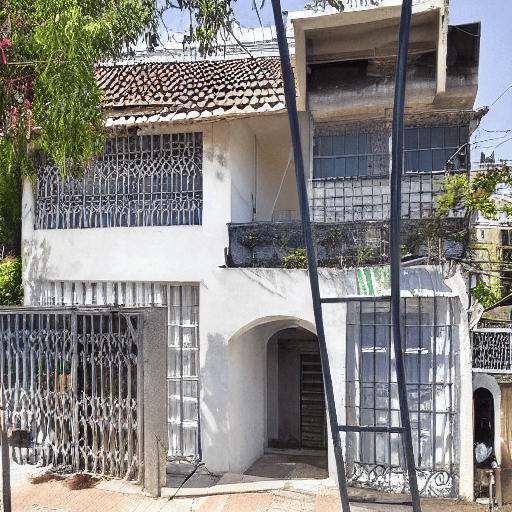}
    {40.0}{62.8}{18.7}
\\
\samplerule


\mintask{Depressing}
&
\inputcell{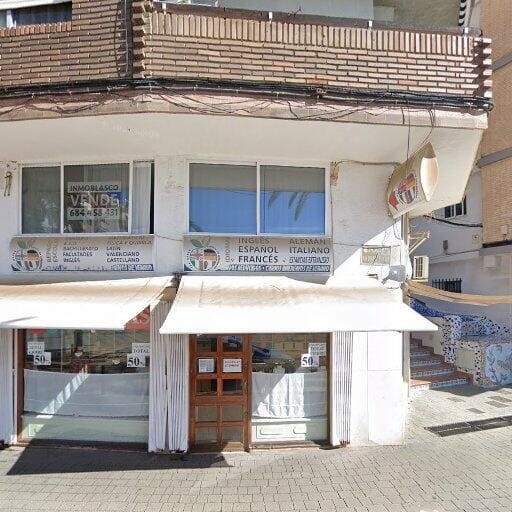}
&
\outputcell
    {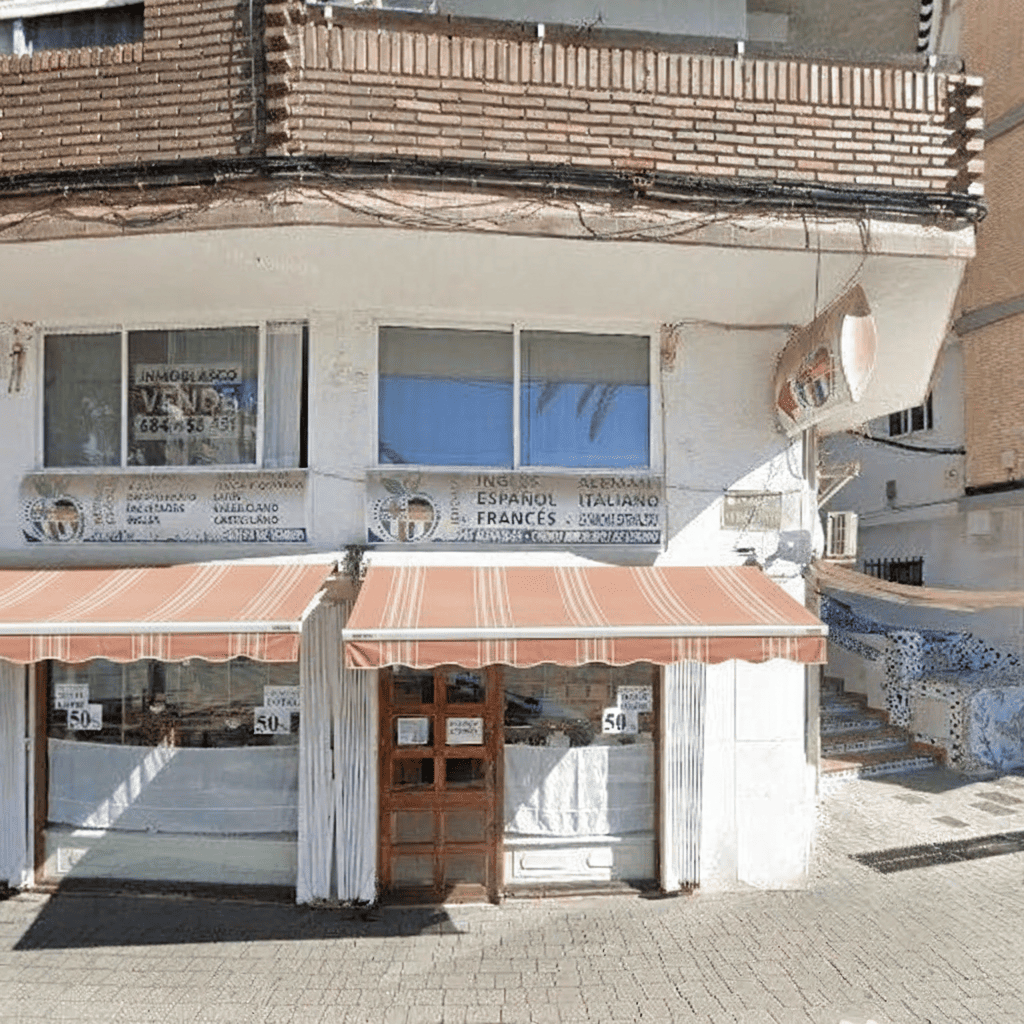}
    {80.0}{61.7}{25.2}
&
\outputcell
    {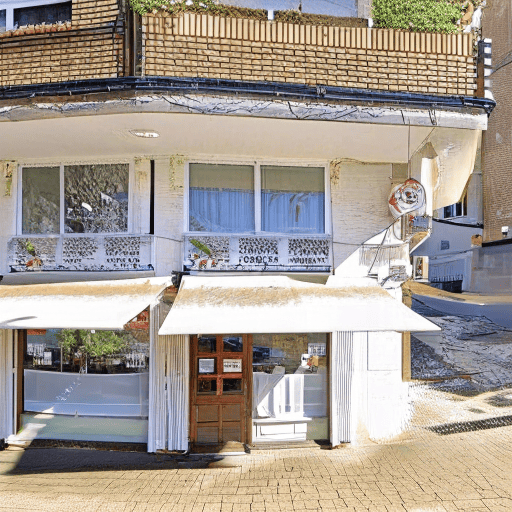}
    {63.3}{65.6}{28.7}
&
\outputcell
    {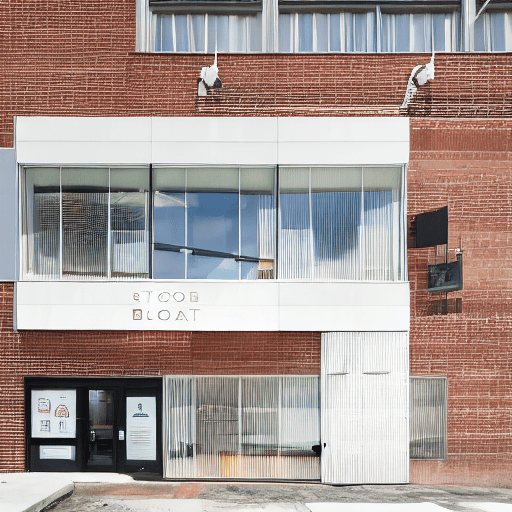}
    {46.7}{71.3}{43.8}
&
\outputcell
    {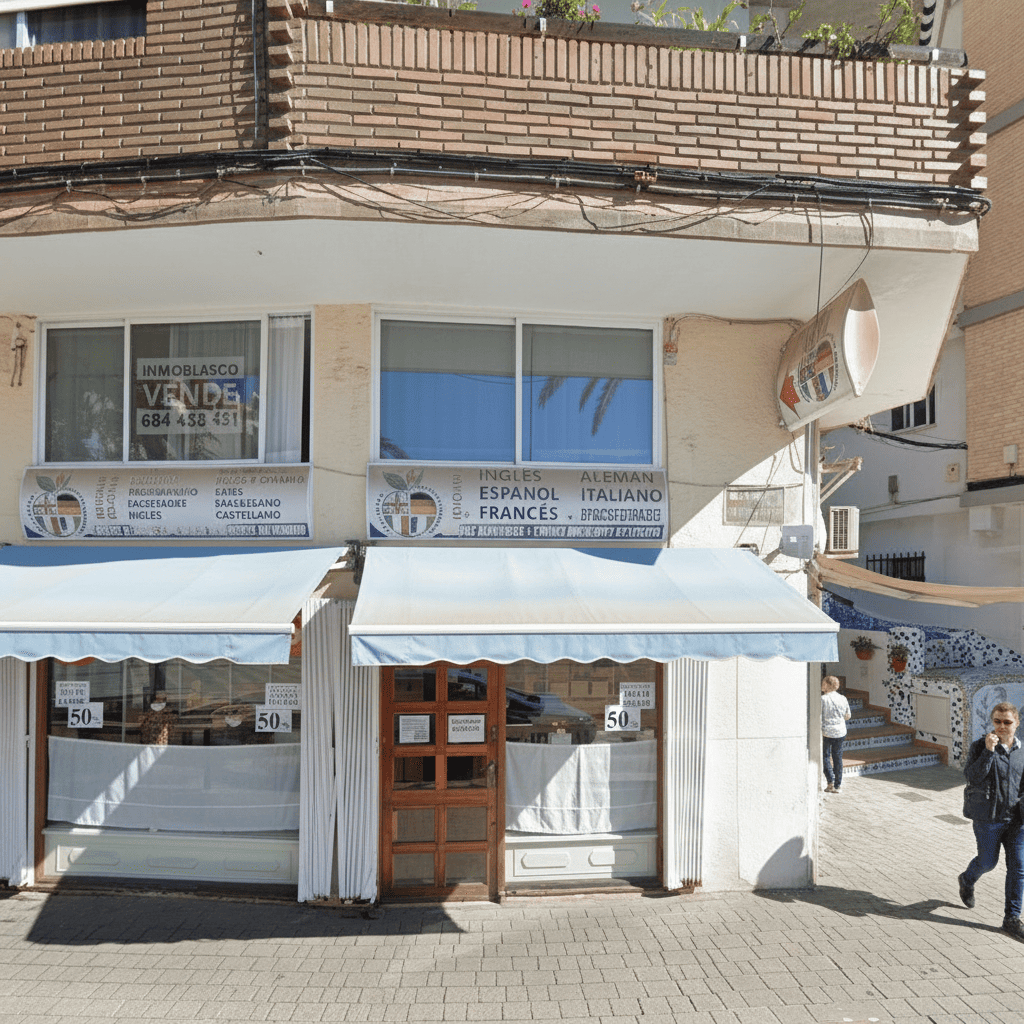}
    {76.7}{58.3}{13.6}
&
\outputcell
    {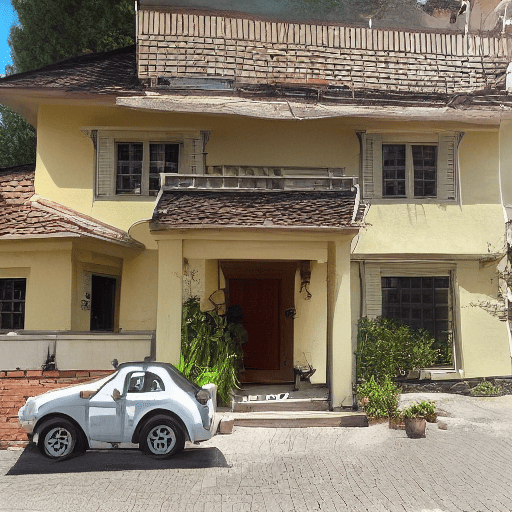}
    {33.3}{70.5}{38.5}
\\
\samplerule


\mintask{Boring}
&
\inputcell{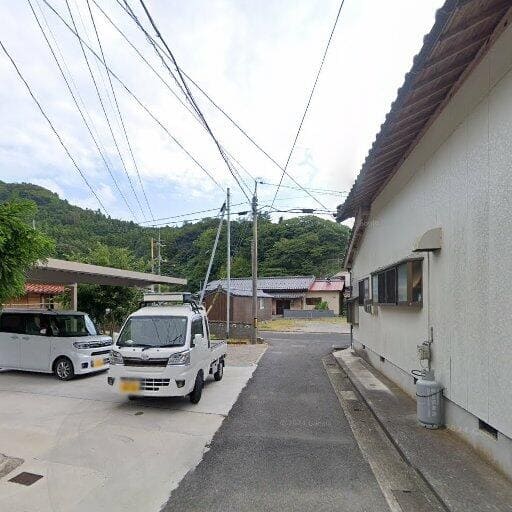}
&
\outputcell
    {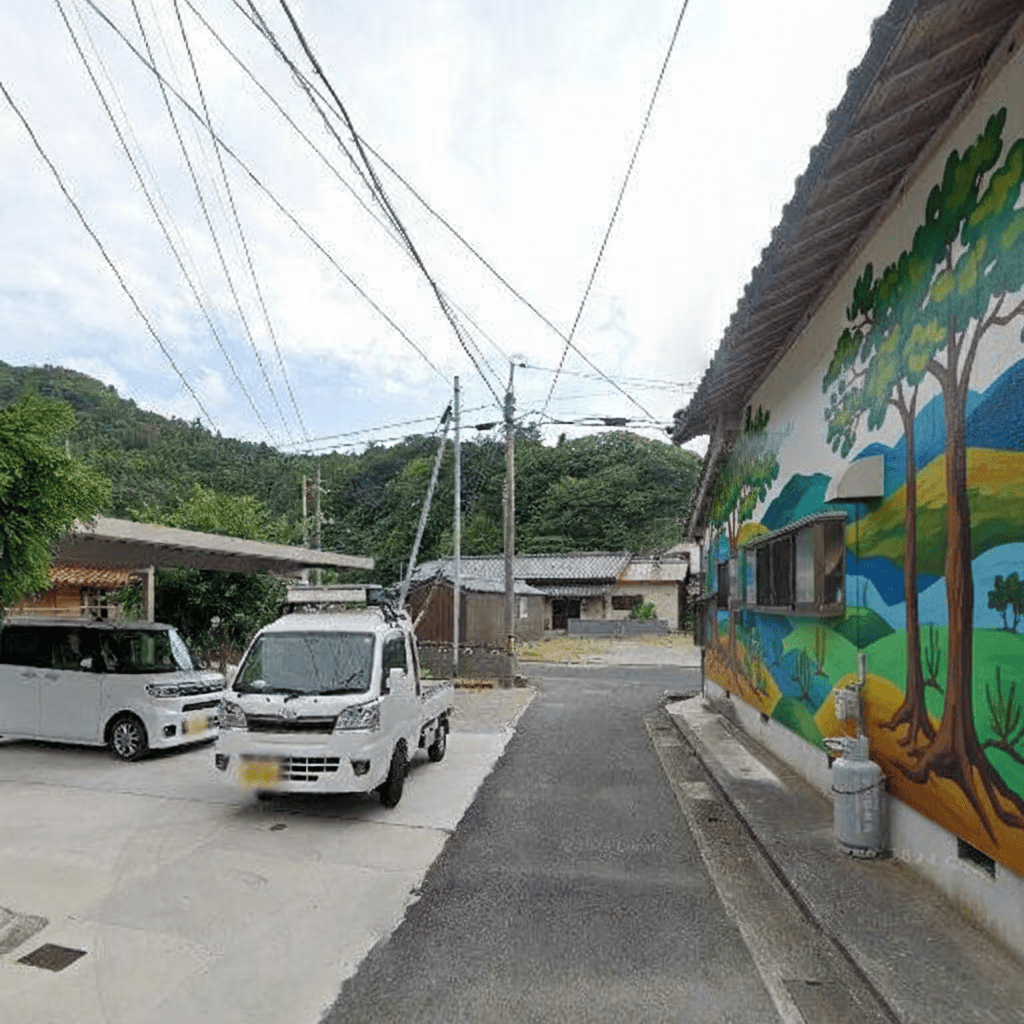}
    {80.0}{46.3}{26.9}
&
\outputcell
    {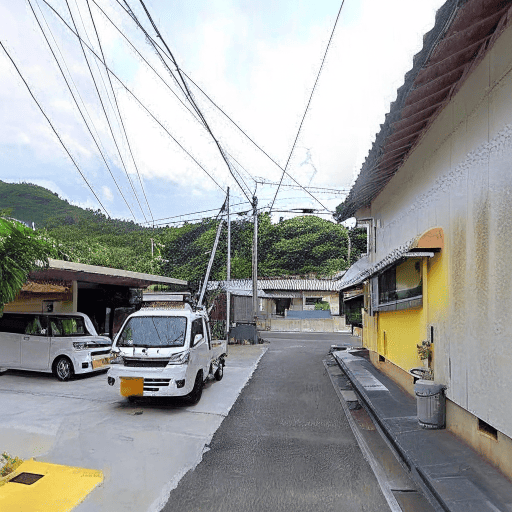}
    {80.0}{47.1}{15.9}
&
\outputcell
    {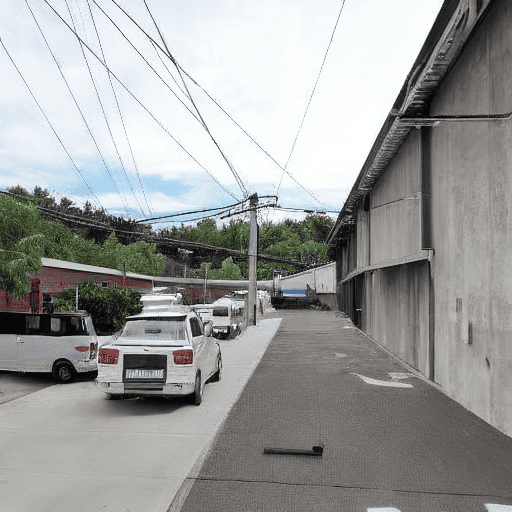}
    {46.7}{60.1}{32.9}
&
\outputcell
    {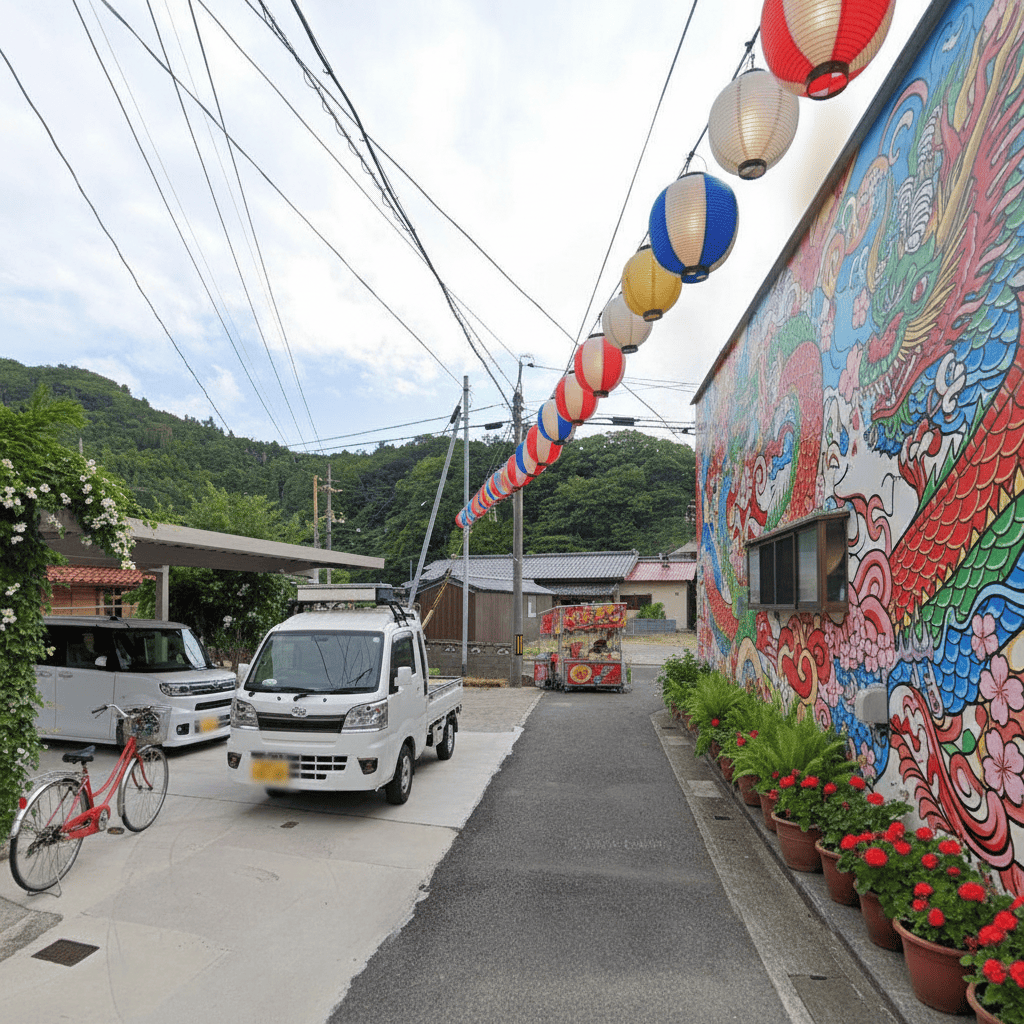}
    {56.7}{60.5}{35.5}
&
\outputcell
    {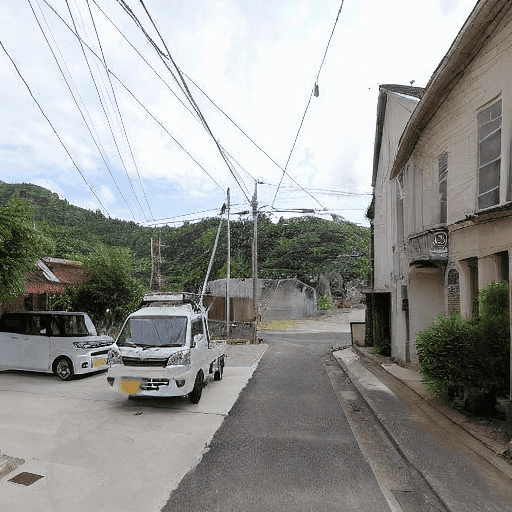}
    {63.3}{48.2}{7.5}
\\
\samplerule


\maxtask{Greenery}
&
\inputcell{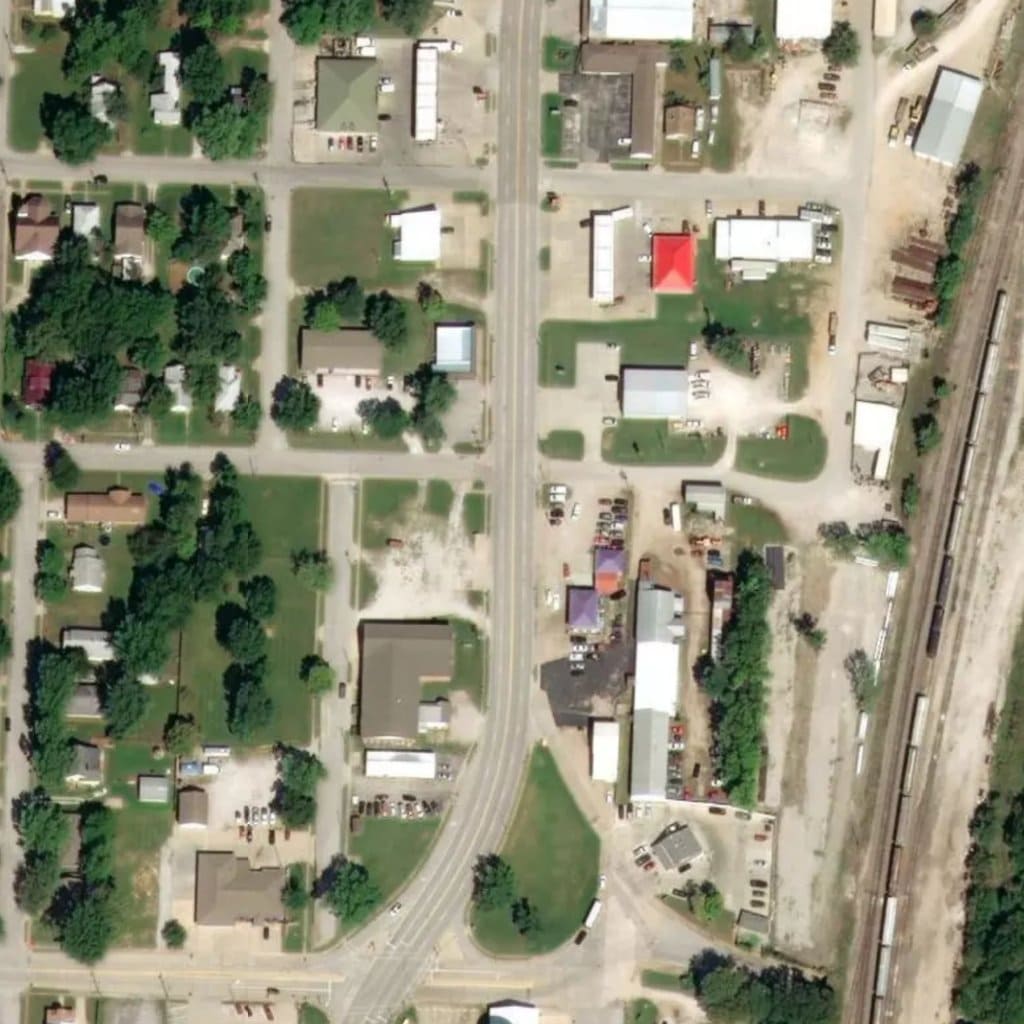}
&
\outputcell
    {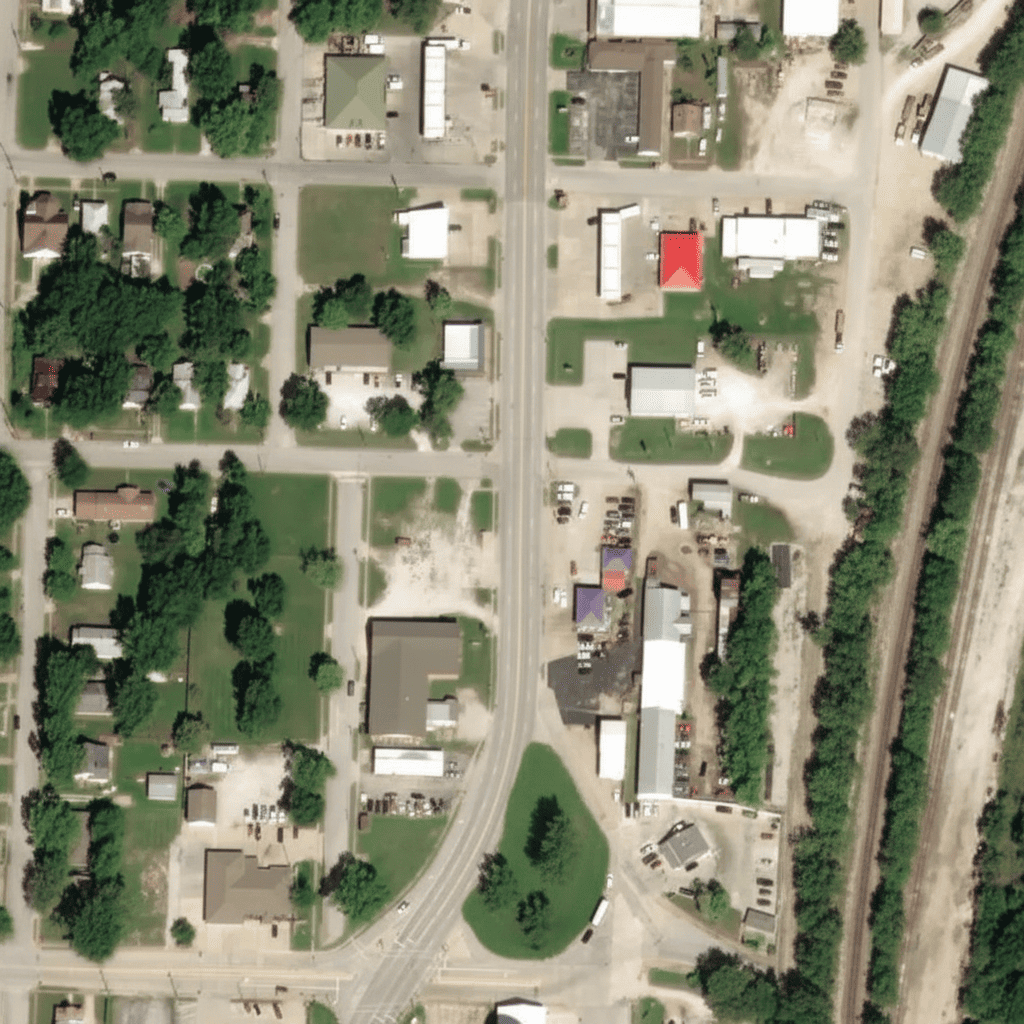}
    {90.0}{45.8}{4.8}
&
\outputcell
    {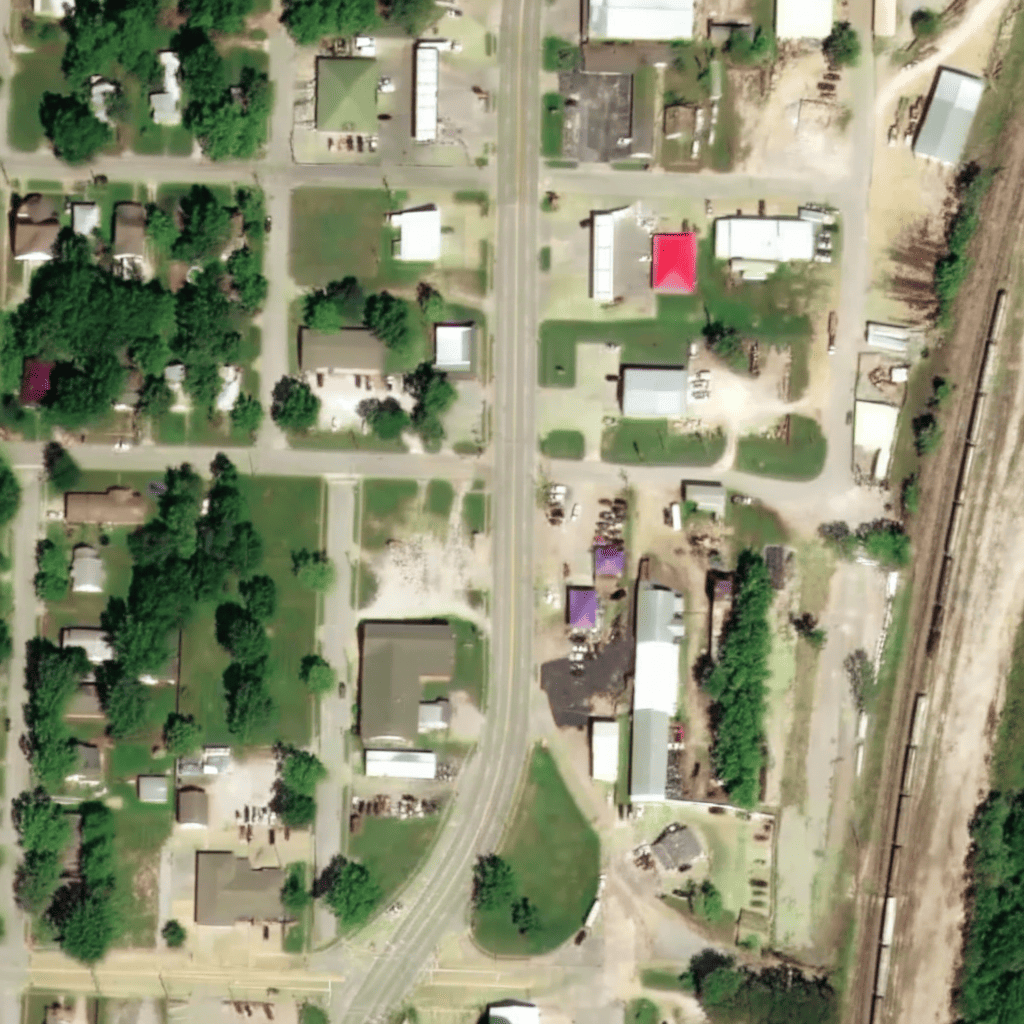}
    {86.7}{43.0}{5.5}
&
\outputcell
    {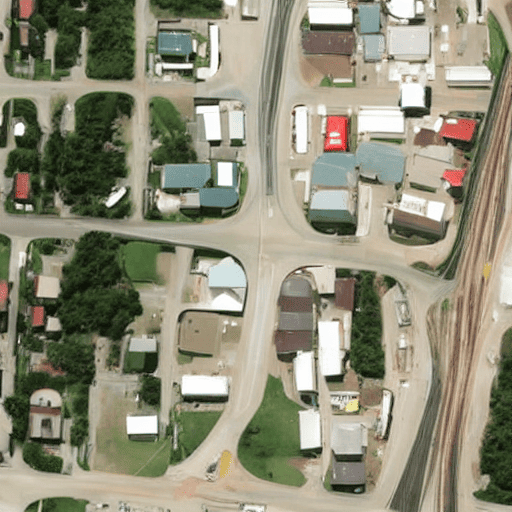}
    {46.7}{52.6}{-12.5}
&
\outputcell
    {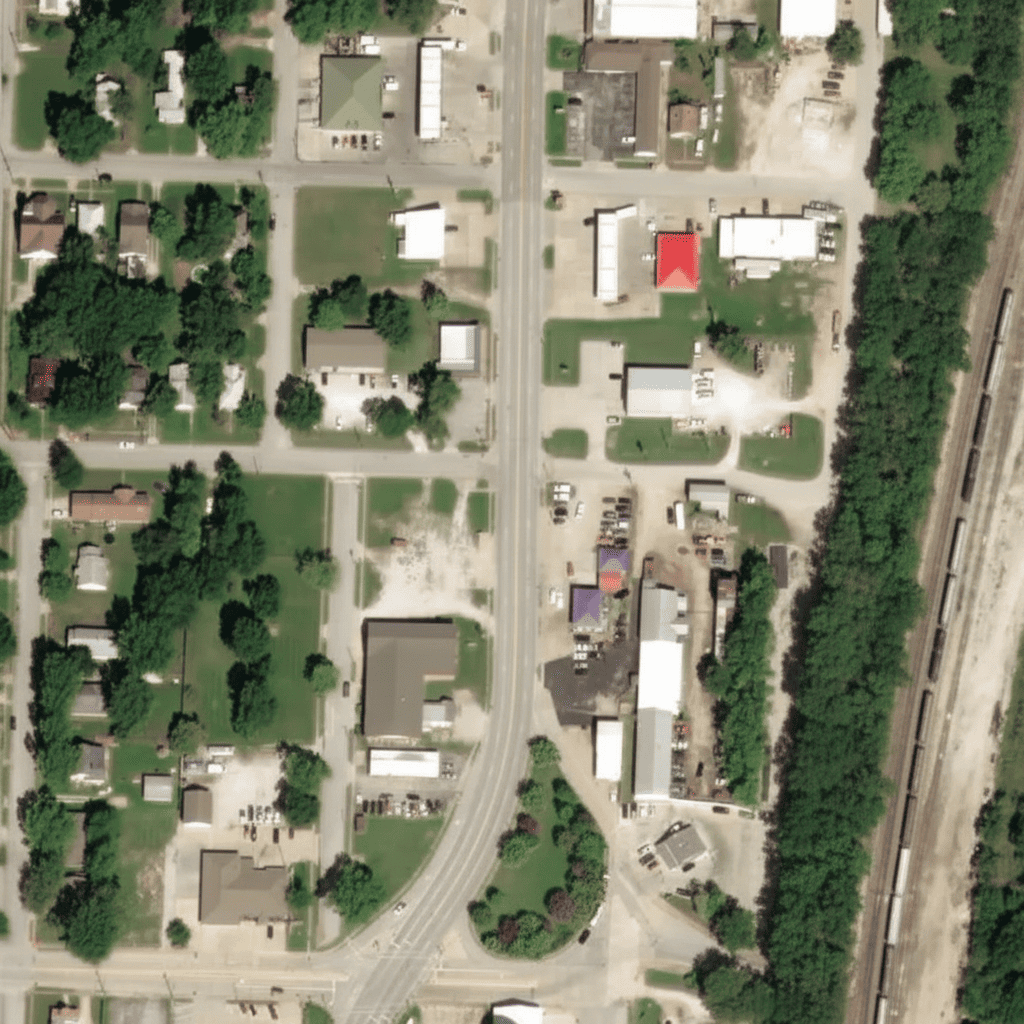}
    {83.3}{40.3}{8.7}
&
\nacell
\\
\samplerule


\mintask{Road Risk}
&
\inputcell{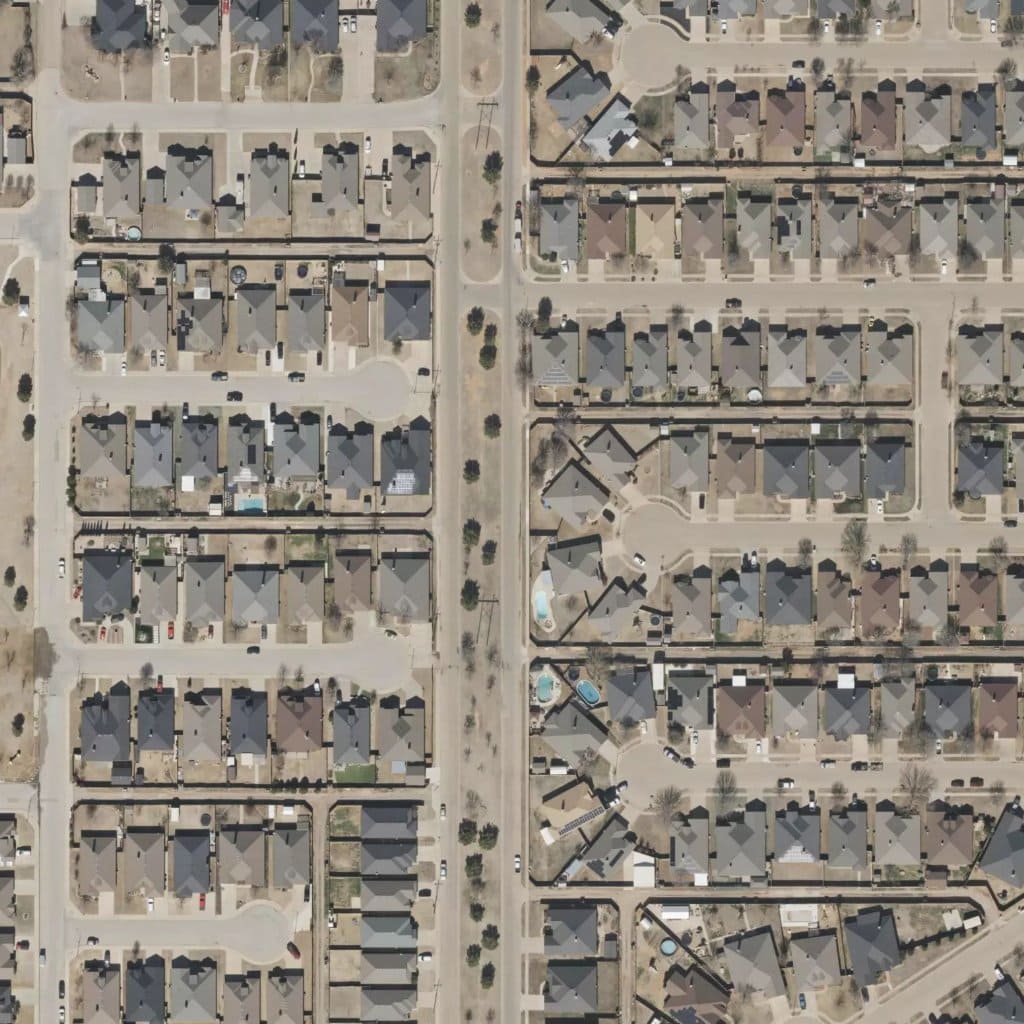}
&
\outputcell
    {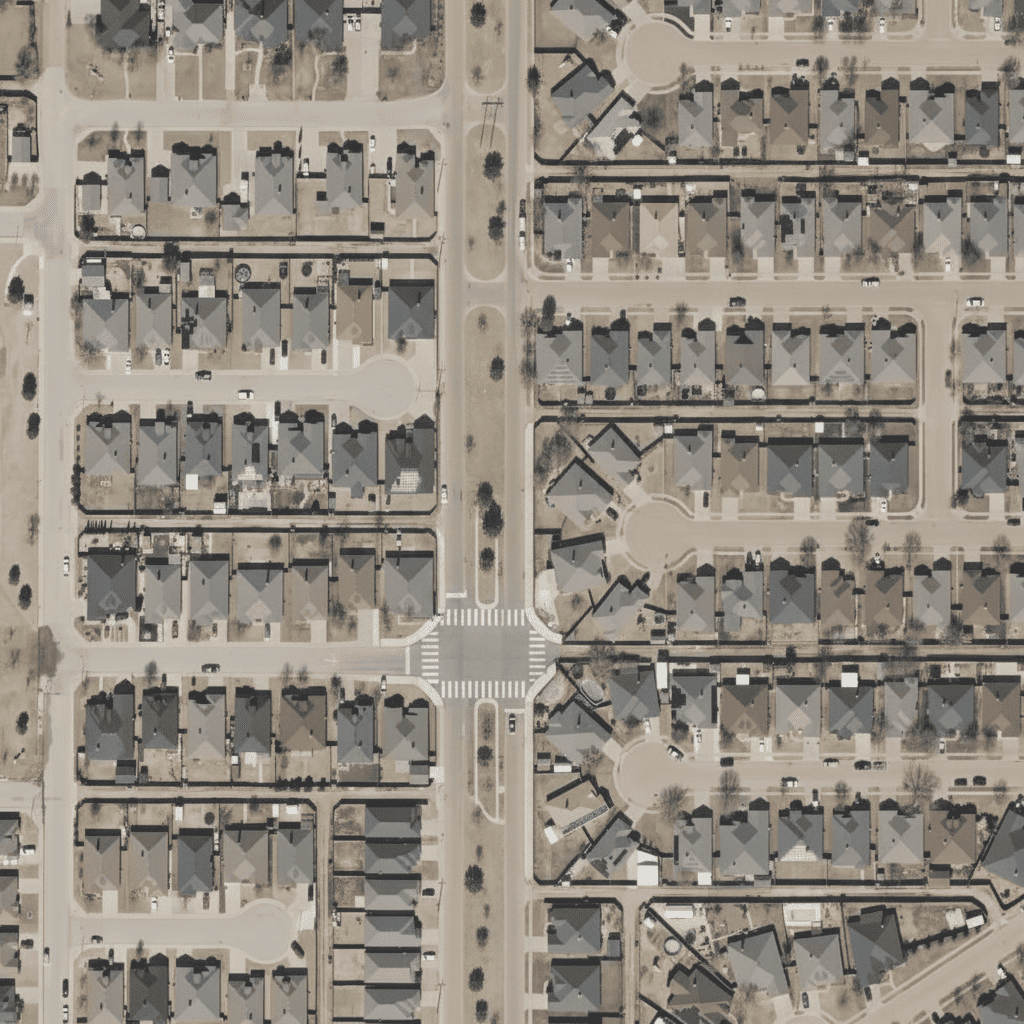}
    {80.0}{47.7}{16.7}
&
\outputcell
    {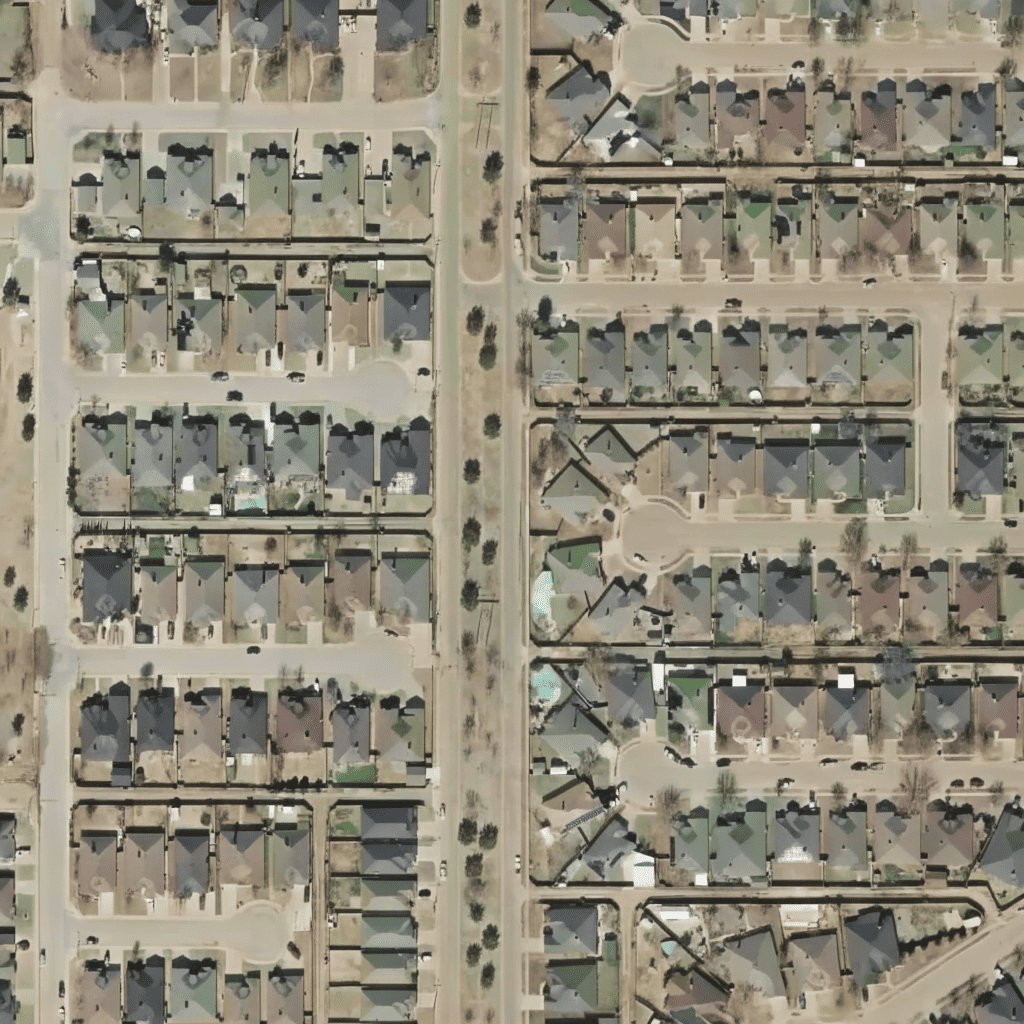}
    {73.3}{46.6}{10.9}
&
\outputcell
    {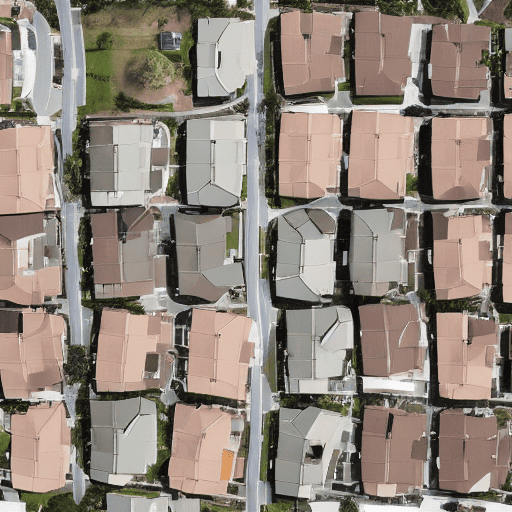}
    {36.7}{82.8}{2.4}
&
\outputcell
    {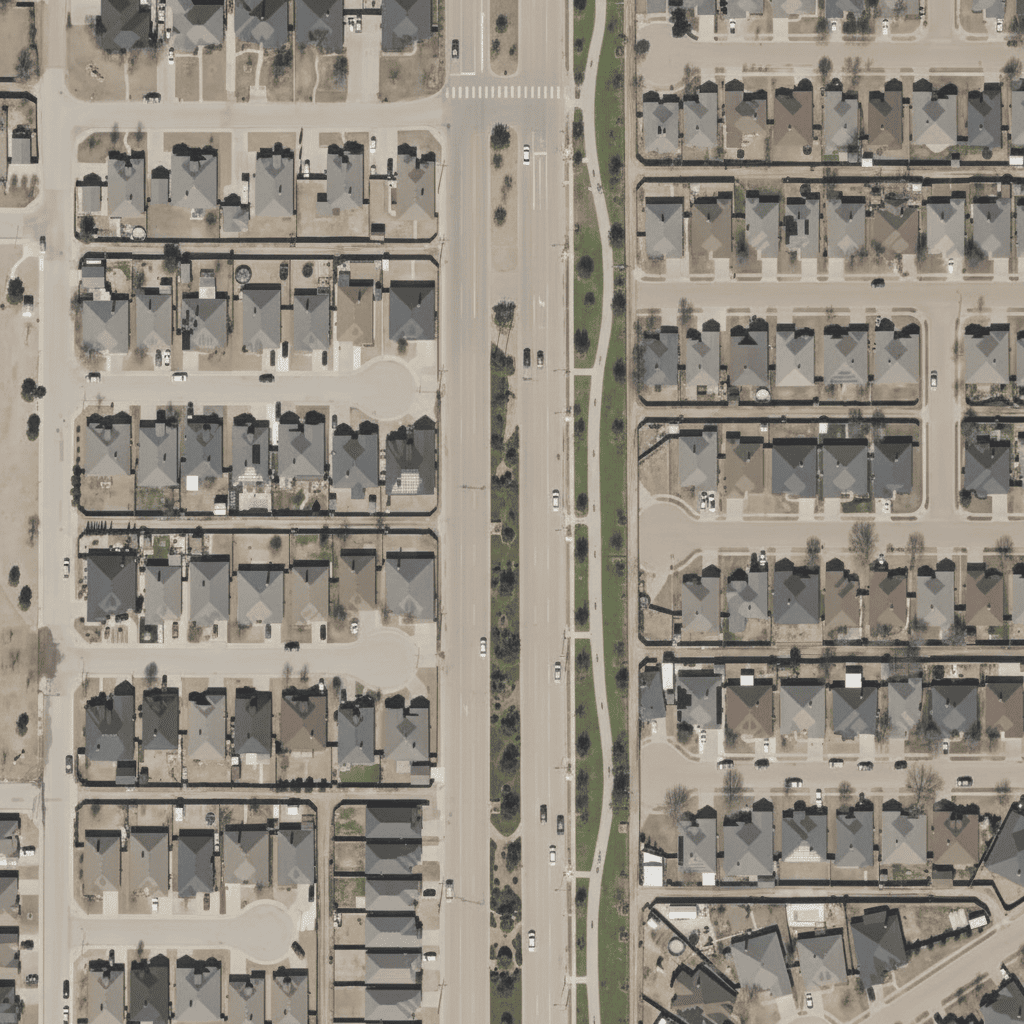}
    {66.7}{51.2}{27.8}
&
\nacell
\\
\samplerule


\mintask{Road Risk}
&
\inputcell{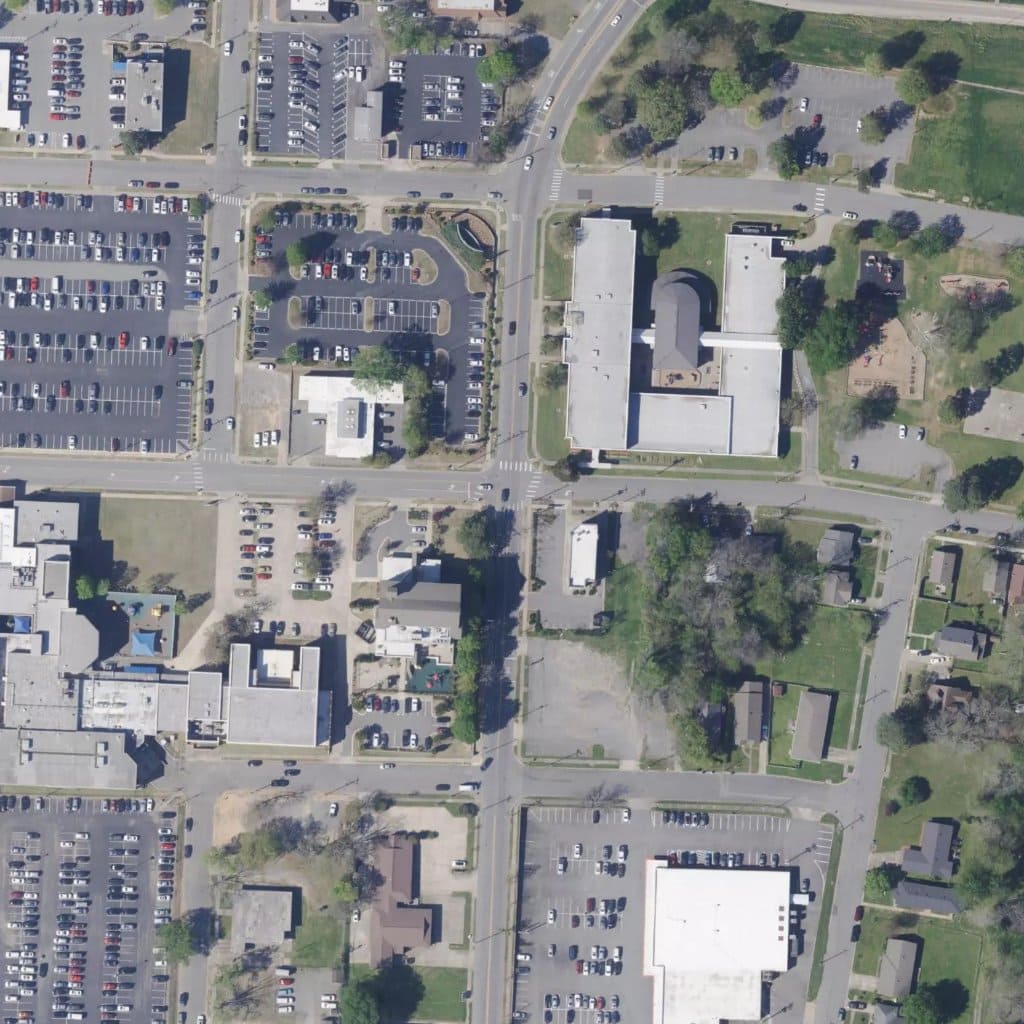}
&
\outputcell
    {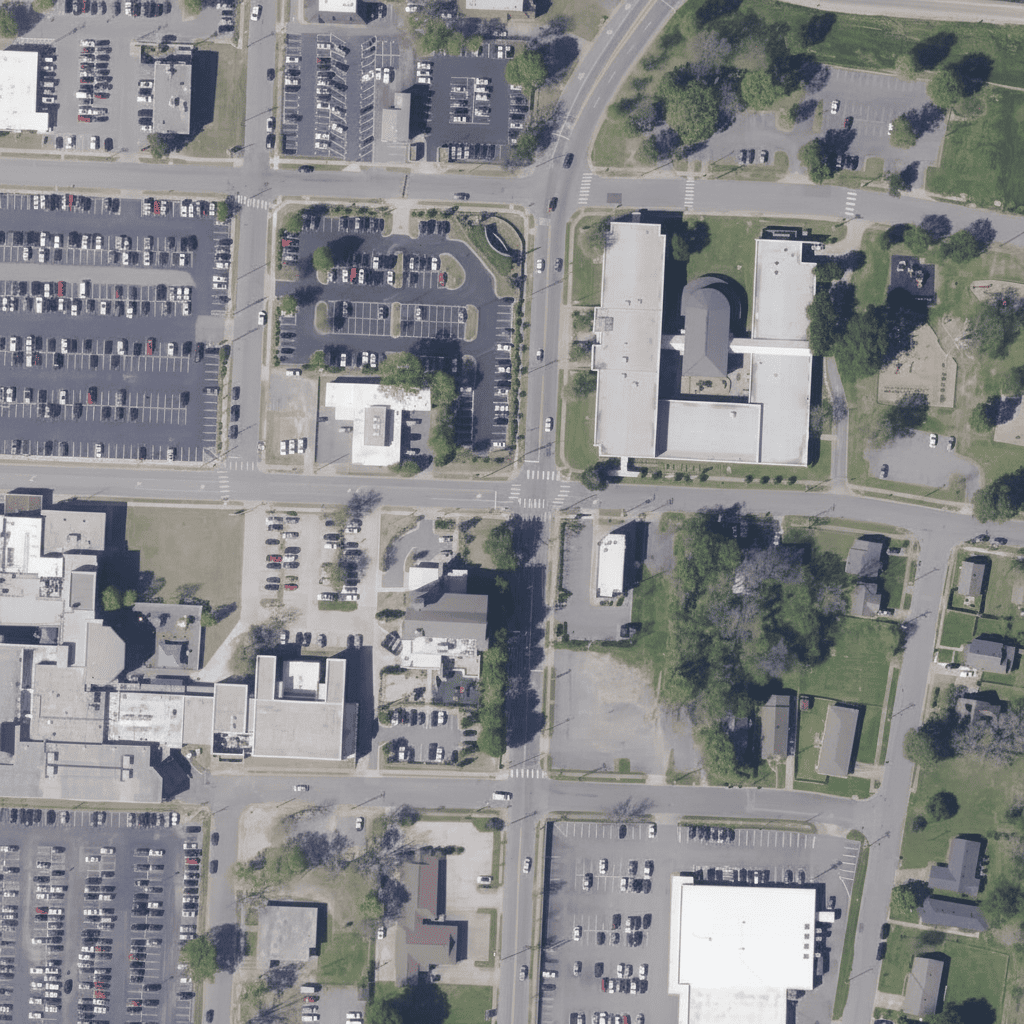}
    {90.0}{37.3}{42.9}
&
\outputcell
    {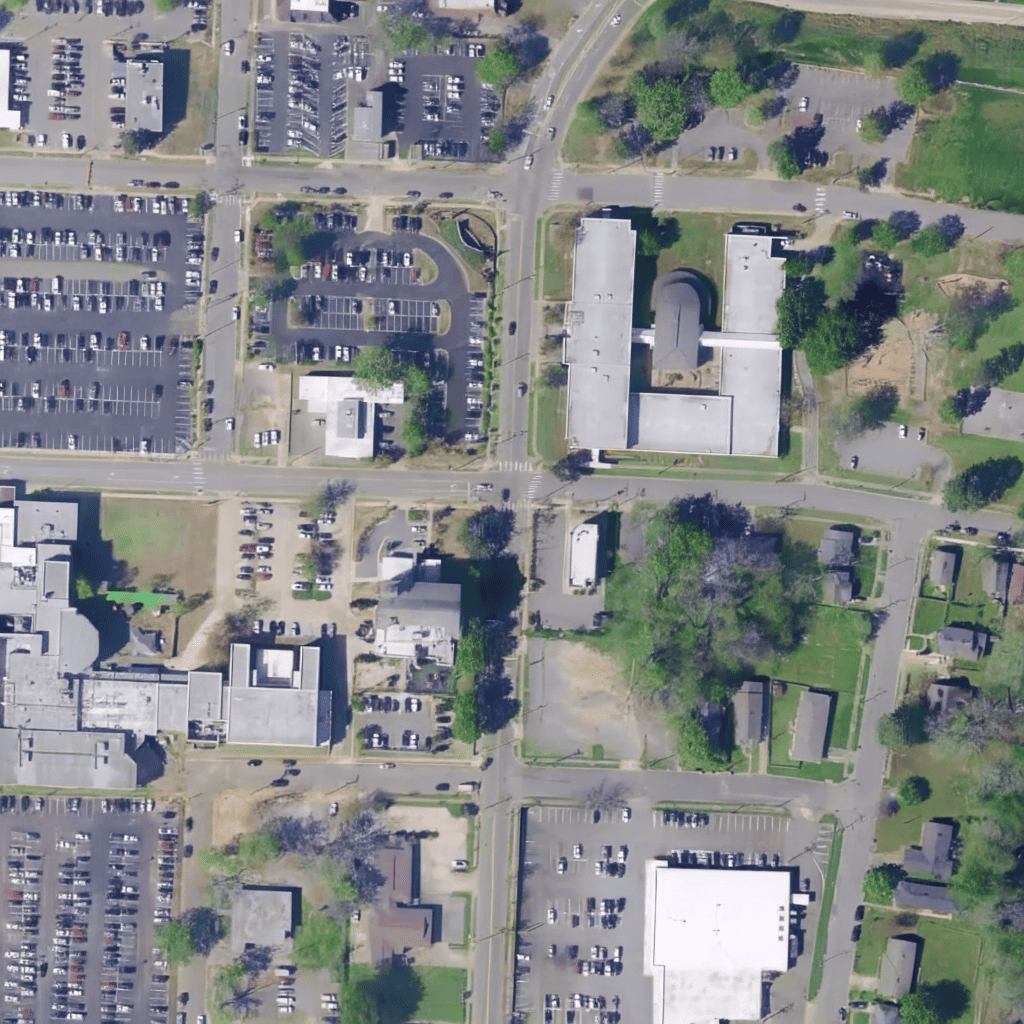}
    {80.0}{35.0}{8.7}
&
\outputcell
    {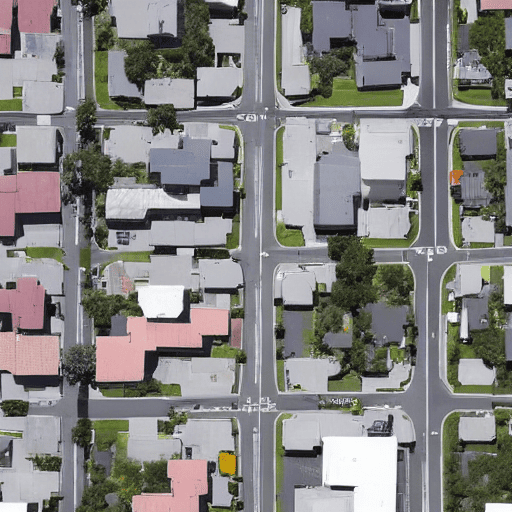}
    {23.3}{57.4}{34.1}
&
\outputcell
    {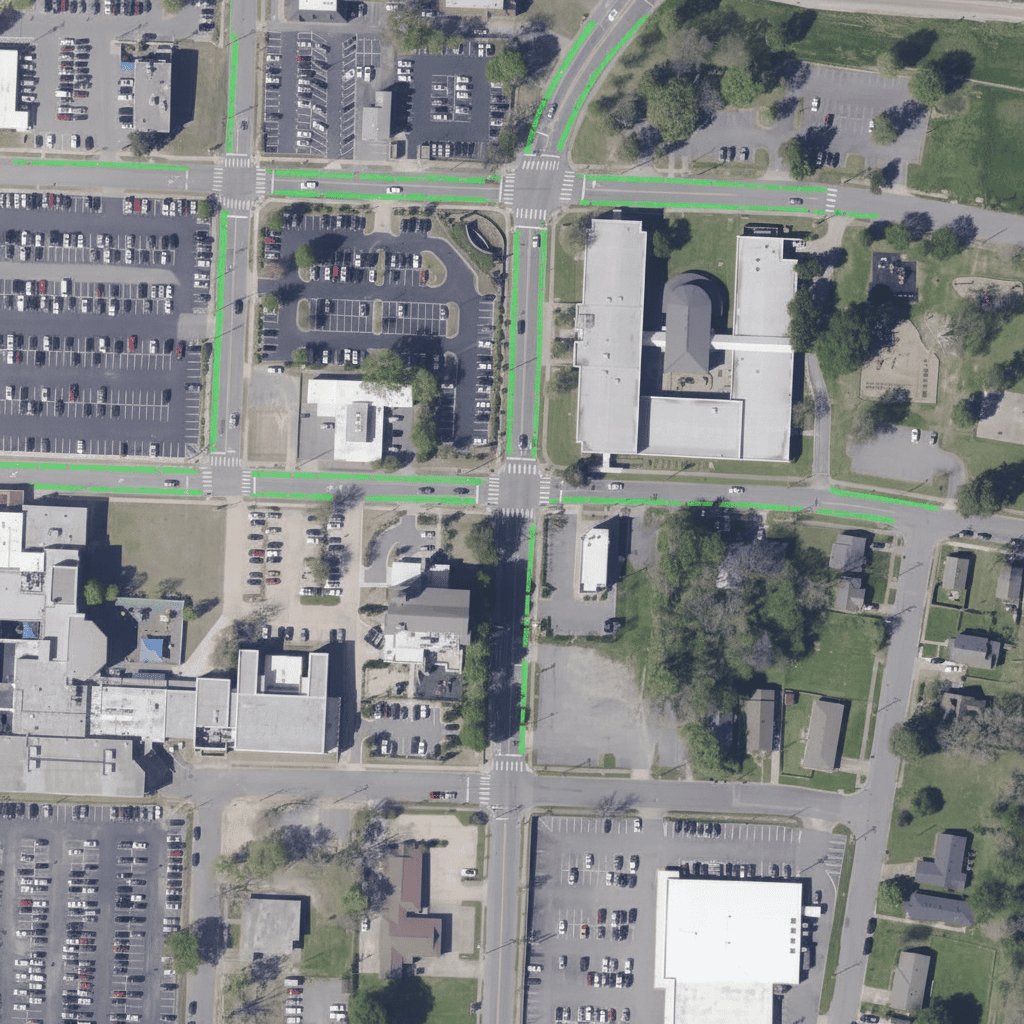}
    {86.7}{38.8}{74.1}
&
\nacell
\\

\bottomrule

\end{tabular*}

\end{center}


\clearpage

\begin{table*}[h]
\centering
\caption{Qualitative examples of counterfactual edits and indicator changes.}
\label{tab:qual_examples}
\footnotesize
\setlength{\tabcolsep}{3pt}
\renewcommand{\arraystretch}{1.15}
\begin{tabularx}{\textwidth}{@{} l C{0.15\textwidth} C{0.15\textwidth} Y C{0.15\textwidth} C{0.09\textwidth} @{}}
\toprule
{\scriptsize\textbf{Task}}
& {\scriptsize\textbf{Input}}
& {\scriptsize\textbf{Seg. Mask}}
& {\scriptsize\textbf{Gen. Prompt}}
& {\scriptsize\textbf{Gen. Edit}}
& {\scriptsize\textbf{Score}} \\
\midrule

\rule{0pt}{6pt}%
\multirow{2}{*}{\scriptsize{\textcolor{mygreen}{\textbf{Greenery}}}}
& \begin{tabular}[c]{@{}c@{}}
    \includegraphics[width=\linewidth]{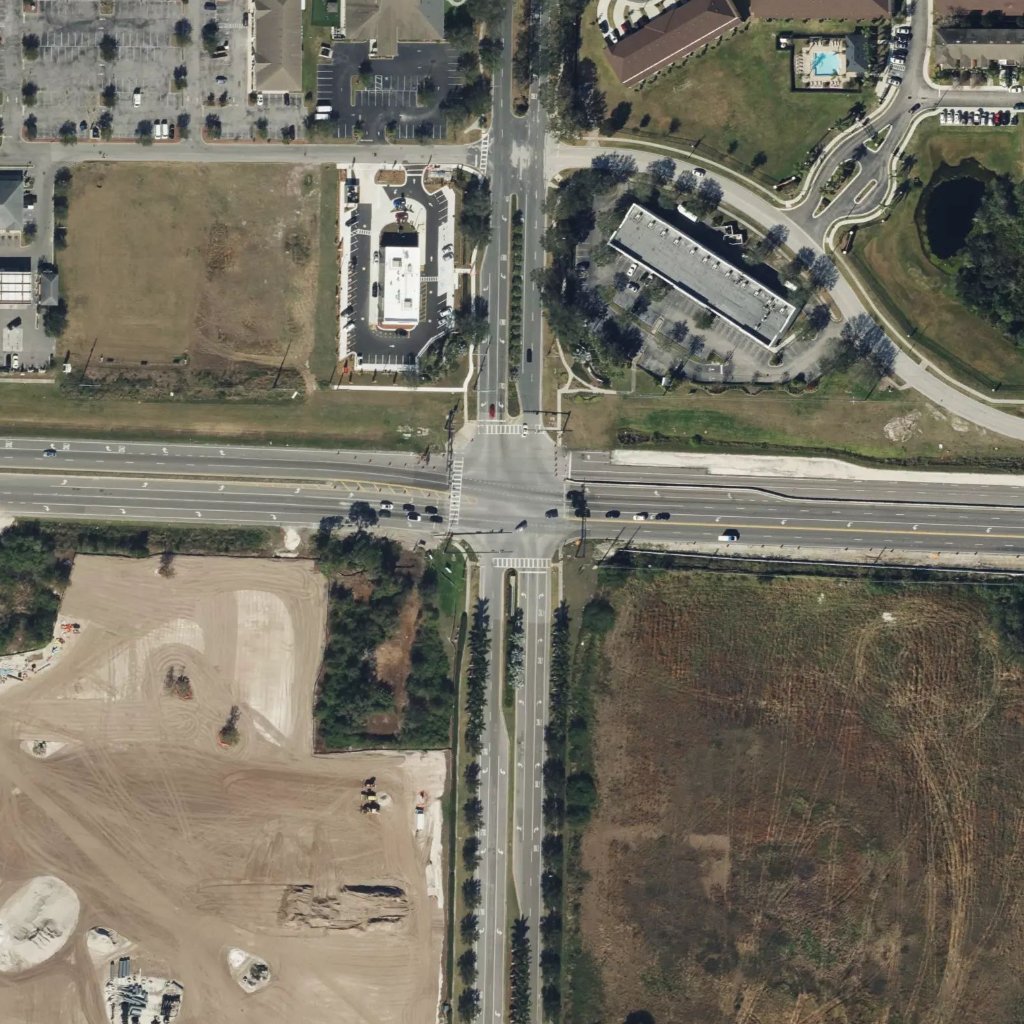} \\[-1pt]
    {\scriptsize\phantom{Seg: LISAt}}
  \end{tabular}
& \begin{tabular}[c]{@{}c@{}}
    {\setlength{\fboxsep}{0pt}\setlength{\fboxrule}{0.6pt}\fbox{\includegraphics[width=\linewidth]{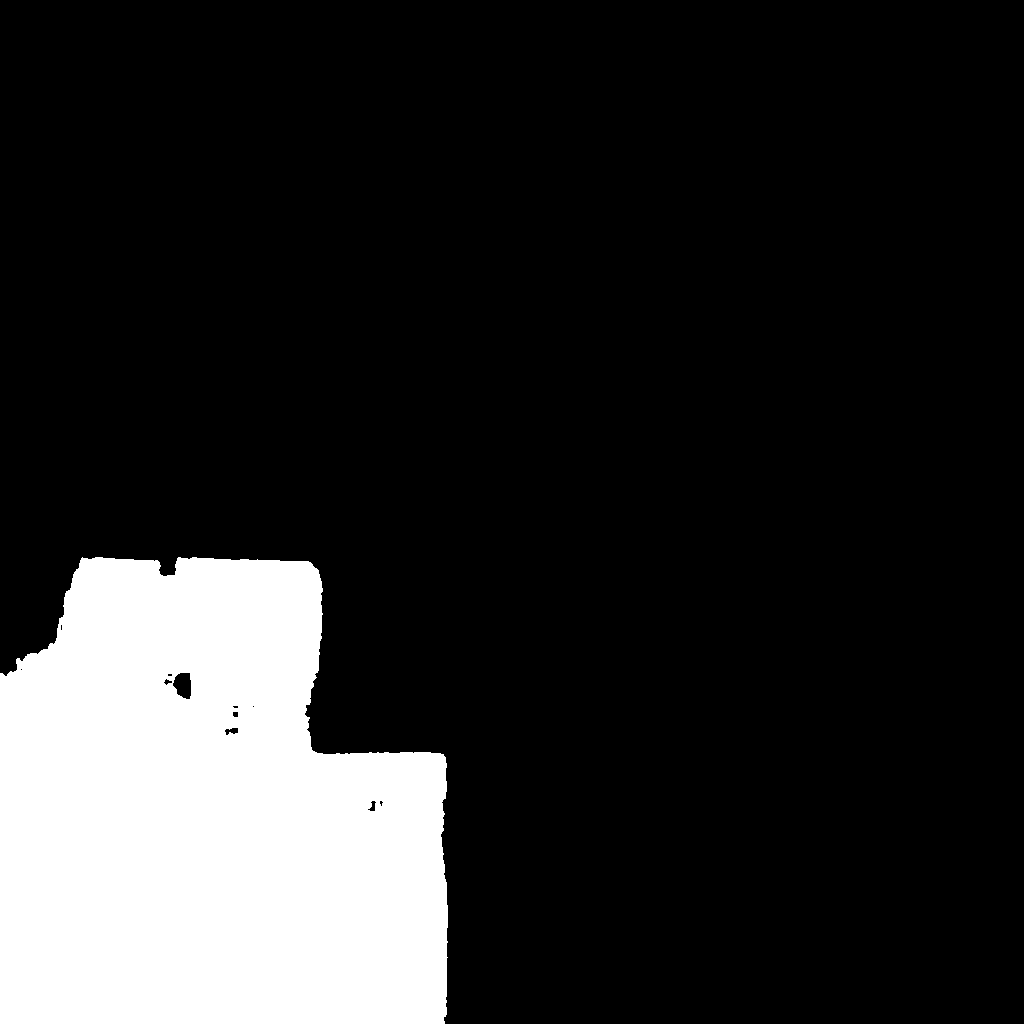}}} \\[-1pt]
    {\scriptsize Seg: \texttt{LISAt}}
  \end{tabular}
& \parbox[c]{\linewidth}{\raggedright\itshape\scriptsize
    ``Introduce a dense mixed-species tree canopy with distinct
    rounded crown shapes and natural shadows.''}
& \begin{tabular}[c]{@{}c@{}}
    \includegraphics[width=\linewidth]{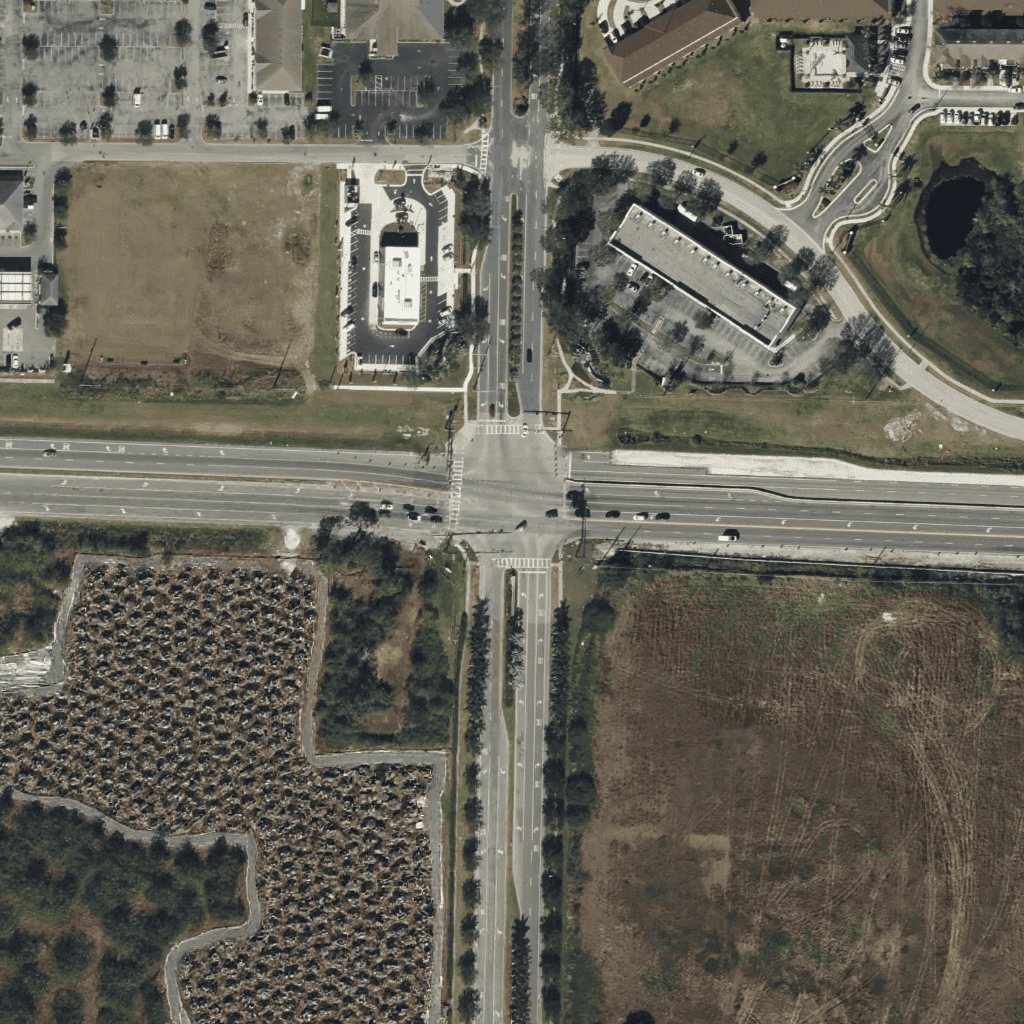} \\[-1pt]
    {\scriptsize Gen: \texttt{FLUX}}
  \end{tabular}
& $+\,$7.1\% \\

\addlinespace[10pt]

& \begin{tabular}[c]{@{}c@{}}
    \includegraphics[width=\linewidth]{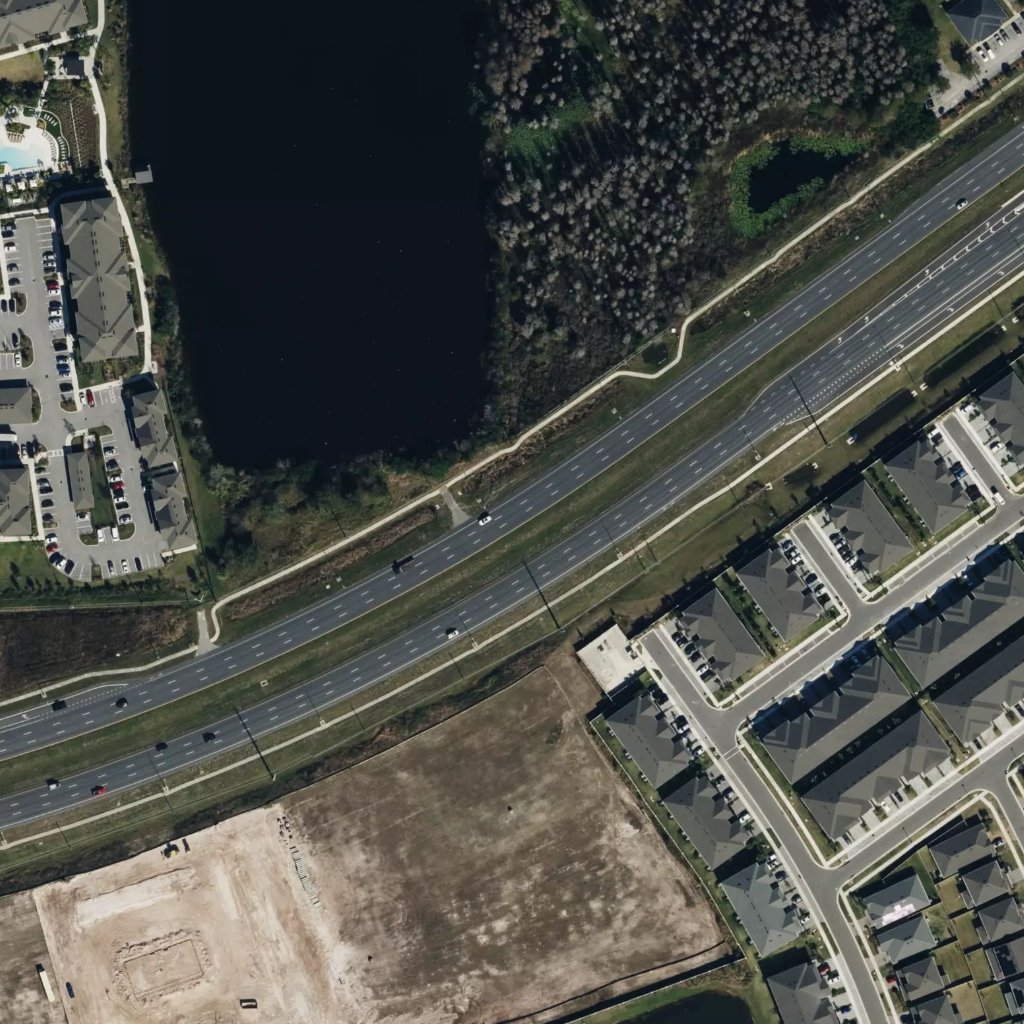} \\[-1pt]
    {\scriptsize\phantom{Seg: SAM3}}
  \end{tabular}
& \begin{tabular}[c]{@{}c@{}}
    {\setlength{\fboxsep}{0pt}\setlength{\fboxrule}{0.6pt}\fbox{\includegraphics[width=\linewidth]{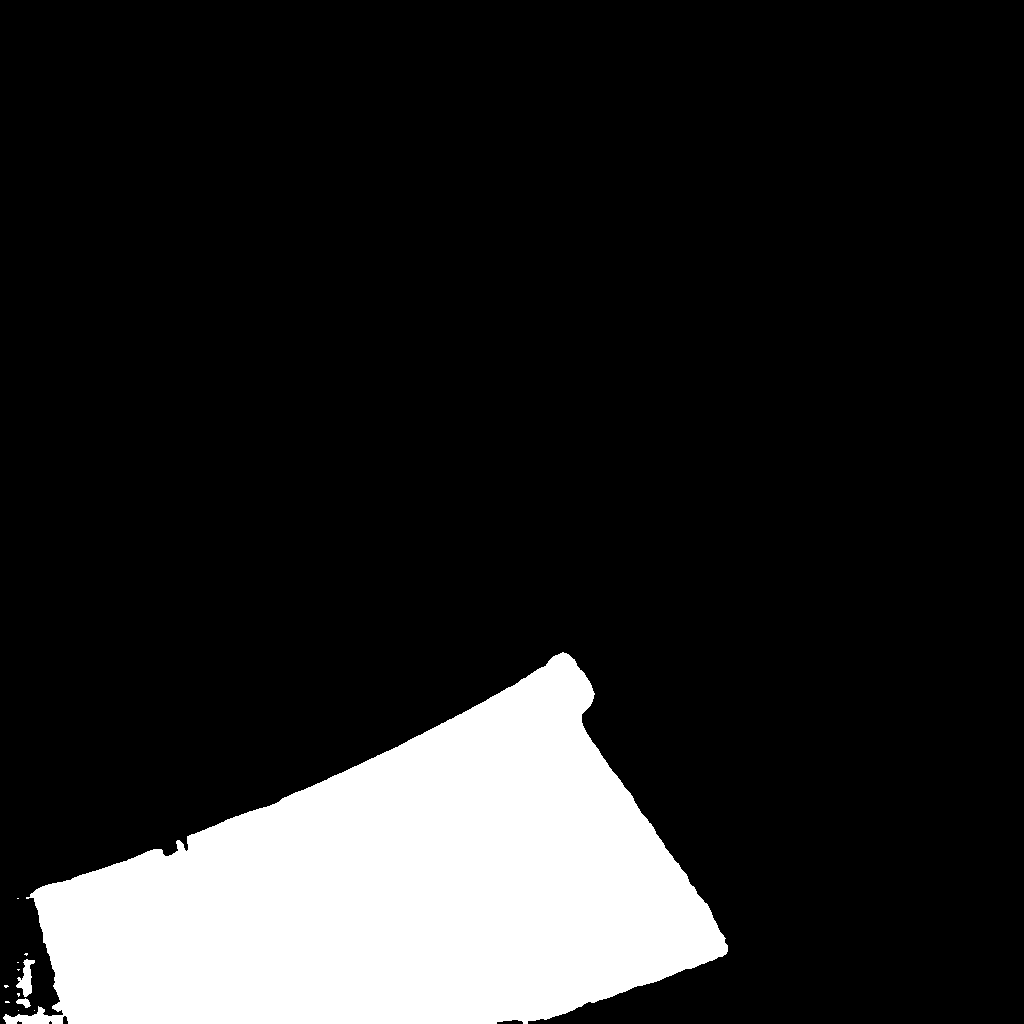}}} \\[-1pt]
    {\scriptsize Seg: \texttt{SAM3}}
  \end{tabular}
& \parbox[c]{\linewidth}{\raggedright\itshape\scriptsize
    ``Convert the masked roadside area into a linear park with a
    dense tree canopy, walking paths, and seating areas.''}
& \begin{tabular}[c]{@{}c@{}}
    \includegraphics[width=\linewidth]{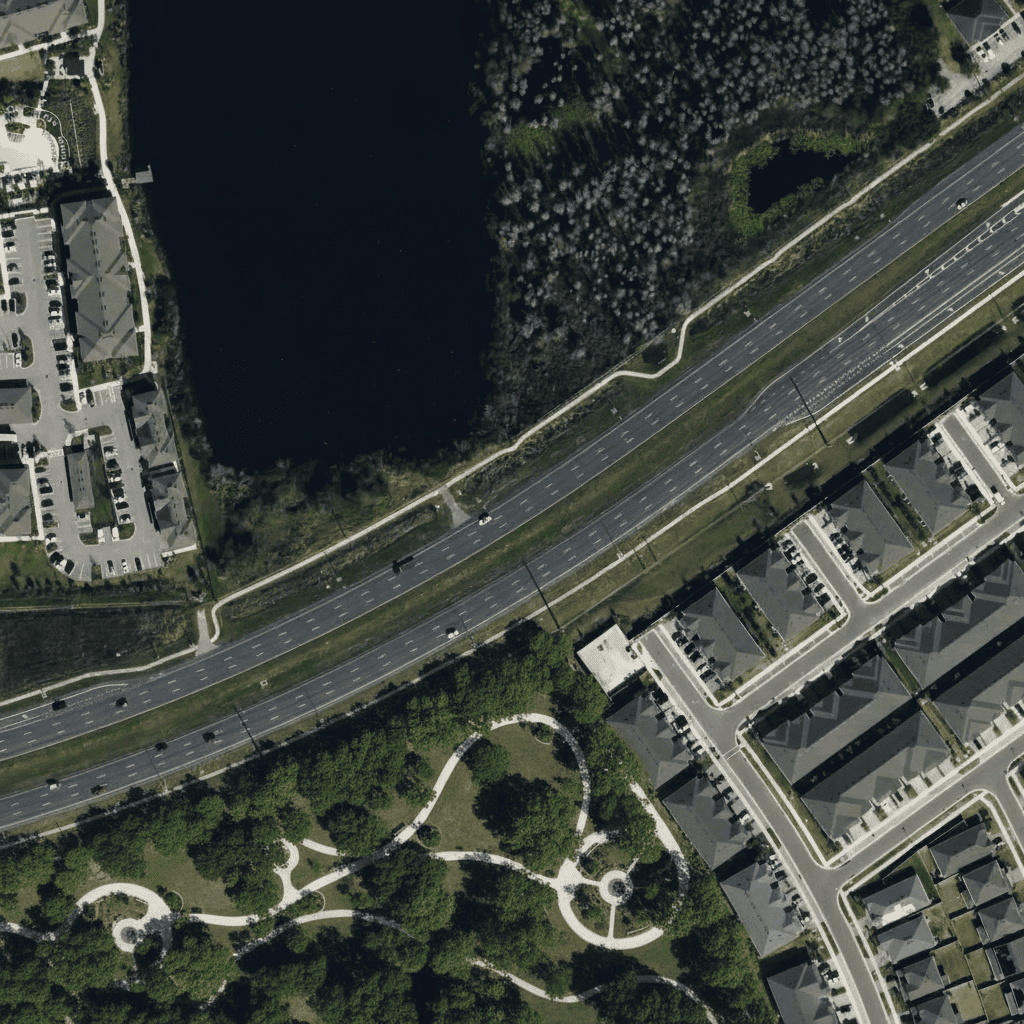} \\[-1pt]
    {\scriptsize Gen: \texttt{NanoBanana}}
  \end{tabular}
& $+\,$5.5\% \\

\addlinespace[4pt]
\midrule
\addlinespace[4pt]

\scriptsize\textbf{\textcolor{myred}{{Road Risk}}}
& \begin{tabular}[c]{@{}c@{}}
    \includegraphics[width=\linewidth]{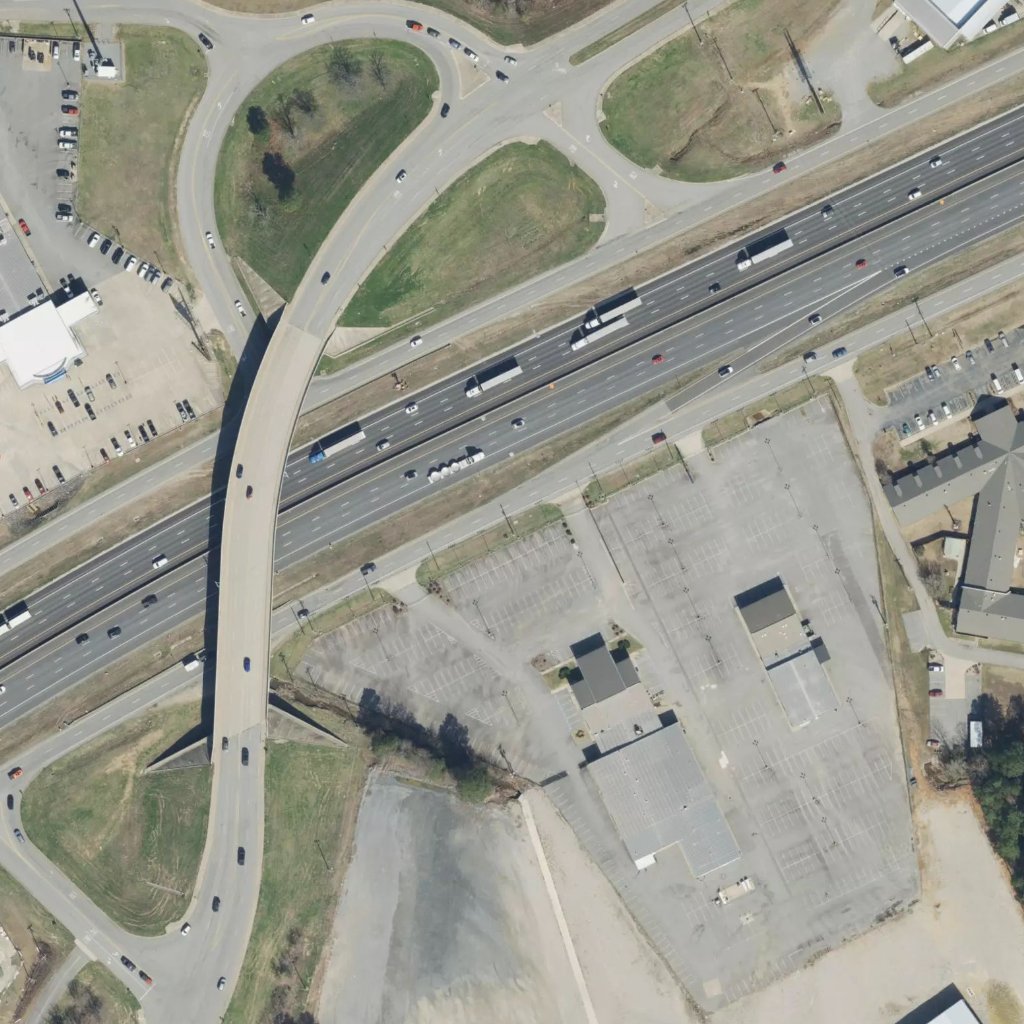} \\[-1pt]
    {\scriptsize\phantom{Seg: SAM}}
  \end{tabular}
& \begin{tabular}[c]{@{}c@{}}
    {\setlength{\fboxsep}{0pt}\setlength{\fboxrule}{0.6pt}\fbox{\includegraphics[width=\linewidth]{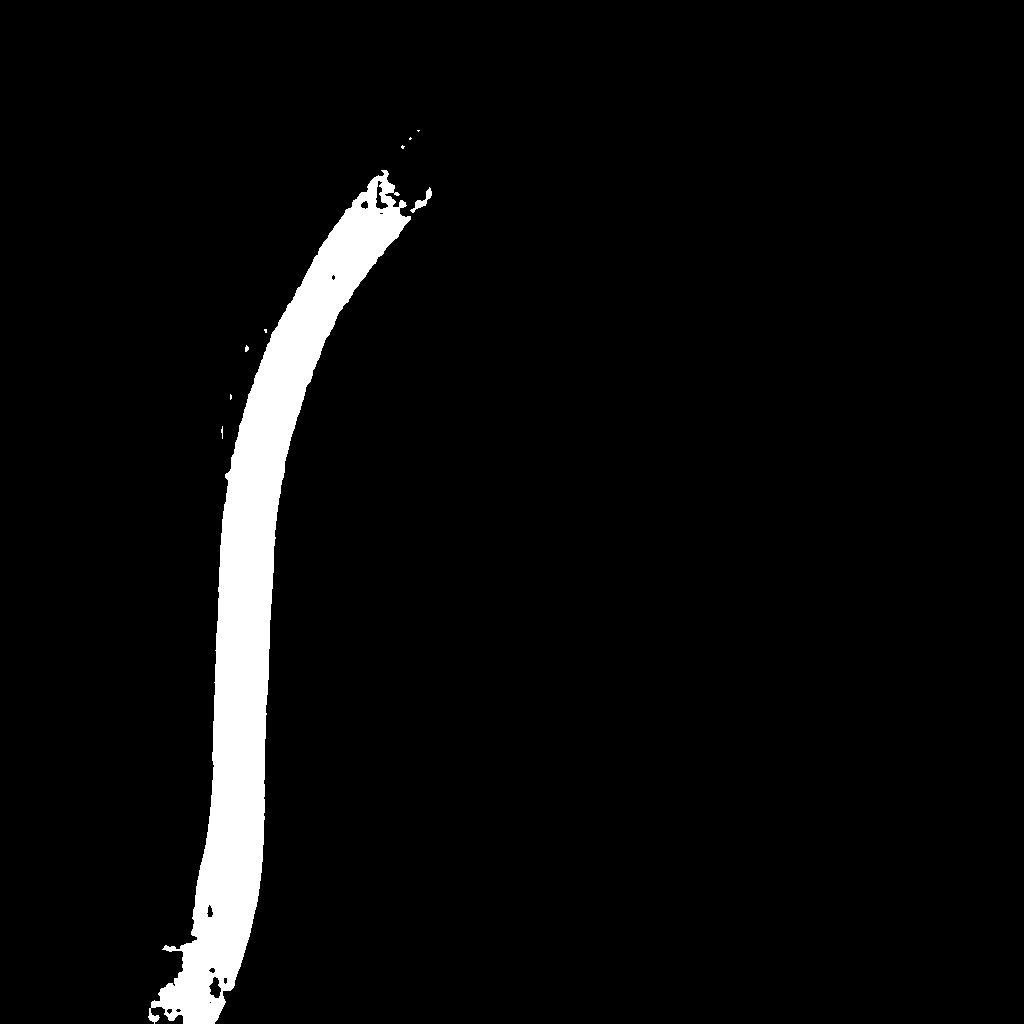}}} \\[-1pt]
    {\scriptsize Seg: \texttt{SAM}}
  \end{tabular}
& \parbox[c]{\linewidth}{\raggedright\itshape\scriptsize
    ``Transform the masked bridge into a pedestrian and cycle bridge with barriers separating it from vehicle lanes.''}
& \begin{tabular}[c]{@{}c@{}}
    \includegraphics[width=\linewidth]{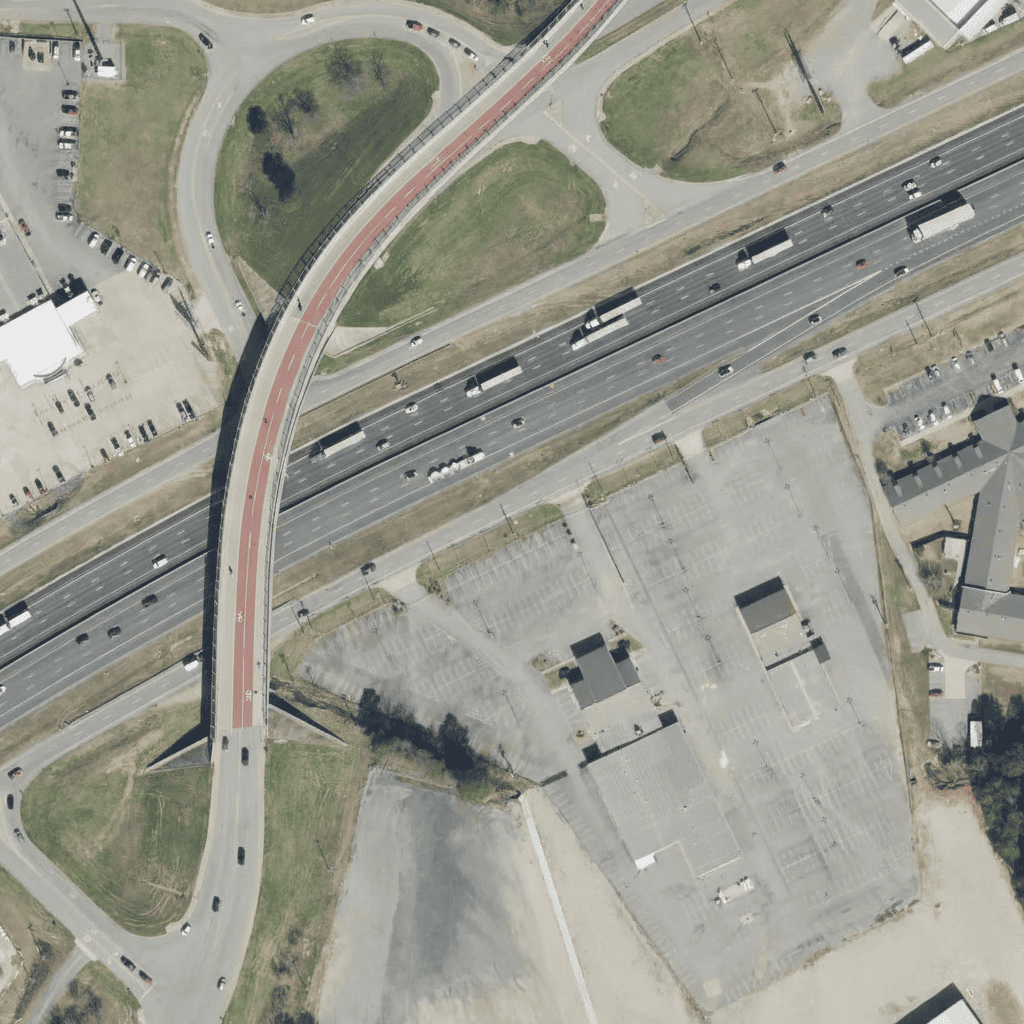} \\[-1pt]
    {\scriptsize Gen: \texttt{NanoBanana}}
  \end{tabular}
& $-\,$9.0\% \\
\addlinespace[4pt]

\bottomrule
\end{tabularx}
\end{table*}

\section{Comparison with Real-World Interventions}
\label{sec:real_edits}
\begin{figure}
    \centering
    \includegraphics[width=\linewidth]{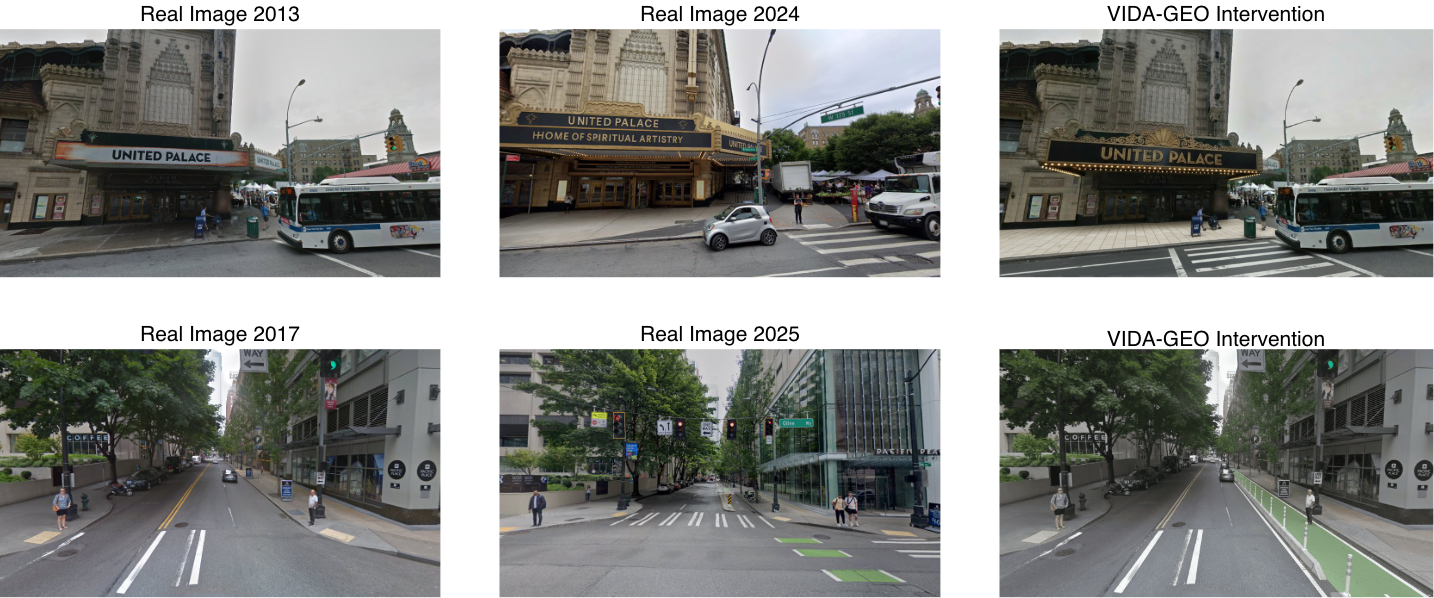}
    \caption{Qualitative comparison of \vida\ interventions with real-world changes. From left to right: the original scene, the observed real-world intervention, and the intervention generated by \vida. In the top example, \vida\ similarly improves the pedestrian crossing, sidewalk, and theater marquee; in the bottom example, it introduces a bike lane similar to the one observed in the real-world intervention.}
    \label{fig:real_edits}
\end{figure}

To examine whether \vida\ proposes interventions that resemble changes already implemented in practice, we qualitatively compare its outputs with real-world interventions. Figure~\ref{fig:real_edits} presents examples of real before-and-after changes alongside interventions generated by \vida\ for similar urban settings. We observe that \vida\ can generate several similar changes that target increases in beautifulness and safety. For example, improving the sidewalks, adding a bike-lane, incorporating a pedestrian-crossing line and improving the theater marquee. 

\section{Ablation Studies Details}
\label{sec:ablation}
Ablation analysis was completed by disabling some of \vida's sub-agents, using a subset of 10 images sampled uniformly from each task. The analysis particularly focused on the Policy Restrictions (restrictor) module, the Prompt Suggestion (suggestor) sub-agent, and the Generation QC and Segmentation QC sub-agents. Around 38.3\% of the outputs of the models with disabled agents do not produce an image that improves upon the baseline score.  Additionally, disabling the QC sub-agents causes a significant drop in scores from 77.7\% to 67.1\% for the Generation Edit QC and 64.2\%  for the Segmentation Mask QC in the average LLM-as-a-Judge score, as detailed in \cref{tab:ablation}.


\begin{table*}[t]
\centering
\caption{Ablation study results. Each row disables one pipeline module (marked with \xmark) while keeping all others active (\cmark). \textit{Visual Quality}, \textit{Realism}, and \textit{Policy Preservation} are scored by an LLM judge; \textit{Judge Avg.} is their mean. \textit{FID-Proxy} measures domain typicality (lower is better).}
\label{tab:ablation}

\footnotesize
\renewcommand{\arraystretch}{1.15}
\setlength{\tabcolsep}{3.5pt}

\resizebox{\textwidth}{!}{%
\begin{tabular}{lccccccccc}
\toprule
Configuration
  & \rotatebox{55}{Edit QC}
  & \rotatebox{55}{Policy}
  & \rotatebox{55}{Mask QC}
  & \rotatebox{55}{Suggestor}
  & \makecell{FID-Proxy\\$\downarrow$}
  & \makecell{Vis.\\Qual. (\%)}
  & Realism (\%)
  & \makecell{Policy\\Pres. (\%)}
  & \makecell{Judge\\Avg. (\%)} \\
\midrule

Full Pipeline (\vida)
  & \cmark & \cmark & \cmark & \cmark
  & \textbf{54.13}
  & \textbf{71.9}
  & \textbf{75.6}
  & \textbf{85.5}
  & \textbf{77.7} \\

\midrule

w/o Edit QC
  & \xmark & \cmark & \cmark & \cmark
  & 57.05 & 65.0 & 68.8 & 67.5 & 67.1 \\

w/o Policy Restrictor
  & \cmark & \xmark & \cmark & \cmark
  & 56.09 & 68.0 & 64.6 & 62.5 & 65.0 \\

w/o Mask QC
  & \cmark & \cmark & \xmark & \cmark
  & 56.79 & 62.5 & 60.0 & 70.1 & 64.2 \\

w/o Suggestor
  & \cmark & \cmark & \cmark & \xmark
  & 55.87 & 60.0 & 65.5 & 80.0 & 68.5 \\

\bottomrule
\end{tabular}%
}
\end{table*}

\section{More Details on the User Interface for Human Evaluation}
\label{sec:UI}
\begin{figure}[ht]
  \centering
  \includegraphics[width=0.7\linewidth]{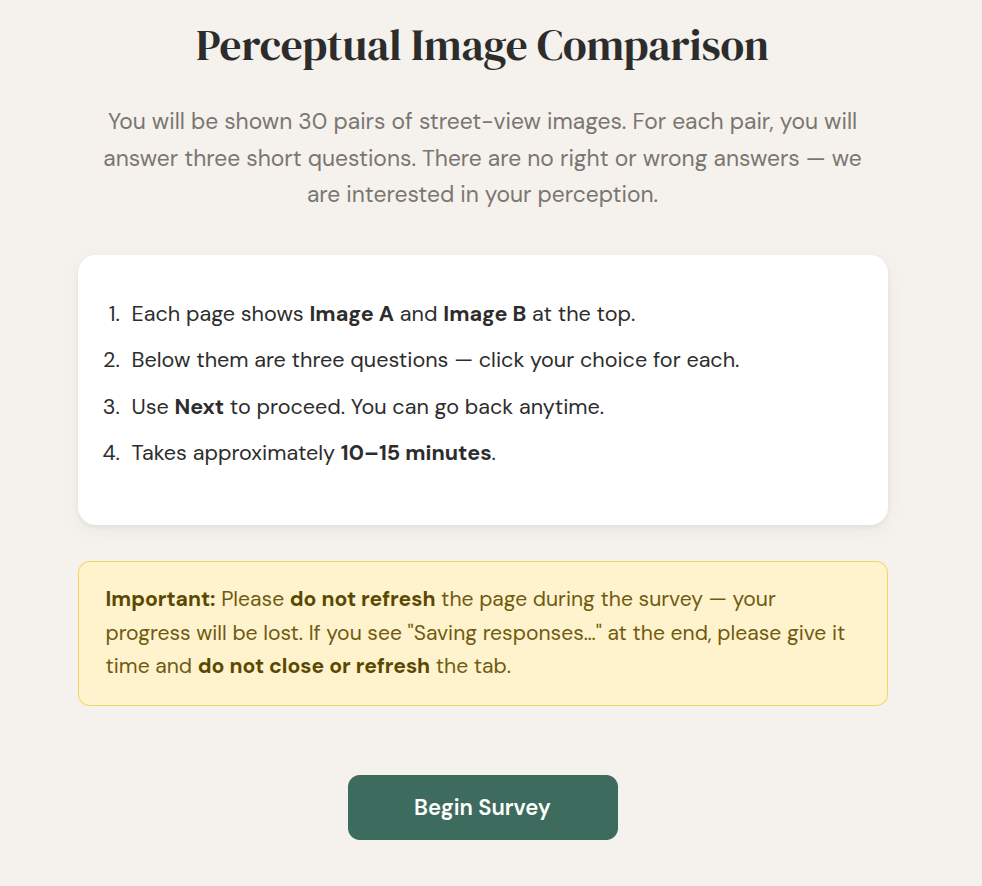}
  \vspace{-6pt}
  \caption{Instructions given to the human evaluators. The instructions were intentionally kept simple to prevent users from biasing themselves to select our method.}
  \label{fig:uiinstr}
  \vspace{-8pt}
\end{figure}

\begin{figure}[ht]
  \centering
  \includegraphics[width=0.65\linewidth]{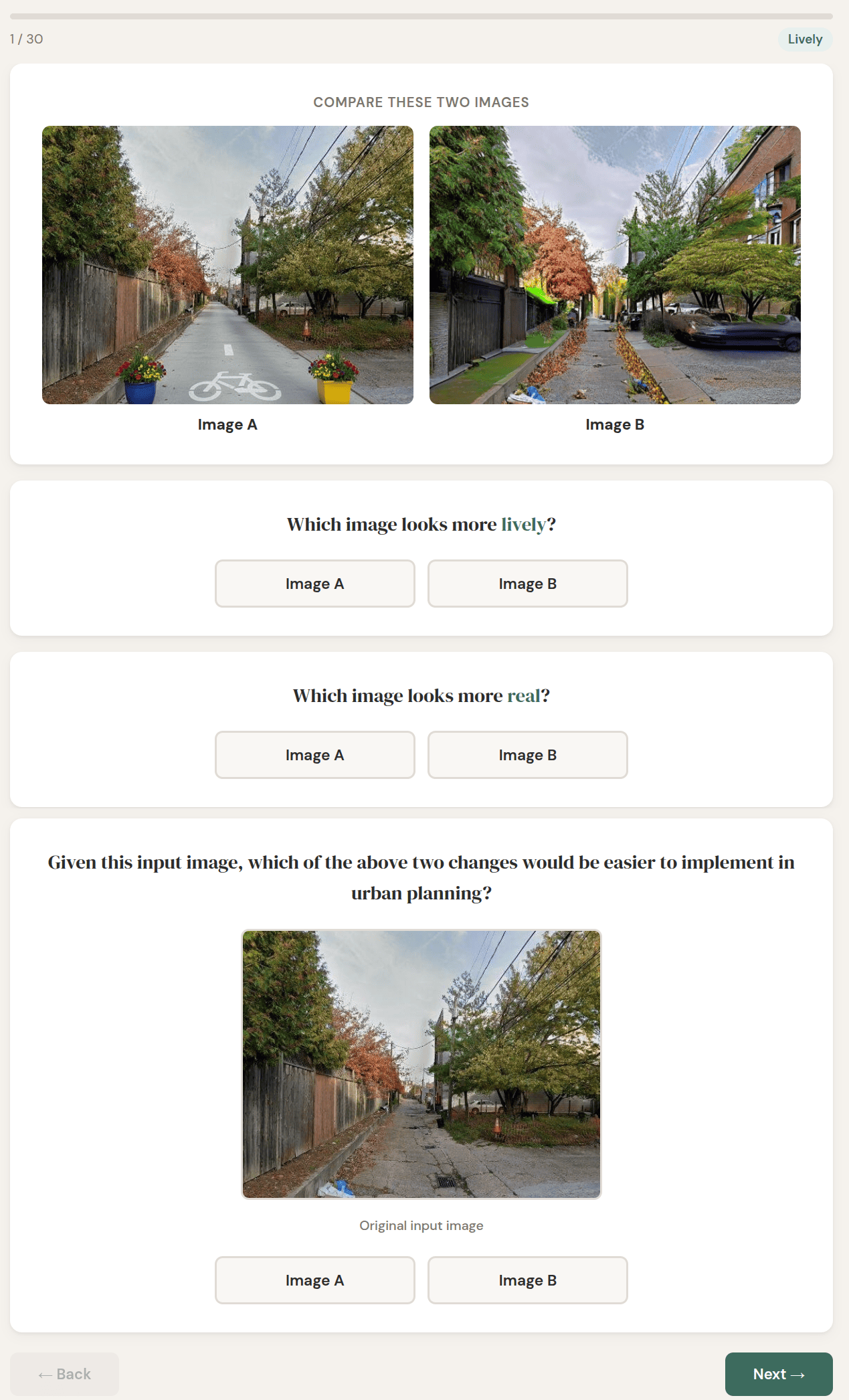}
  \vspace{-6pt}
  \caption{User interface with questions shown to the human annotators. 
  They are randomly shown the best-performing DIFFusion intervention and the best \vida\ intervention, and asked three questions about the indicator, perceptual quality, and policy alignment.
  }
  \label{fig:ui}
  \vspace{-8pt}
\end{figure}

Here we provide more details on the instructions and the user interface (UI) for the human evaluation.
\Cref{fig:uiinstr}, shows the list of instructions given to the human evaluators. 
Since diffusion outputs look very unrealistic, to prevent them from biasing always select \vida\ , we intentionally did not give any examples or explained what we mean exactly by indicator names such as ``safety" or ``wealthiness."
\Cref{fig:ui} shows an example of a UI where the users have to answer three questions about the pair of images: Which image do they believe has a higher change in the indicator score? Which image looks perceptually more real? And changes in which image align better with policy?

\section{Broader Impact of this Work}
\label{sec:impact}
\vida\ can support urban planning by making proposed interventions easier to visualize, compare, and prioritize before costly real-world deployment. It is the first of its kind to help in decision making to support engineers and urban planners in planning and modifying regions with multiple indicators in mind. At the same time, the system should not be treated as a replacement for community engagement, expert review, or detailed engineering studies. Because the generated edits and indicator scores depend on pretrained models, they may inherit geographic, socioeconomic, or cultural biases, and visually plausible interventions may still be legally, financially, or physically infeasible. There is also a risk that such tools could be used to optimize for superficial visual improvements rather than deeper structural needs.

\section{Licenses for Images}
\label{sec:licenses}
We do not redistribute any Google Street View, MapBox or third-party aerial imagery. The released benchmark contains only independently curated coordinates for each of our tasks. Users may obtain imagery from sources whose licenses permit their intended use.


\end{document}